\documentclass[journal]{IEEEtran}
\ifCLASSINFOpdf
\else
\fi
\usepackage{color}
\usepackage{epsfig}
\usepackage{graphicx}
\usepackage{amsmath}
\usepackage{amssymb}
\usepackage{bm}
\usepackage{algorithm}
\usepackage{url}
\usepackage{multirow}
\usepackage{subfigure}
\usepackage{caption}
\usepackage{cases}
\usepackage{multirow}
\usepackage{tabularx}
\usepackage{booktabs} 
\usepackage{algpseudocode} 
\usepackage{longtable}
\usepackage{amsthm}
\usepackage{setspace}
\usepackage{makecell}
\usepackage{array}
\usepackage{booktabs}
\usepackage{amssymb}
\usepackage{pifont}
\newcommand{\xmark}{\ding{55}}%

\begin{document}
%
\title{Real-world Underwater Enhancement: Challenges, Benchmarks, and Solutions}
%
%
%

\author{Risheng~Liu,~\IEEEmembership{Member,~IEEE,}  Xin~Fan,~\IEEEmembership{Senior~Member,~IEEE,}
         Ming~Zhu, Minjun~Hou, and Zhongxuan~Luo
        \thanks{R. Liu, X. Fan, M. Zhu, H. Hou, and Z. Luo are with DUT-RU International School of Information Science \& Engineering and Key Laboratory for Ubiquitous Network and Service Software of Liaoning Province, Dalian University of Technology, Dalian, Liaoning Province, P. R. China. (E-mail: xin.fan@ieee.org).}}
\markboth{Manuscript submitted to IEEE Transactions on Image Processing}%
{Shell \MakeLowercase{\textit{et al.}}: Bare Demo of IEEEtran.cls for IEEE Journals}
%



\maketitle

\begin{abstract}

	Underwater image enhancement is such an important low-level vision task with many applications that numerous algorithms have been proposed in recent years. These algorithms developed upon various assumptions demonstrate successes from various aspects using \emph{different} data sets and \emph{different} metrics. In this work, we setup an undersea image capturing system, and construct a large-scale \textit{Real-world Underwater Image Enhancement} (RUIE) data set divided into three subsets. The three subsets target at three challenging aspects for enhancement, i.e., image visibility quality, color casts, and higher-level detection/classification, respectively. We conduct extensive and systematic experiments on RUIE to evaluate the effectiveness and limitations of various algorithms to enhance visibility and correct color casts on images with hierarchical categories of degradation.
	Moreover, underwater image enhancement in practice usually serves as a preprocessing step for mid-level and high-level vision tasks. We thus exploit the object detection performance on enhanced images as a brand new \emph{task-specific} evaluation criterion. The findings from these evaluations not only confirm what is commonly believed, but also suggest promising solutions and new directions for visibility enhancement, color correction, and object detection on real-world underwater images.
	
\end{abstract}


\IEEEpeerreviewmaketitle

\section{Introduction}
\label{sec:intro}
The development and utilization of ocean resources is of great significance to human beings, demanding remotely operated vehicles (ROVs) and autonomous underwater vehicles (AUVs) equipped with imaging systems for effective investigation. Unfortunately, underwater images of low quality bring failures to intelligent computer vision systems for visual inspections, environmental sensing, and object detection and recognition. Therefore, it is crucial to develop underwater image enhancement technology for the benefit of more underwater computer vision tasks.

\begin{figure}[t]
	\begin{center}
		\begin{tabular}{c@{\extracolsep{0.4em}}c}	
			\includegraphics[width=.23\textwidth]{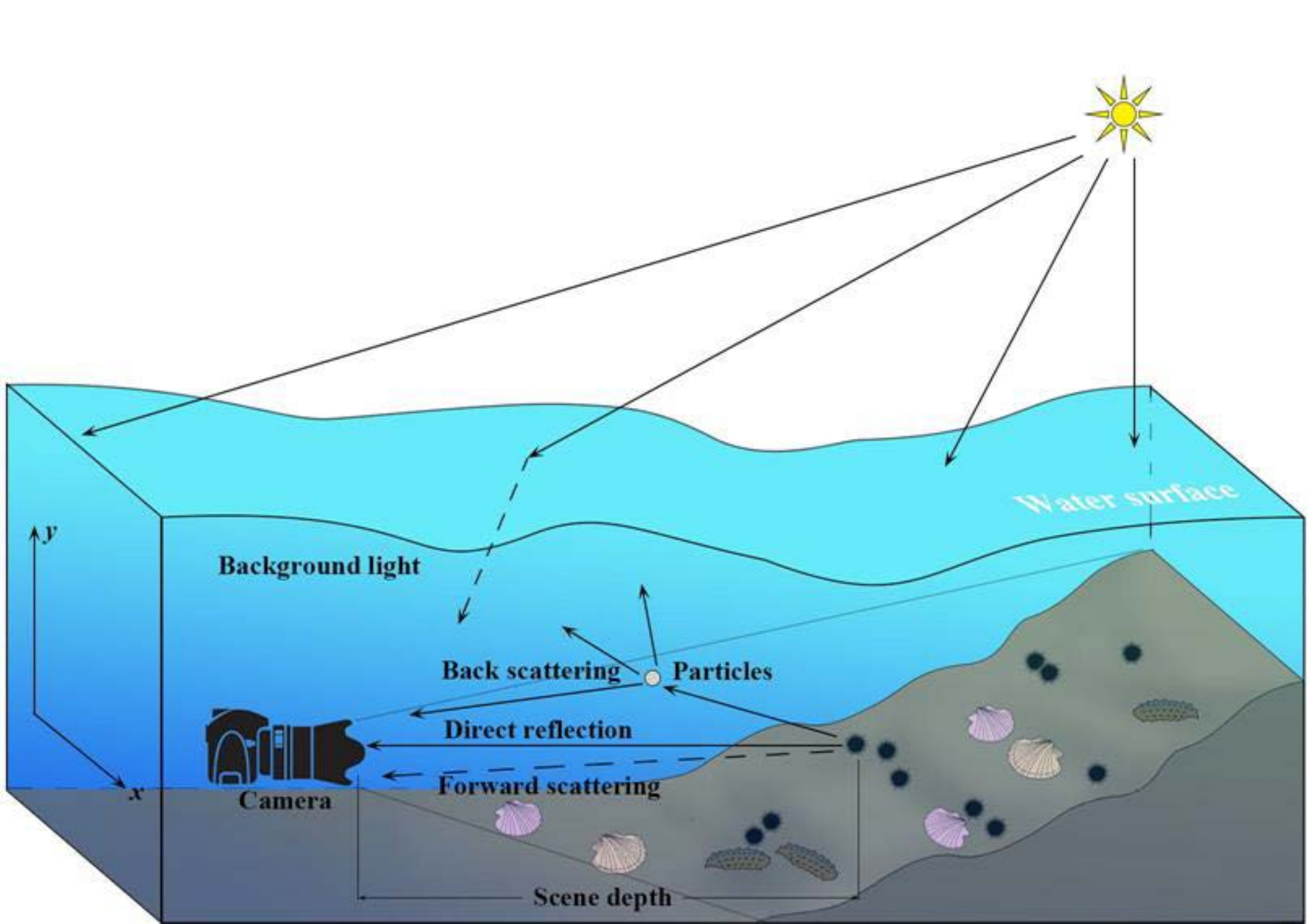}&
			\includegraphics[width=.23\textwidth]{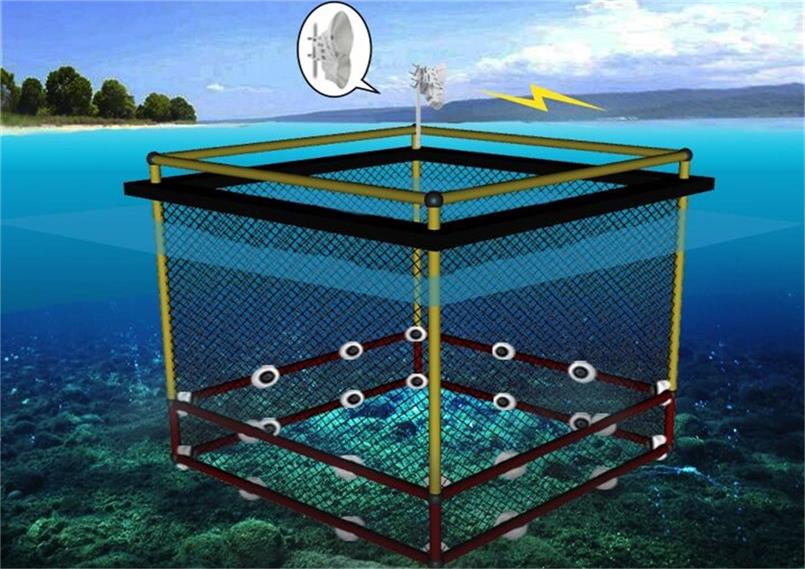}\\
			(a)	&(b)
		\end{tabular}
	\end{center}
	\caption{Schematic diagram of underwater optical imaging and the underwater image capturing system of the proposed RUIE benchmark.}\label{fig:model}		
\end{figure}

As shown in Fig.~\ref{fig:model}, there are two major factors leading to the degradation of underwater images. Firstly, the reflected light from underwater scene is absorbed and scattered by suspending particles in the medium before reaching the camera, resulting in low contrast and haze-like effects. Secondly, the attenuation of light, depending on optical wavelength, dissolved organic compounds, and water salinity, causes various degrees of color casts. For examples, underwater images always look bluish or greenish as the red light having a longer wavelength is absorbed more than the green and blue. 

Accordingly, two essential objectives of underwater image enhancement (UIE) algorithms are to combat the effect of light scattering (similar to dehazing) and to correct color casts. Improving the accuracy for following higher-level detection/classification tasks is one additional objective of enhancement when the UIE algorithms serve as a preprocessing step. Numerous UIE algorithms have been proposed upon various assumptions to address these degradation issues. Generally, UIE algorithms can be categorized into three types including model-free, model-based, and learning-based convolutional neural networks (CNNs). Traditional white balance adjustment, histogram equalization and fusion techniques~\cite{Jobson2002A,pizer1990contrast,Ancuti2012Enhancing,gibson2015preliminary} fall into the model-free category. Researchers also develop underwater imaging models to characterize the physical formation process of underwater images. The prior-driven methods employ various types of field knowledge to estimate  depth-dependent transmission map, and further recover images of higher visibility~\cite{Carlevaris2010Initial,chiang2012underwater,UnderwaterHL,Wen2013Single,Jr2013Transmission,Galdran2015Automatic,Li2016Underwater,Peng2017Underwater}. 
Recent data-driven frameworks either design end-to-end CNNs~\cite{Li2017WaterGAN,Chen2017Towards}, or integrate CNNs with physical priors~\cite{liu2018learning,Liu2019learning} to learn these essential parameters or transmission maps from the degraded inputs.

Existing UIE algorithms in literature are generally evaluated using \emph{different} data sets, among which many are synthetic, and using \emph{different} metrics upon a certain quality index such as contrast, saturation and luminance. In view of the rapid progress of underwater image enhancement algorithms, it is necessary to enrich a large-scale real-world benchmark for the algorithm evaluation as well as the generation of synthetic images for training data-driven networks\footnote{Considering the emerging and popularity of generative adversarial networks (GANs), real-world images are required for discriminators in GANs to generate `sythentic' training examples.}. Moreover, the evaluation merely on the visibility is not enough for the enhancement having multiple objectives/aspects. Specifically, three major limitations exist in current though rare underwater data sets~\cite{chiang2012underwater,Ancuti2012Enhancing,Galdran2015Automatic,li2018watergan}, especially for the purpose of performance evaluation: 1) many existing data sets are unsuitable for evaluating the performance of visibility enhancement, especially those prior model driven methods, as the scene depth is shallow and the scattering effect is subtle in these sets; 2) the scenes and tones of these data sets are relatively monotonous, making it difficult to evaluate how the algorithms work under different illuminations and color casts; 3) there are few marine organisms in images, which limits the application of these databases for evaluating the effectiveness of enhancement to higher-level tasks.

In this study, we build a benchmark data set with \emph{real world sea} images to overcome the three limitations, and also compare the state-of-the-art to suggest the effective and efficient solutions to the problem of underwater image enhancement. Our contributions are summarized as follows:

\begin{itemize}
	\item 
	
	We setup a multi-view imaging system under sea water, and construct a large-scale underwater benchmark, the \textit{Real-world Underwater Image Enhancement} (RUIE) data set, with over $4,000$ images\footnote{All images are available at https://github.com/dlut-dimt/Realworld-Underwater-Image-Enhancement-RUIE-Benchmark}. Table~\ref{table:subset} lists the profile of RUIE, and Fig.~\ref{fig:example1} shows several image examples. Compared with exiting realistic image sets from underwater scenes, the RUIE includes a large diversity of images, which are divided into three subsets targeting at the evaluation for three objectives of image enhancement algorithms.
	\item We conduct substantial and systematic experiments on RUIE to evaluate the performance of various algorithms in processing images with multiple degrees of degradation and different types of color casts. Both quantitative and qualitative analysis demonstrate the advantages and limitations of every algorithm for evaluation. Not only do the findings from these experiments confirm what is commonly believed, but also bring insights for new research directions in underwater image enhancement.
	\item We also devise a task-specific evaluation criterion for enhancement algorithms that exploits the object detection accuracy on the enhanced images.
	Experimental results reveal that there exists no strong correlation between current quality metrics for underwater images and the accuracy for object classification when the underwater images are preprocessed by the existing enhancement methods. This discovery may suggest a new research perspective that considers low-level enhancement and higher-level detection/classification as a whole instead of two cascaded independent processes.  
		
\end{itemize}

This paper is organized as follows. Section~\ref{sec:related_work} surveys algorithms addressing the challenges for the enhancement of underwater images. Section~\ref{sec:data_set} details the RUIE benchmark set, followed by experimental evaluations on RUIE and discussions on the results that suggest solutions to enhancement in Section~\ref{sec:results}. Section~\ref{sec:conclusions} concludes the paper.

\begin{table*}[t]
	\caption{Overview of representative UIE algorithms.  The labels ``R", ``S", and ``CC" in the ``test data" column represent real world, synthetic, and ColorChecker images. The ``Criterion" column lists the metrices of Mean Squared Error (MSE), Root Mean Squard Error (RMSE), Peak Signal to Noise Ratio (PSNR), Structural Similarity Index (SSIM), Contrast-to-noise ratio (CNR), Patch-based Contrast Quality Index (PCQI) \cite{Wang2015A}, UCIQE, UIQM,  and Blind/ Referenceless Image Spatial Quality Evaluator (BRISQUE) \cite{Mittal2012No}.} \label{table:method}
	\centering
	\small	
	\renewcommand\arraystretch{1.3}
	\begin{tabular}		
		{l|p{4.3cm}<{\centering}|p{1.6cm}<{\centering}|p{2cm}<{\centering}|p{4.3cm}<{\centering}|p{0.5cm}<{\centering}}\toprule
		\multicolumn{5}{c}{model-free methods}\\\hline
		Method & \multicolumn{2}{c|}{Enhancement technique}& Test data &Criterion&Code\\\hline
		UCM \cite{Iqbal2010Enhancing}   &\multicolumn{2}{c|}{Unsupervised color balance and histogram stretching}&R &Histogram distribution&$\checkmark$\\ 
		MSRCR \cite{Jobson2002A}		&\multicolumn{2}{c|}{Multiscale retinex with color restoration}	&R &\xmark	&$\checkmark$\\
		CLAHE \cite{pizer1990contrast}  &\multicolumn{2}{c|}{Contrast-limited adaptive histogram equalization}	&R &\xmark	&$\checkmark$\\
		CLAHE-MIX\cite{Hitam2013Mixture}&\multicolumn{2}{c|}{Mixture RGB and HSV CLAHE}&R &	MSE, PSNR  &\xmark \\
		Fusion \cite{Ancuti2012Enhancing}&\multicolumn{2}{c|}{White Balance, bilateral filtering, image fusion}&R, CC& Local feature points matching &$\checkmark$\\	
		Ghani \cite{Ghani2015Underwater} &\multicolumn{2}{c|}{minimizes under-enhanced and over-enhanced areas}&R  &	Entropy, MSE, PSNR &	\xmark\\
		\toprule
		\multicolumn{5}{c}{Prior-based methods}\\ \hline
		Method   & Physical prior& Post process& Test data &Criterion&Code\\  \hline
		BP \cite{Carlevaris2010Initial}	&Radiance attenuation &\xmark&R/ CC&\xmark &$\checkmark$\\
		P. Drews-Jr \cite{Jr2013Transmission} &Underwater DCP on g,b&\xmark&R/ CC&RMSE&$\checkmark$\\
		UHP \cite{UnderwaterHL} &Color distribution&$\checkmark$ &R & RGB Median angle	&$\checkmark$	\\
		NOM \cite{Wen2013Single}& Underwater DCP&\xmark&R &\xmark	&$\checkmark$	\\
		Li \cite{li2014a}&Underwater DCP&$\checkmark$&R	&\xmark	&$\checkmark$\\
		LDP \cite{lcy2016underwater}&Histogram distribution prior& $\checkmark$	&R, S, CC	&MSE, PSNR, Entropy, PCQI, UCIQE 	&\xmark\\			
		Peng \cite{Peng2017Underwater}&Blurriness\& Light Absorption&\xmark &	R, S &PSNR, SSIM, BRISQUE, UIQM	&\xmark\\
		WCID \cite{chiang2012underwater}&Residual energy ratios&$\checkmark$ &R, CC&	\xmark  &$\checkmark$\\
		Galdran \cite{Galdran2015Automatic}&Red channel prior&\xmark &R&	Edge number, Gradient ratio &	\xmark \\
		Lu \cite{Lu2015Contrast}&UDCP with median filter&$\checkmark$  &R  &	PSNR, CNR, SSIM& 	\xmark\\
		Li \cite{Li2016Underwater} &UDCP with median filter&$\checkmark$  &R  &	CNR &	\xmark\\
		Yang \cite{Yang2012Low}  &UDCP with median filter &$\checkmark$&R  &	\xmark&	\xmark \\		
		DPATN \cite{liu2018learning}&Learning-based UDCP&$\checkmark$&R&\xmark &$\checkmark$\\
		\hline
	\end{tabular}
\end{table*}

\section{Underwater Image Enhancement Algorithms}
\label{sec:related_work}

Typical UIE algorithms aim to produce a high-quality image that human favor from a single degraded input. These algorithms either increase the visibility or alleviate color casts by combating the light scattering and other ambient circumstance factors during capturing underwater scenes. According to the means of modeling imaging process, we roughly categorize existing UIE algorithms into the following three types.
\subsection{Model-free Methods}
This type of algorithms adjusts pixel values of a given image without explicitly modeling the image formation process. The adjustments may be performed in the spatial or transform domain. The spatial domain methods include histogram equalization~\cite{Hummel1977Image}, the Gray World algorithm~\cite{Buchsbaum1980A}, contrast limited adaptive histogram equalization (CLAHE)~\cite{pizer1990contrast}, multi-scale retinex with color restore \cite{Jobson2002A}, automatic white balance \cite{Liu2004Automatic}, and color constancy \cite{van2010Edge, Foster2011Color}. The transform domain methods map image pixels into a specific domain where we exploit the physical properties to perform adjustments. The commonly used transforms include Fourier and wavelets~\cite{singh2015underwater}.

The spatial domain methods can improve the visual quality to some extent, but may accentuate noise, introduce artifacts, and cause color distortions. The transform domain methods perform well in smearing noise, but suffer from low contrast, detail loss, and color deviations. Due to the complexity of underwater environment and illumination conditions, these enhancement techniques, merely relying on the observed information, can hardly recover high quality images from underwater degradation.

\subsection{Model-based Methods}
\begin{figure*}[t]
	\begin{center}
		\begin{tabular}{c@{\extracolsep{0.5em}}c@{\extracolsep{0.5em}}c@{\extracolsep{0.5em}}c@{\extracolsep{0.5em}}c@{\extracolsep{0.5em}}c}
			\includegraphics[width=.19\textwidth]{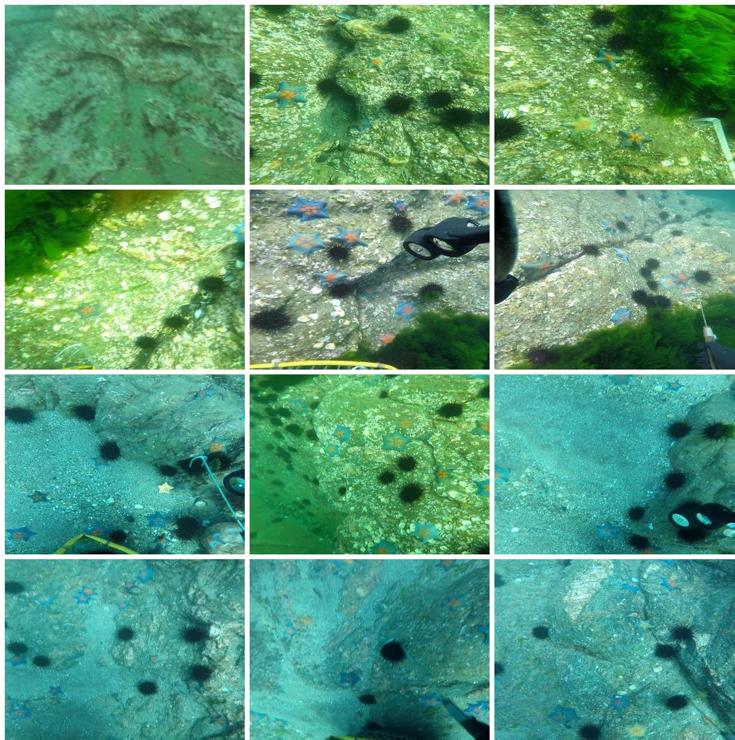}
			&\includegraphics[width=.19\textwidth]{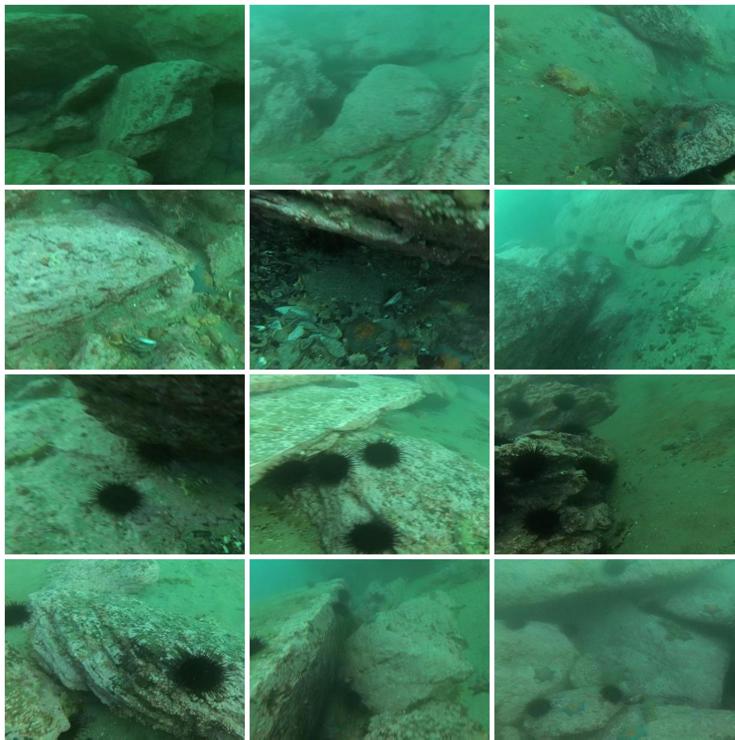}		
			&\includegraphics[width=.19\textwidth]{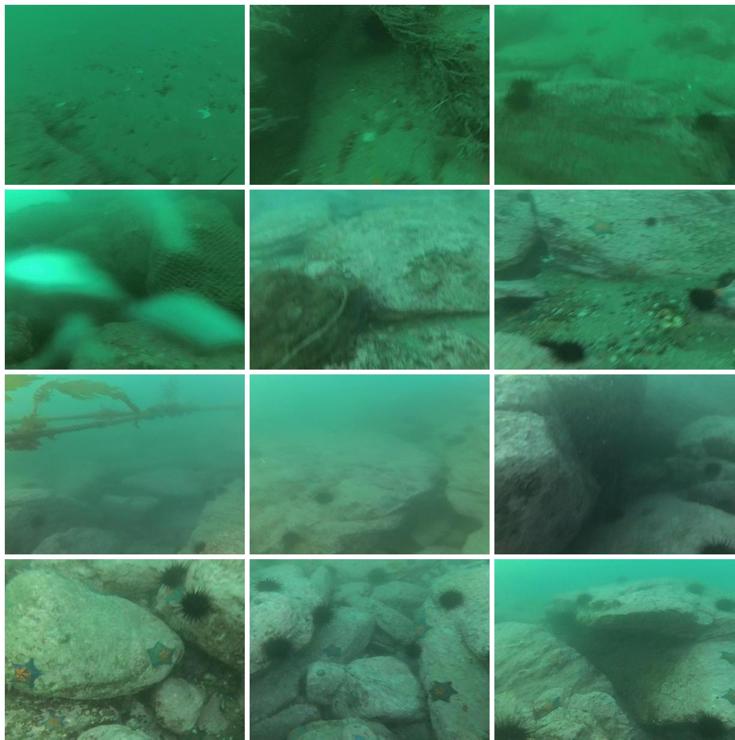}
			&\includegraphics[width=.19\textwidth]{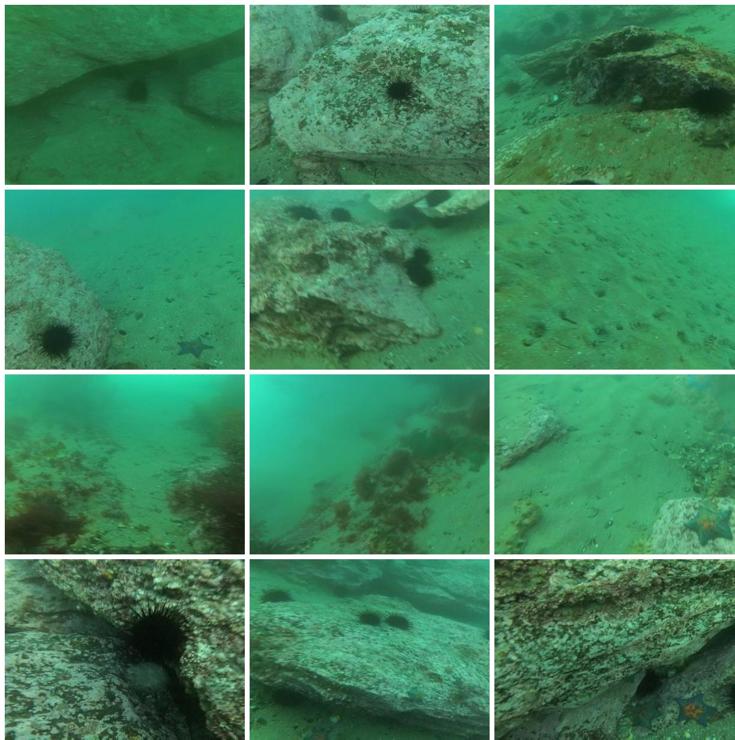}
			&\includegraphics[width=.19\textwidth]{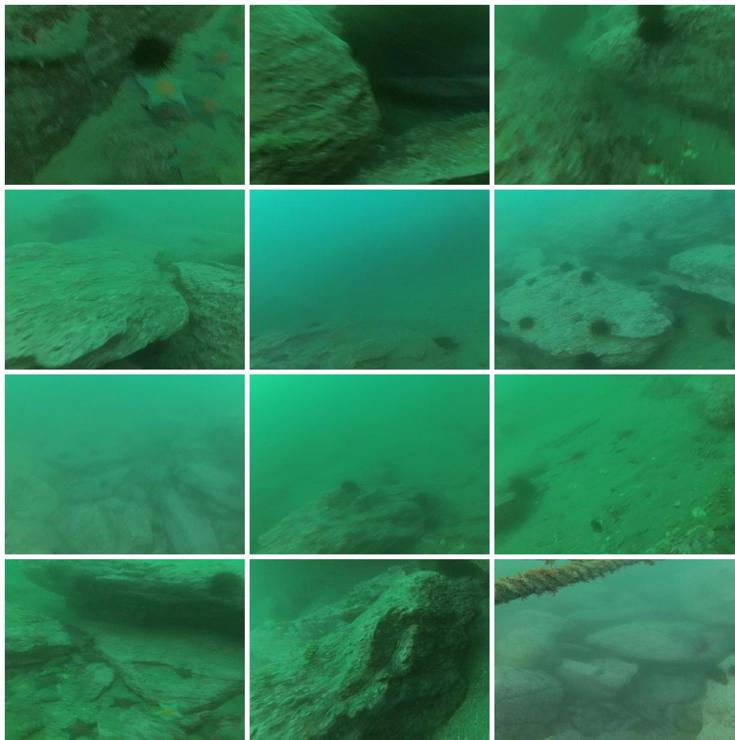}\\
			\multicolumn{5}{c}{\shortstack{(a) Underwater Image Quality Set (UIQS). \\From left to right, images from the five subsets A$\sim$E are ranked according to a non-reference image quality metric, \\and the corresponding image quality successively decreases.}\vspace{0.5em}}​
			
		\end{tabular}	
		\begin{tabular}{c@{\extracolsep{0.5em}}c@{\extracolsep{0.5em}}c}
			\includegraphics[width=.323\textwidth]{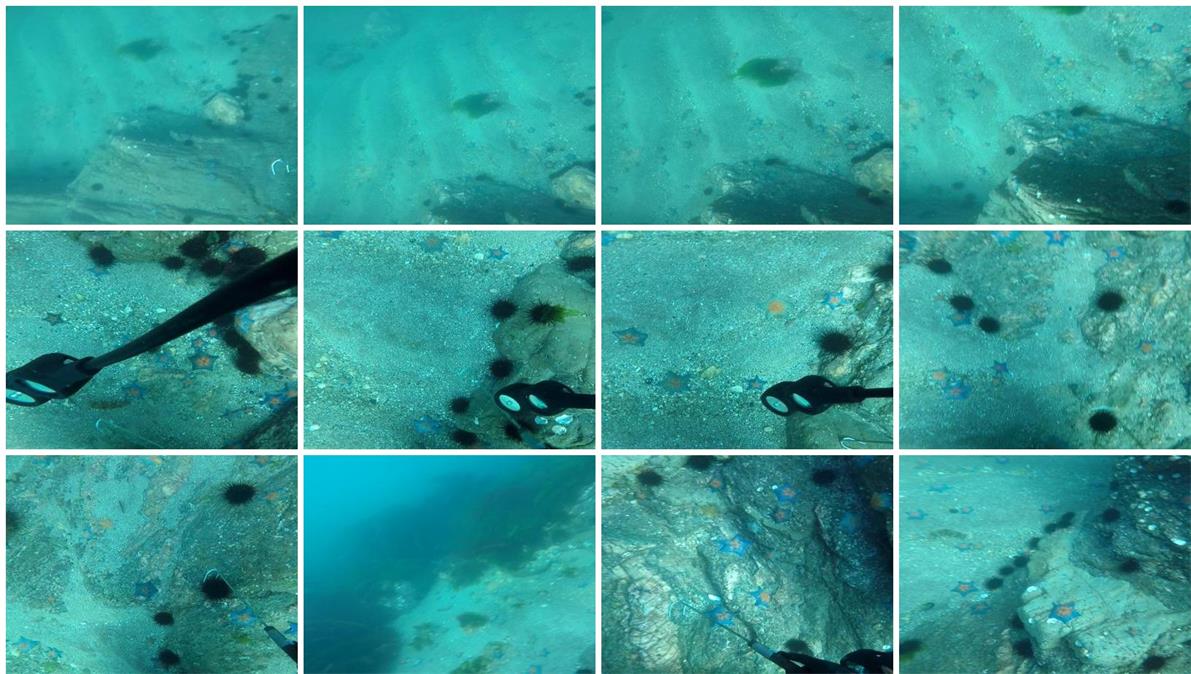}
			&\includegraphics[width=.323\textwidth]{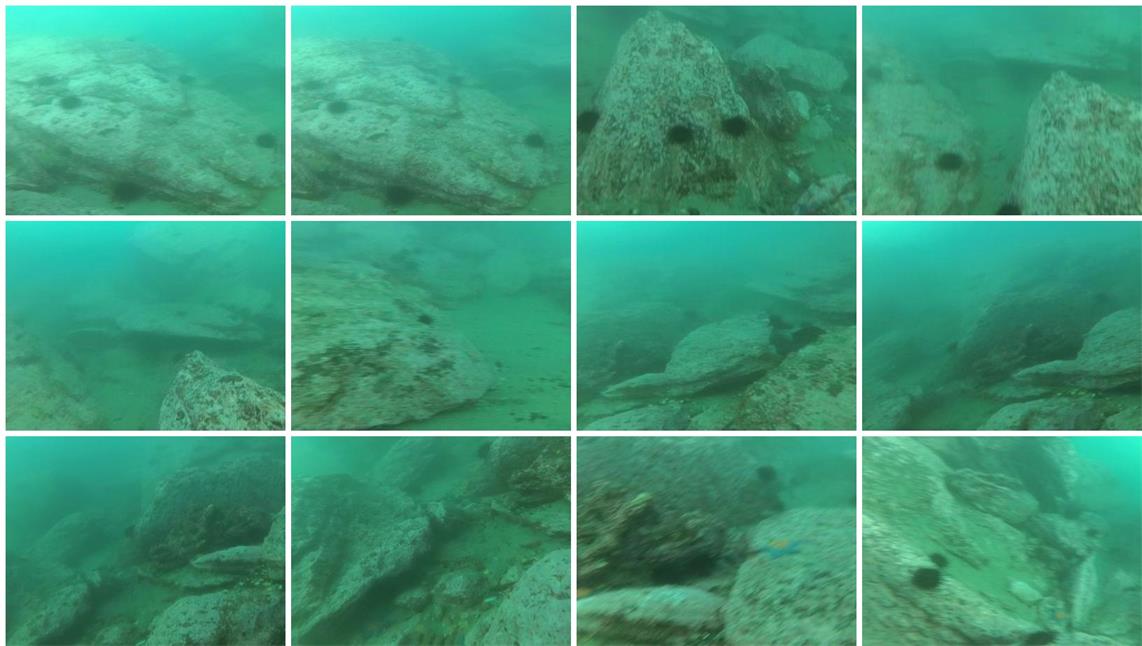}
			&\includegraphics[width=.323\textwidth]{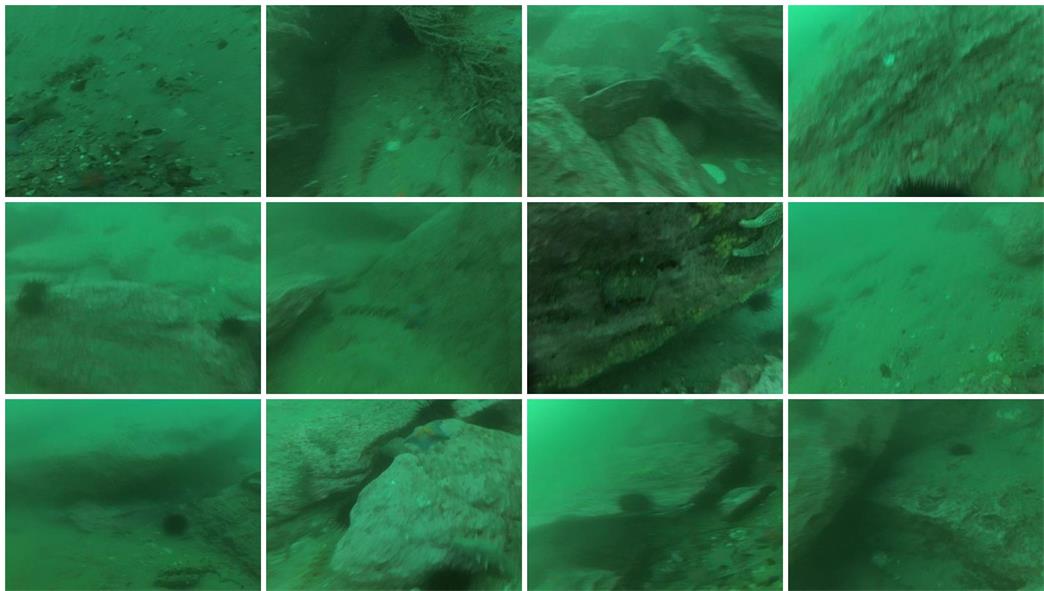}\\
			\multicolumn{3}{c}{\shortstack{(b) Underwater Color Cast Set (UCCS). The set  is divided into ``Green", ``Green-blue", \\and ``Blue" according to the degree of color cast.}\vspace{0.5em}}​
		\end{tabular}
	    
		\begin{tabular}{c@{\extracolsep{0.5em}}c@{\extracolsep{0.5em}}c@{\extracolsep{0.5em}}c@{\extracolsep{0.5em}}c@{\extracolsep{0.5em}}c}
			\includegraphics[width=.19\textwidth]{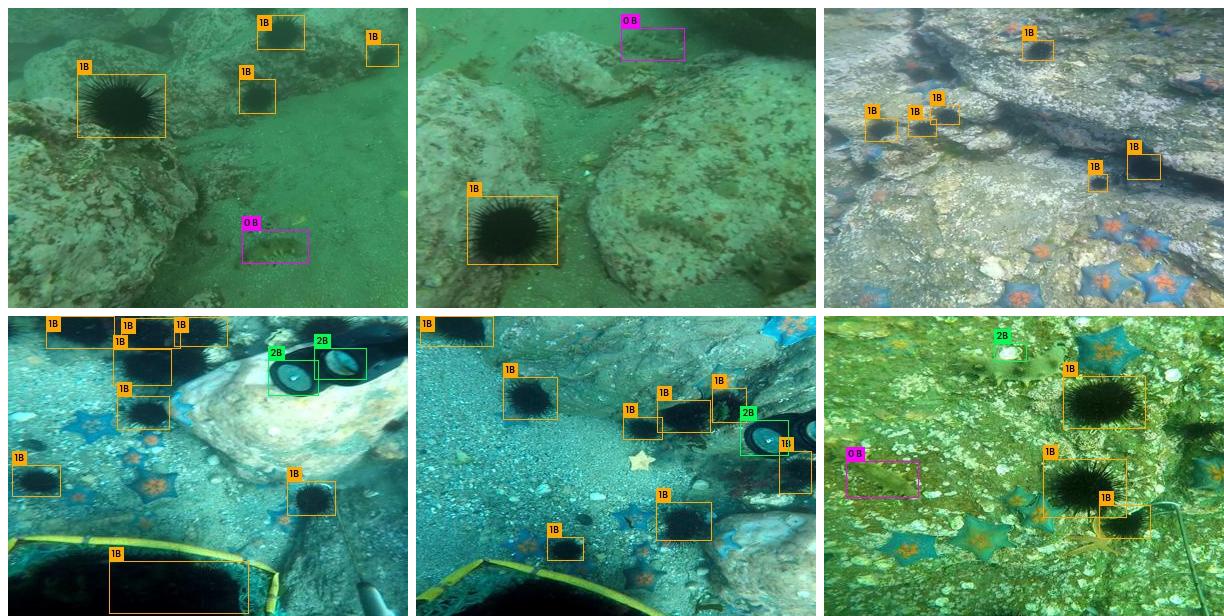}
			&\includegraphics[width=.19\textwidth]{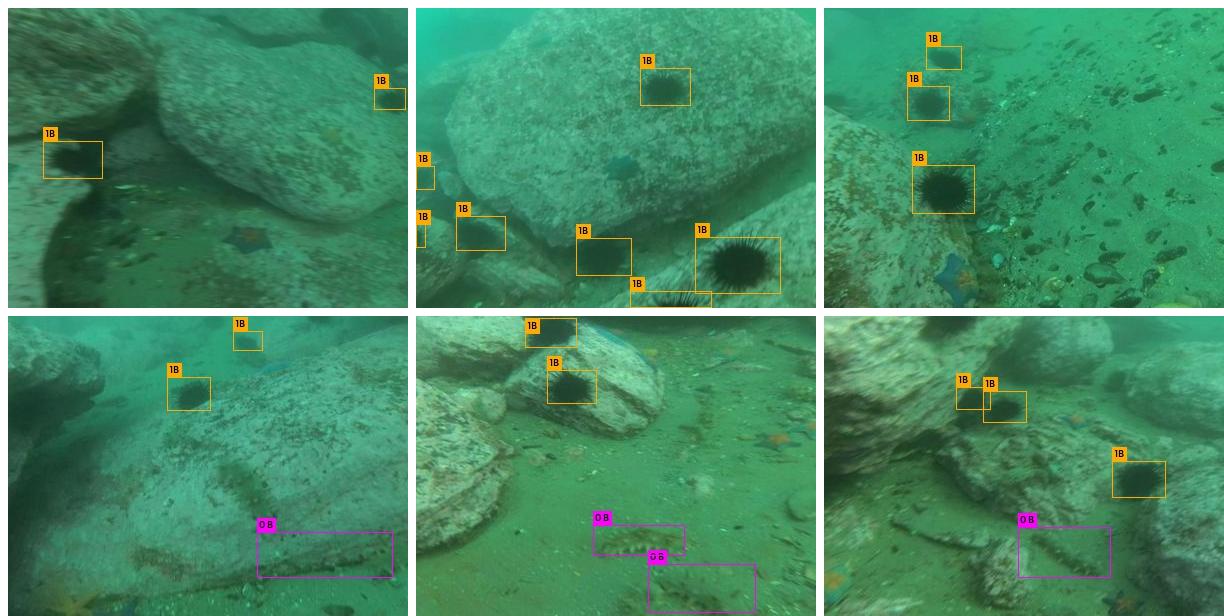}		
			&\includegraphics[width=.19\textwidth]{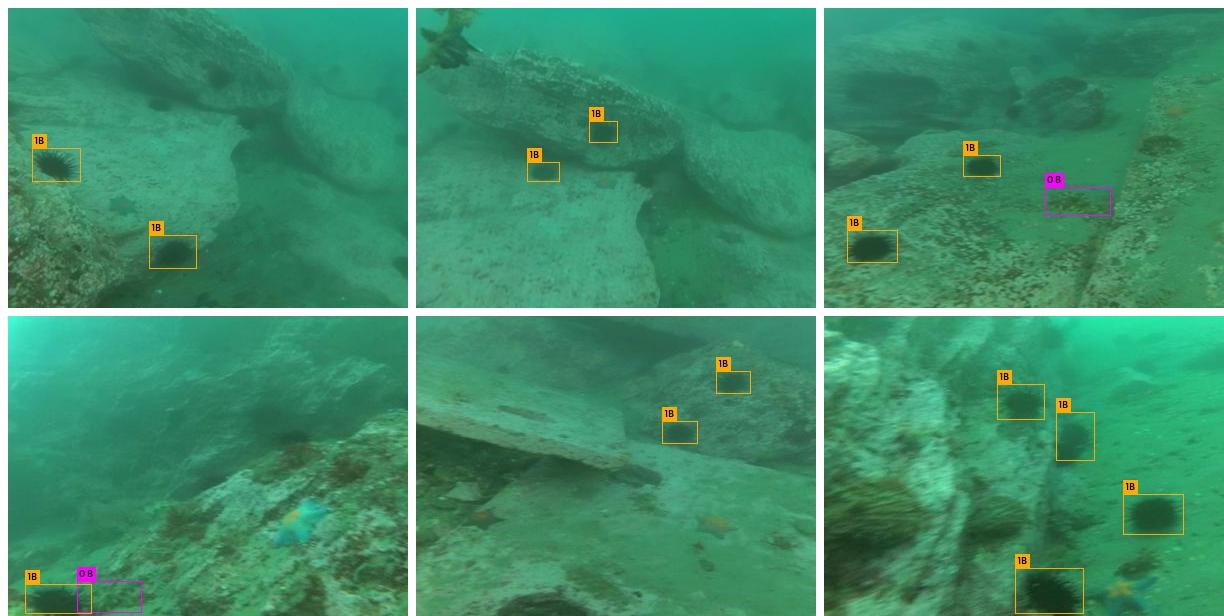}
			&\includegraphics[width=.19\textwidth]{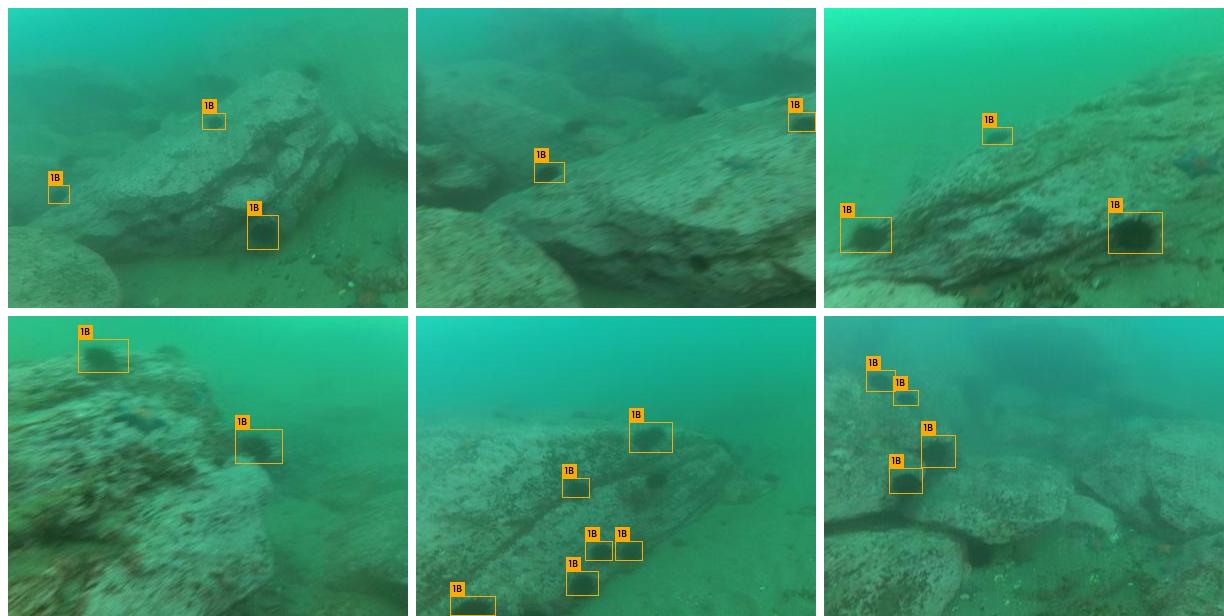}
			&\includegraphics[width=.19\textwidth]{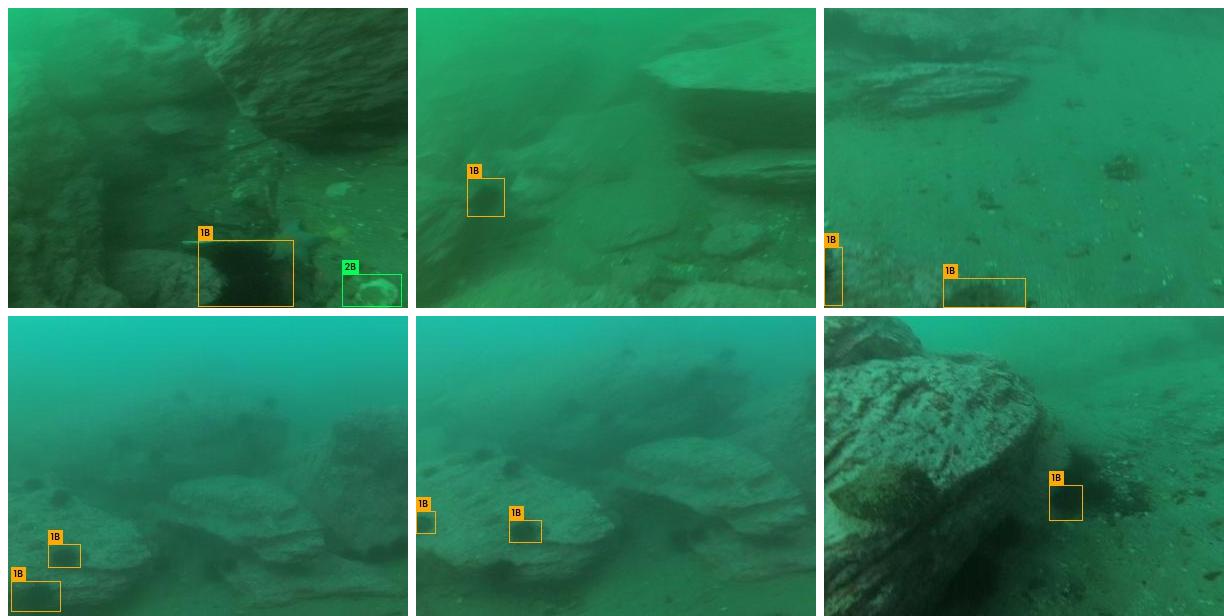}\\
			\multicolumn{5}{c}{\shortstack{(c) Underwater Higher-level Task-driven Set (UHTS). Various sea lif appear in the images of this set. \\The five subsets A$\sim$E are also ranked according to the image quality.}}​
		\end{tabular}
	\end{center}
	\caption{Example images from the three sets of RUIE. 
	}\label{fig:example1}		
\end{figure*}
The model based methods explicitly characterize the physical imaging process, and estimate the parameters of the imaging model from the observation and various prior assumptions. The clear underwater scene can be restored by inverting this degradation process. One common underwater imaging model derives from the Jaffe-McGlamery model as \cite{jaffe1990computer,Mcglamery1980A}:
\begin{equation}
I_c(x)=J_c(x)t_c(x)+A_c(1-t_c(x)),c\in\{r,g,b\},  \label{eq:model}
\end{equation} 
where $I_c(x)$ is the observed degraded image, $J_c(x)$ is the clear scene radiance to be recovered, $c$ represents a color channel. There are two critical parameters for the restoration,~i.e., the global atmospheric light $A_c$, and transmission matrix $t_c(x)$. The transmission denotes the portion of the scene radiance that reaches the camera, defined as:
\begin{equation}
t_c(x)=e^{-\beta_cd(x)},   \label{eq:t}
\end{equation} 
where $d(x)$ represents the scene depth, and $\beta_c$ is the scattering coefficient of the transmission medium depending on water quality,depth and salinity, for underwater images. Most recent underwater enhancement methods estimate these two key parameters $A_c$ and $t_c(x)$ in order to improve visibility, and correct color casts using traditional model-free techniques, e.g., color balance or histogram equalization.

 
Many UIE algorithms attempt to extend the prior model based dehazing algorithms to underwater scenes by noting that the underwater imaging model shares commons with the one for hazy images. The original priors for dehazing have to adapt to the serious attenuation of red light through water so that those dehazing algorithms are applicable to underwater scenarios. For instance, several prior-based UIE methods derive from the dark channel prior (DCP)~\cite{he2009single}, one of the most effective means to estimate the transmission (depth) map of a hazy image. Chiang \emph{et al.} modified DCP by compensating the attenuation to restore the color balance~\cite{chiang2012underwater}. Drews-Jr~\emph{et al.} applied the modified underwater DCP only to the blue and green channels~\cite{Jr2013Transmission}. Galdran~\emph{et al.} proposed the Red Channel Prior (RCP) upon DCP by characterizing the attenuation on the red channel~\cite{Galdran2015Automatic}.

Researchers have also proposed various physical priors designated to underwater images, other than those derived from DCP. Nicholas~\emph{et al.} took advantage of the characteristics of channel discrepancies to estimate the  underwater transmission~\cite{Carlevaris2010Initial}. Wang~\emph{et al.} developed a maximum attenuation identification method (MAI), which only use the red channel information to generate the depth map and atmospheric light~\cite{Wang2017Underwater}. Peng~\emph{et al.} presented a depth estimation method using image blurriness and light absorption~\cite{Peng2017Underwater}. Wang~\emph{et al.} proposed an adaptive attenuation-curve prior applicable to both UIE and dehazing~\cite{Wang2017Single}. 

Nevertheless, one of the common drawbacks of the aforementioned prior-based UIE algorithms lie in that these priors are invalid to some specific environmental/scenery configurations and/or severe color casts. For example, it is well known that DCP is inapplicable to white objects or regions. 

\subsection{Data-driven Enhancement Neural Networks}
In the past decade, the community has witnessed the great success of deep neural networks in many low-level and high-level computer vision tasks. These heuristically constructed networks trained on a large amount of data examples yield superior performance~\cite{lecun2015deep,he2016deep}. In the context of image dehazing, the convolutional neural networks (CNNs) may work in an end-to-end manner, which learns the direct mapping from a degraded image to the corresponding clear scene~\cite{li2017aod, ren2016single, cai2016dehazenet}. Alternatively, the feature representation power of CNNs can also significantly improve the accuracy for depth or transmission estimation in complex scenes and uncontrolled outdoor environments~\cite{zhang2018densely}.

The similarities on imaging models between underwater and hazy images motivate researchers to apply the network architecture similar with dehazing CNNs to data-driven underwater image enhancement. Meanwhile, more complex factors in underwater imaging, including dynamic water flow, color deviations, and low illuminations, require a more complex network structure and/or a well-designed loss function.
Li~\emph{et~al.} designed the WaterGAN to synthesize training examples and an end-to-end network consisting of a depth estimation module followed by a color correction module~\cite{Li2017WaterGAN}. Chen~\emph{et al.} combined a filtering-based with a GAN-based restoration scheme that adopted a multi-stage loss strategy for the training~\cite{Chen2017Towards}. 
Recently, Hou~\emph{et al.} proposed an underwater residual network to jointly optimize transmission and correct color casts~\cite{hou2018joint}.
Different from the above end-to-end UIE networks, Liu~\emph{et al.} established the basic propagation scheme based on the fundamental image modeling cues and then introduce CNNs~\cite{Liu2019learning} or a lightweight residual learning framework~\cite{liu2018learning}, to integrate both physical priors and data driven cues for solving various image enhancement tasks including the underwater aspects.

The three categories of UIE algorithms were evaluated on different data sets in terms of various metrics for visual quality. A comprehensive study is still highly demanded on the extent to which these UIE algorithms achieve the objectives of improving visibility, correcting color casts, and increasing accuracy for the following higher-level vision tasks.

\section{The Proposed Data Set}
\label{sec:data_set}
A successful UIE algorithm has to address one or all of the following issues in underwater imaging including visibility degradation, color casts, and accuracy decrease of higher-level detection tasks. These multiple objectives of UIE algorithms require a diverse portfolio of testing examples in a benchmark for UIE. The images to evaluate the capability of visibility improvements typically need a larger scene depth\footnote{The depth is defined as the distance from the scene to the imaging plane.} so that the degradation effects caused by water scattering are evident. On the other hand, the data set containing a wide range of color tones suffices to evaluate the performance of correcting color casts in underwater imaging. Moreover, the calculation of detection/classification accuracy demands object/target labels as the groundtruth in the benchmark. However,  most existing underwater image data sets generally target at evaluating either one or two of the three objectives for UIE algorithms.
Therefore, the establishment of a large-scale, diverse, and task-specific database has great importance for fair and comprehensive comparisons of UIE algorithms. Additionally, this type of benchmark may lay a `data' foundation for training intelligent underwater vehicles equipped with automatic computer vision systems.

We setup a multi-view underwater image capturing system with twenty-two water-proof video cameras in order to collect image examples for our RUIE benchmark, this system is simplified and modeled as shown in Figure \ref{fig:model} (b). These cameras were mounted along a $10$ meters by $10$ meters square frame, and placed $0.5$ meter above the sea bed close to the Zhangzi island in the Yellow Sea, of which the geographic coordinate locates at $(N39.186,E44.625)$. We carefully adjusted the view angles of the cameras so that the maximum scene depths may vary from $0.5$ to $8$ meters. All videos were captured during two time slots from 8 AM to 11 AM and 1 PM to 4 PM each day between September 21st and 22nd, 2017. The water depth varied from $5$ to $9$ meters owing to the periodic tide. The changing lighting and water depth produce varying color tones. More importantly, this area maintains a natural marine ecosystem containing abundant sea life including fish, sea urchin, sea cucumber, scallops, etc. This ecosystem makes it possible for us to provide labels for underwater object detection tasks.

The captured videos over $250$ hours cover a wide range of diversities on illuminations, depths of fields, blurring degrees, and color casts. We manually picked about four thousand images, and divided them into three subsets according to specific tasks of UIE algorithms. We list their respective profiles and objectives for evaluation as follows.

\textbf{\textit{Underwater Image Quality Set (UIQS):}} This subset is used to test UIE algorithms for the improvement of image visibility. Specifically, we assessed the quality of images according to the underwater color image quality evaluation (UCIQE) metric~\cite{Yang2015An}, and ranked these images by their corresponding UCIQE scores. The UCIQE metric is a linear combination of chroma, saturation, and contrast of underwater images. Then we equally divided them into five subsets, denoted as [A, B, C, D, E], in descending order of the UCIQE values, in order to facilitate testing the performance of different algorithms in various underwater conditions. Figure~\ref{fig:example1} shows image examples with different levels of image quality.

\textbf{\textit{Underwater Color Cast Set (UCCS):}} This set aims to evaluate the ability of correcting color casts for UIE algorithms. According to the average value of the blue ($b$) channel (red-green bias) in the CIElab color space, we collected 300 images from UIQS and produced the UCCS set. It contains three 100-image subsets of the bluish, greenish and blue-green tones. The corresponding example images are shown in the second row of Fig.~\ref{fig:example1}.

\textbf{\textit{Underwater Higher-level Task-driven Set (UHTS):}} The UHTS set contains 300 images containing several types of sea life for the purpose of evaluating the effects of UIE algorithms to higher-level computer vision tasks, e.g., classification and detection. Currently, we label the bounding boxes and types of three classes of sea life, i.e., scallop, sea cucumbers and sea urchins, in these underwater images. The detection and classification of these three types greatly challenge recent computer vision algorithms because their appearance is quite similar to the ambient environment thus difficult to distinguish. The accuracy of higher-level algorithms is sensitive to image quality. Additionally, these sea lives are of great interest for marine ecology. Therefore, we provide these labels for our higher-level task driven set. Furthermore, similar to UIQS, the images UHTS are sorted into five subsets according to the UCIQE scores in order to explore the impact of image quality on the detection accuracy. 

Since enhanced images are often subsequently fed to higher-level computer version tasks, it is noted  that the objective of UIE is not only pixel-level or perceptual-level quality improvements, but also the utility of enhanced images in given semantic analysis tasks. We thus propose the higher-level task-driven evaluation for UIE algorithms, and study the problem of object detection in the presence of visibility degradation as an example. We trained an underwater object detection CNN using the network structure of YOLO-v3 as the baseline~\cite{redmon2018yolov3}. The training set consists of $1,800$ labeled pictures captured from shallow waters with the depth of less than three meters. We apply the trained CNN to detect three types of objects from the enhanced results given by various UIE algorithms. The detection accuracy is evaluated in terms of the mean Average Precision (mAP).

%

\begin{table}[t]
	\centering
	\caption{Subsets of RUIE for training and testing} \label{table:subset}
	\begin{tabular}{ll}
		\toprule
		Subset& Image Number\\\midrule
		Underwater image quality set (UIQS) & 3630 (726\xmark5)\\
		Underwater color cast set (UCCS) & 300 (100\xmark3)\\
		Underwater task-driven testset (UHTS) &300 (60\xmark5)\\\bottomrule
	\end{tabular}
\end{table}

\begin{figure*}[!htbp]
	\begin{center}
		\begin{tabular}{c@{\extracolsep{0.4em}}c@{\extracolsep{0.4em}}c@{\extracolsep{0.4em}}c@{\extracolsep{0.4em}}c@{\extracolsep{0.4em}}c@{\extracolsep{0.4em}}c}
			\includegraphics[width=.155\textwidth]{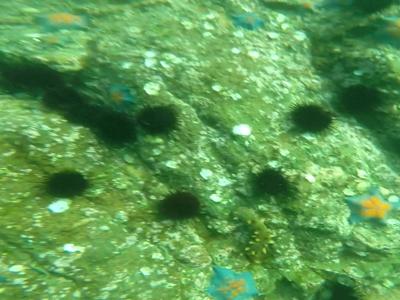}
			&\includegraphics[width=.155\textwidth]{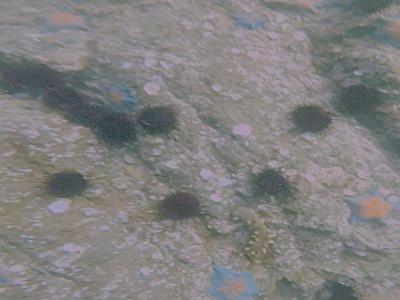}
			&\includegraphics[width=.155\textwidth]{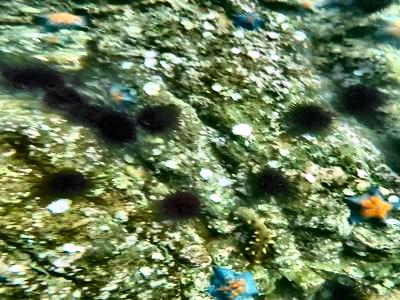}
			&\includegraphics[width=.155\textwidth]{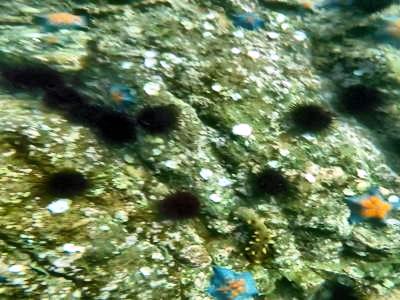}
			&\includegraphics[width=.155\textwidth]{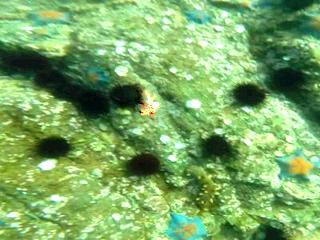}
			&\includegraphics[width=.155\textwidth]{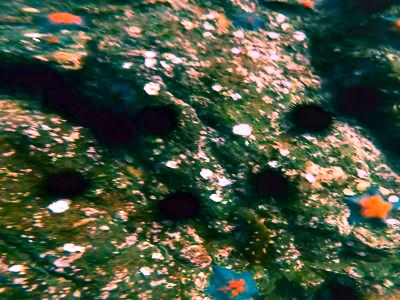}
			\\
			\includegraphics[width=.155\textwidth]{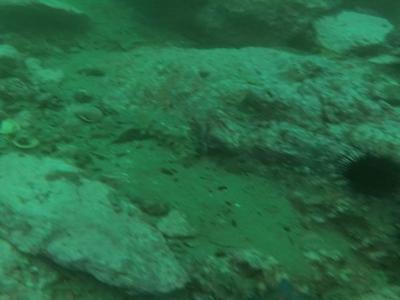}
			&\includegraphics[width=.155\textwidth]{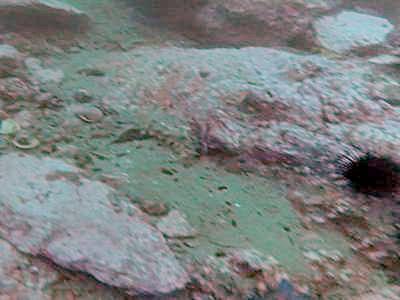}
			&\includegraphics[width=.155\textwidth]{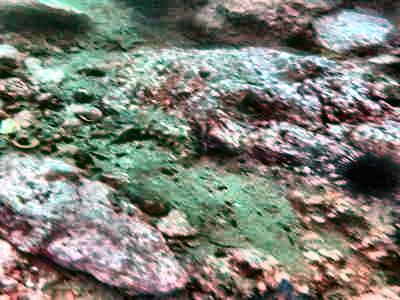}
			&\includegraphics[width=.155\textwidth]{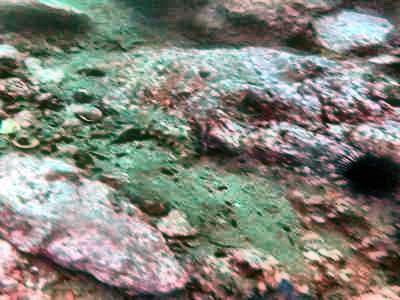}
			&\includegraphics[width=.155\textwidth]{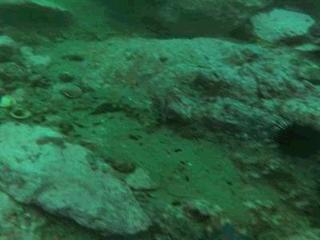}
			&\includegraphics[width=.155\textwidth]{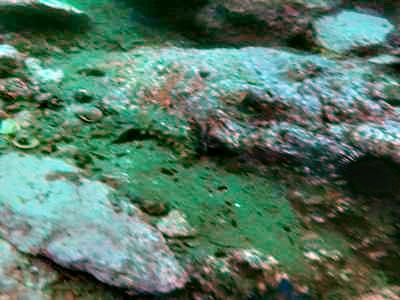}
			\\
			\includegraphics[width=.155\textwidth]{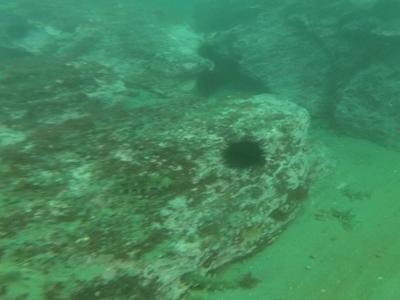}
			&\includegraphics[width=.155\textwidth]{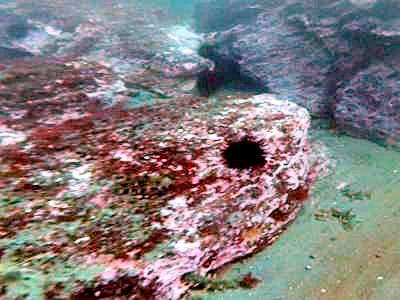}
			&\includegraphics[width=.155\textwidth]{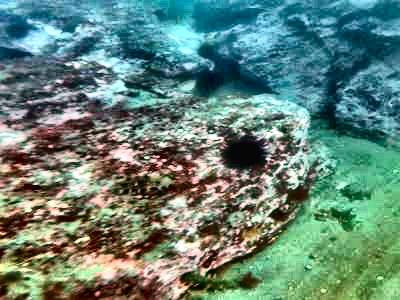}
			&\includegraphics[width=.155\textwidth]{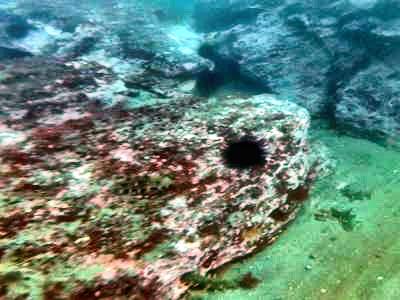}
			&\includegraphics[width=.155\textwidth]{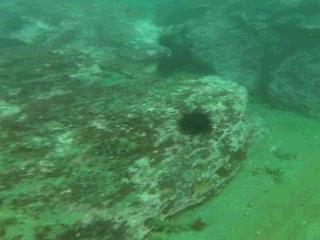}
			&\includegraphics[width=.155\textwidth]{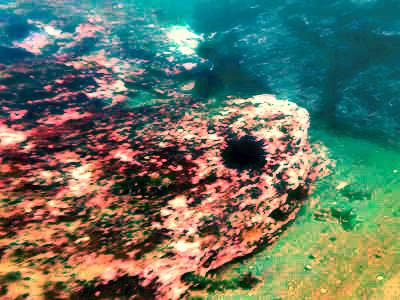}
			\\
			\includegraphics[width=.155\textwidth]{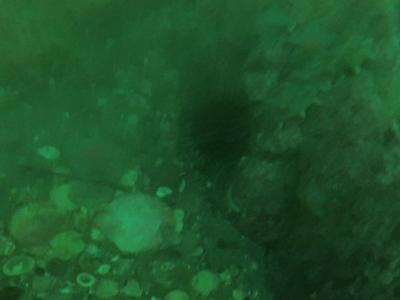}
			&\includegraphics[width=.155\textwidth]{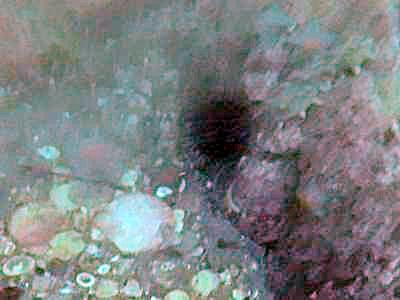}
			&\includegraphics[width=.155\textwidth]{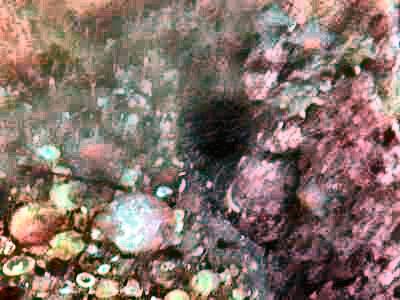}
			&\includegraphics[width=.155\textwidth]{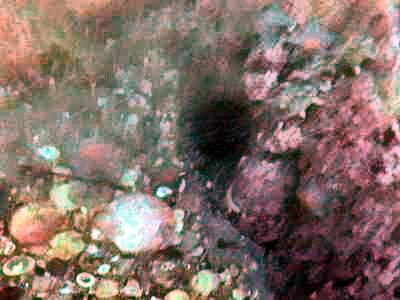}
			&\includegraphics[width=.155\textwidth]{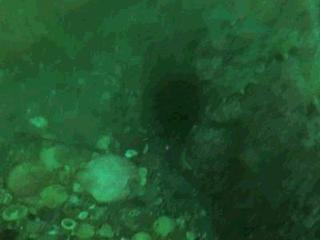}
			&\includegraphics[width=.155\textwidth]{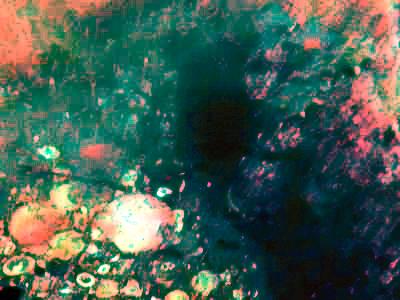}
			\\
			\includegraphics[width=.155\textwidth]{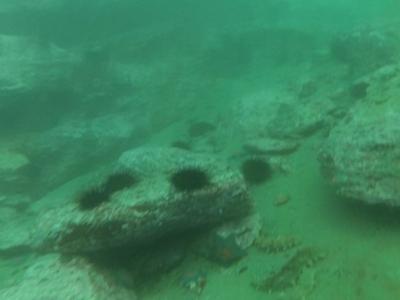}
			&\includegraphics[width=.155\textwidth]{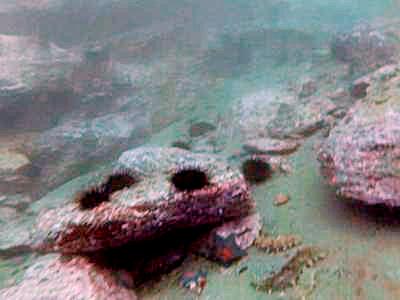}
			&\includegraphics[width=.155\textwidth]{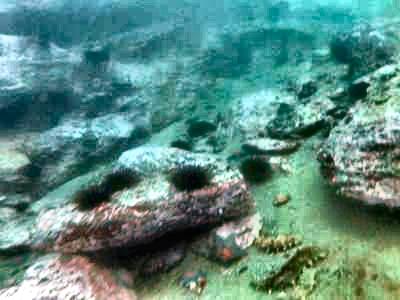}
			&\includegraphics[width=.155\textwidth]{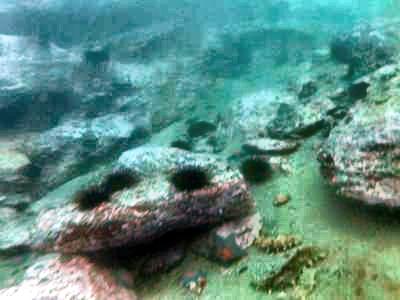}
			&\includegraphics[width=.155\textwidth]{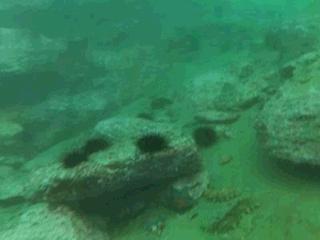}
			&\includegraphics[width=.155\textwidth]{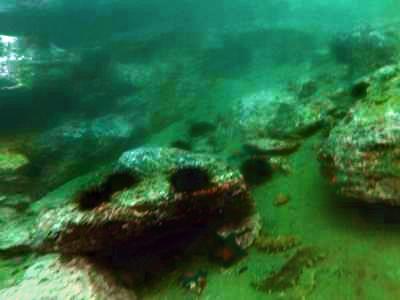}
			\\
			Input &MSRCR &CLAHE &Fusion &BP&UHP \\			
			\includegraphics[width=.155\textwidth]{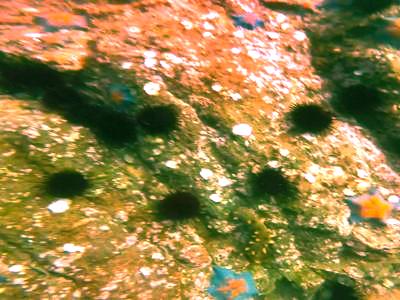}
			&\includegraphics[width=.155\textwidth]{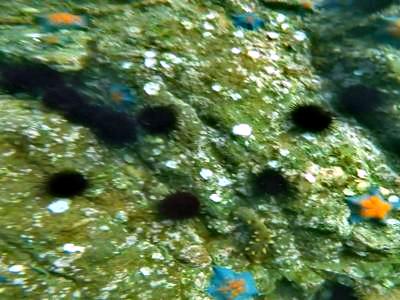}
			&\includegraphics[width=.155\textwidth]{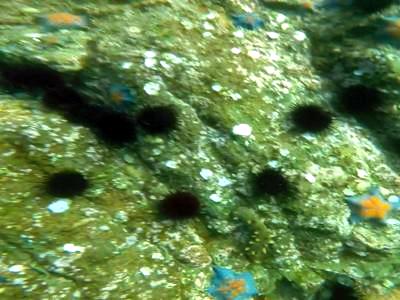}
			&\includegraphics[width=.155\textwidth]{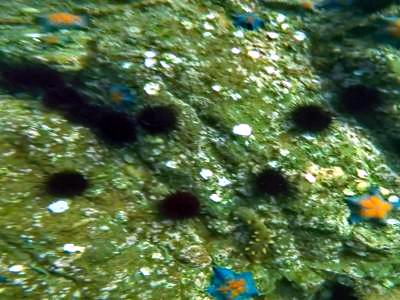}
			&\includegraphics[width=.155\textwidth]{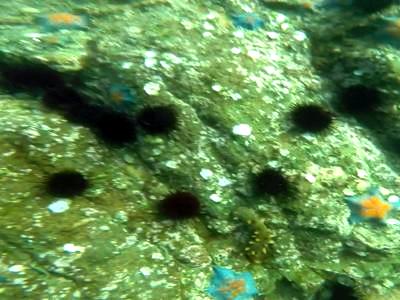}
			&\includegraphics[width=.155\textwidth]{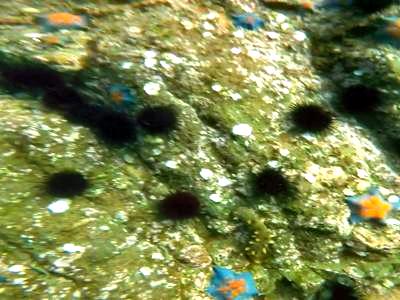}
			\\
			\includegraphics[width=.155\textwidth]{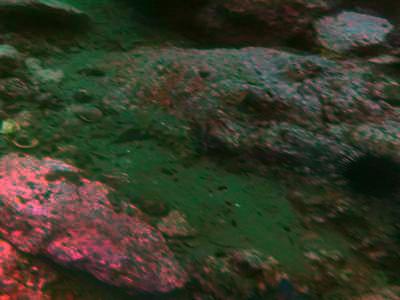}
			&\includegraphics[width=.155\textwidth]{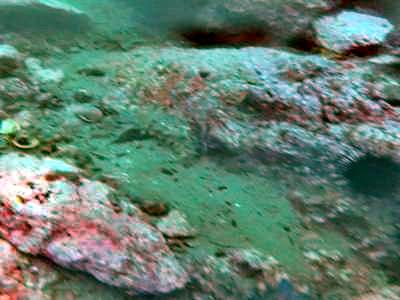}
			&\includegraphics[width=.155\textwidth]{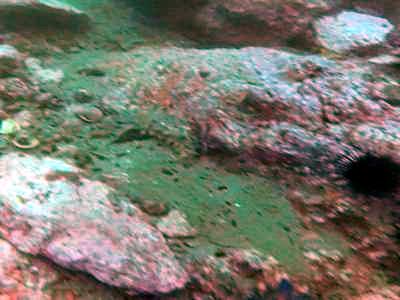}
			&\includegraphics[width=.155\textwidth]{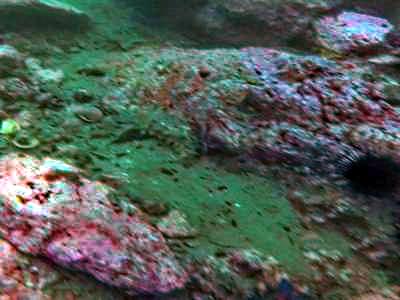}
			&\includegraphics[width=.155\textwidth]{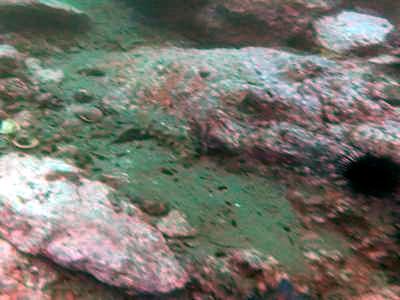}
			&\includegraphics[width=.155\textwidth]{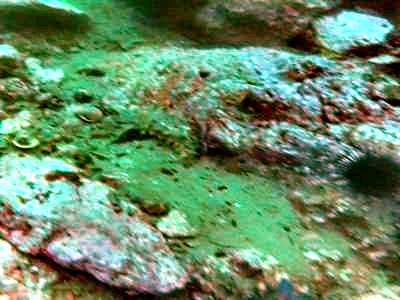}
			\\
			\includegraphics[width=.155\textwidth]{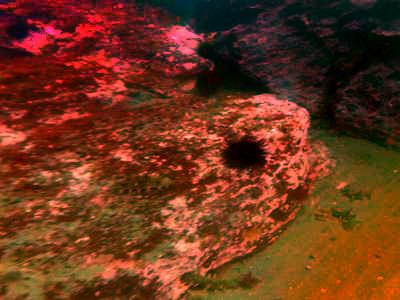}
			&\includegraphics[width=.155\textwidth]{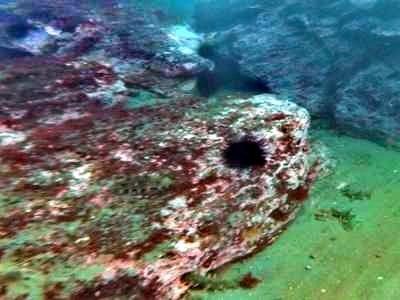}
			&\includegraphics[width=.155\textwidth]{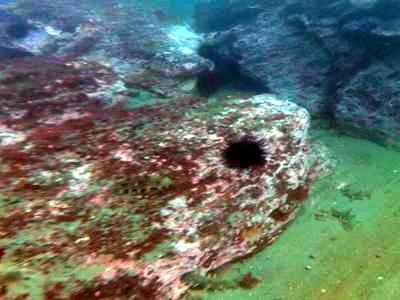}
			&\includegraphics[width=.155\textwidth]{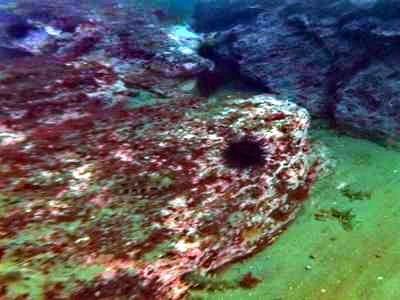}
			&\includegraphics[width=.155\textwidth]{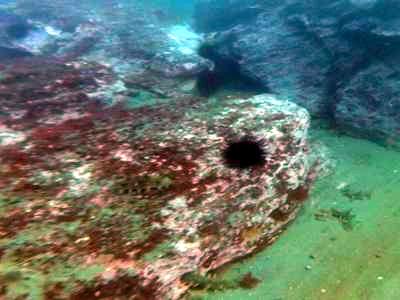}
			&\includegraphics[width=.155\textwidth]{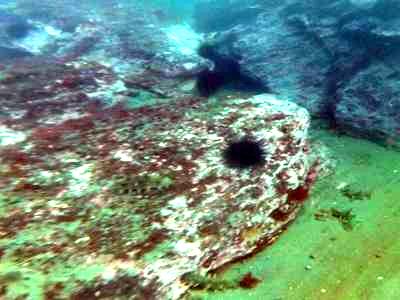}
			\\
			\includegraphics[width=.155\textwidth]{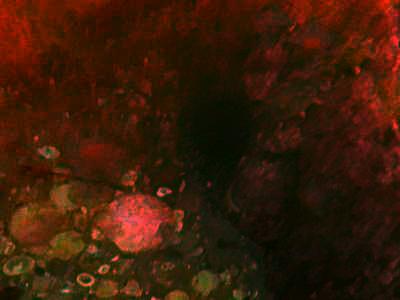}
			&\includegraphics[width=.155\textwidth]{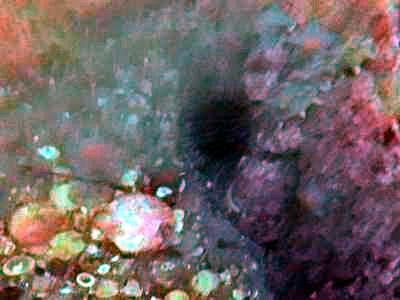}
			&\includegraphics[width=.155\textwidth]{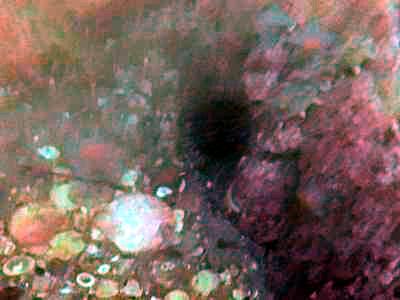}
			&\includegraphics[width=.155\textwidth]{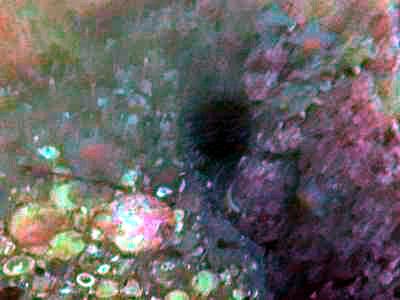}
			&\includegraphics[width=.155\textwidth]{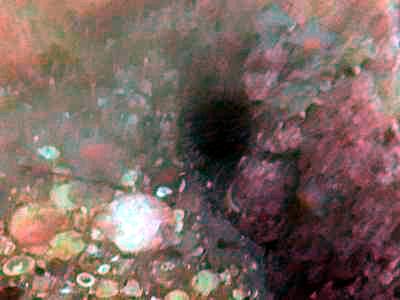}
			&\includegraphics[width=.155\textwidth]{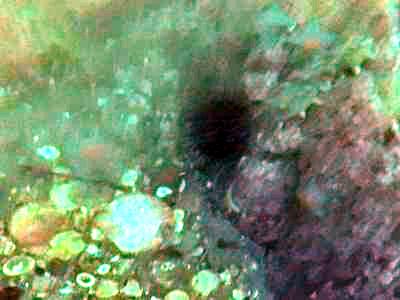}
			\\
			\includegraphics[width=.155\textwidth]{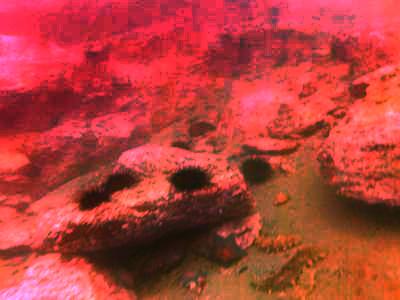}
			&\includegraphics[width=.155\textwidth]{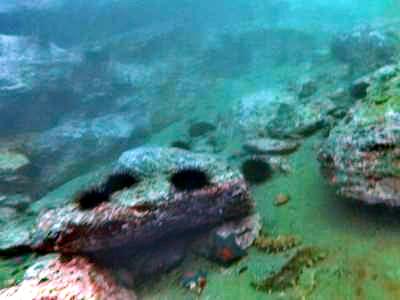}
			&\includegraphics[width=.155\textwidth]{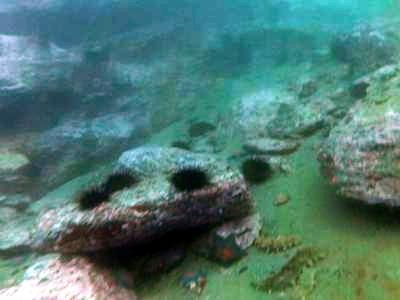}
			&\includegraphics[width=.155\textwidth]{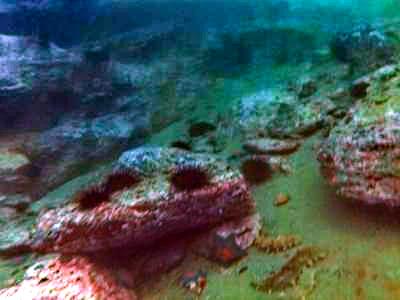}
			&\includegraphics[width=.155\textwidth]{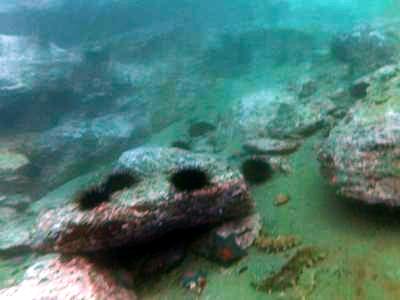}
			&\includegraphics[width=.155\textwidth]{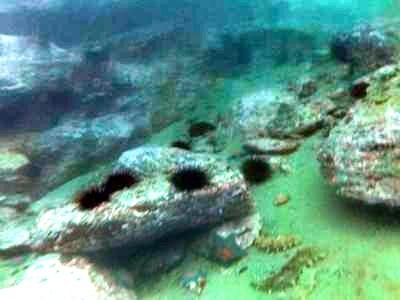}
			\\
			NOM&DPATN & $\text{DCP}_{cb}$ & $\text{BCCR}_{cb}$ &$\text{CAP}_{cb}$ &$\text{HLP}_{cb}$			
		\end{tabular}
	\end{center}
	\caption{Comparison on UIQS sub-dataset. In sequence the quality levels of the five inputs are A , B, C, D and E respectively. 
	}\label{fig:grade_compare}		
\end{figure*}

\section{Evaluation Results and Discussions}
\label{sec:results}
We applied our RUIE benchmark to quantitatively and qualitatively evaluate \emph{eleven} representative underwater image enhancement algorithms \footnote{all images and implementations of compared algorithm are available at https://github.com/dlut-dimt/Underwater-image-enhancement-algorithms}including 
Multi-Scale Retinex with Color Restore (MSRCR)~\cite{Jobson2002A}, Contrast Limited Adaptive Histogram Equalization (CLAHE)~\cite{pizer1990contrast}, Fusion~\cite{Ancuti2012Enhancing}, Bianco Prior~(BP)~\cite{Carlevaris2010Initial}, 
Underwater Haze-line Prior (UHP)~\cite{UnderwaterHL},
New Optical Model (NOM)~\cite{Wen2013Single}, 
Dark Channel Prior ($\text{DCP}_{cb}$)~\cite{he2009single},
Boundary Constrained Context Regularization ($\text{BCCR}_{cb}$)~\cite{meng2013efficient}, 
Color Attenuation Prior ($\text{CAP}_{cb}$)~\cite{zhu2015a},
Haze Line Prior ($\text{CAP}_{cb}$\footnote{The subscript $cb$ indicates that the algorithm was cascaded with color balance as a postprocessing step.})~\cite{berman2016non-local}, 
Data and Prior Aggregated Transmission Network (DPATN)~\cite{liu2018learning}.
Table~\ref{table:method} gives more information about these algorithms. In this section, 
we provide experimental results on the three subsets of our benchmark, and also discuss the results especially on the higher-level task-driven subset. For the sake of evaluation in reasonable time, we resized all images to 300\xmark400, that produces stable outputs for both enhancement and detection, in the following experiments.



%

\subsection{Comprehensive Image Quality Comparisons on UIQS} 

\begin{table*}[t]
	\begin{center}
		\caption{Non-reference Underwater Image Quality Evaluation of algorithms on UIQS.} \label{table:RUI-I}
		\begin{tabular}{p{0.3cm}<{\centering}p{0.92cm}<{\centering}p{0.92cm}<{\centering}p{0.92cm}<{\centering}p{0.92cm}<{\centering}p{0.92cm}<{\centering}p{0.92cm}<{\centering}p{0.92cm}<{\centering}p{0.92cm}<{\centering}p{0.92cm}<{\centering}p{0.92cm}<{\centering}p{0.92cm}<{\centering}p{0.92cm}<{\centering}p{0.92cm}<{\centering}}\toprule
			&Metric&Input &MSRCR &CLAHE &Fusion &BP& UHP &NOM&DPATN & $\text{DCP}_{cb}$ & $\text{BCCR}_{cb}$ &$\text{CAP}_{cb}$ &$\text{HLP}_{cb}$\\ \midrule
			&UICM& -74.29&  \textbf{-2.07}& -20.80& -21.75& -69.04& -22.82& \textcolor{green}{42.45}&-16.18& -10.11&  \textbf{ 0.057}& -14.12& -31.96\\
			&UISM&   1.657&   4.925&   \textbf{5.532}&   4.784&   0.162&   3.251&   2.317&  \textbf{4.947}& 3.042&   3.805&   3.098&   3.650\\
			E&UIConM&   0.459&   0.712&   0.807&   \textbf{0.818}&   0.563&   0.743&   0.684& 0.811&  0.778&  \textbf{ 0.833}&   0.759&   0.782\\
			&UIQM&   0.035&   \textbf{3.942}&   3.933&   3.725&   0.114&   2.974&   \textcolor{green}{4.327}&  3.902& 3.396&   \textbf{4.103}&   3.232&   2.971\\
			&UCIQE&   0.240&   0.493&   0.451&   0.469&   0.264&   \textbf{0.500}&   0.486&  0.479& 0.486&   \textbf{0.501}&   0.475&   0.462\\
			\midrule
			&UICM& -77.51&  \textbf{-0.718}& -16.32& -16.72& -69.59& -19.40& \textcolor{green}{35.48}& -15.24& -2.001&   \textbf{5.208}&  -5.765& -28.98\\
			&UISM&   1.825&   4.852&   \textbf{5.598}&   4.845&   0.257&   3.369&   2.298&  \textbf{4.857}& 3.136&   3.805&   3.193&   3.798\\
			D&UIConM&   0.512&   0.717&   0.805&   \textbf{0.825}&   0.619&   0.745&   0.665& 0.806&  0.781&   \textbf{0.822}&   0.759&   0.784\\
			&UIQM&   0.184&   3.975&   \textbf{4.071}&   3.909&   0.328&   3.112&   \textcolor{green}{4.058}&  3.886& 3.661&   \textbf{4.210}&   3.493&   3.108\\
			&UCIQE&   0.266&   0.483&   0.452&   0.470&   0.293&   \textbf{0.500}&   0.477& 0.475&  0.491&   \textbf{0.505}&   0.476&   0.472\\
			\midrule
			&UICM& -82.01&  \textbf{-1.497}& -19.50& -19.19& -70.57& -18.59& \textcolor{green}{37.84}& -14.83& -1.85&   \textbf{6.256}&  -7.920& -31.69\\
			&UISM&   1.836&   4.663&   \textbf{5.525}&   4.719&   0.262&   3.376&   2.251& \textbf{4.793}&  3.094&   3.793&   3.180&   3.731\\
			C&UIConM&   0.520&   0.703&   0.791&   \textbf{0.811}&   0.635&   0.742&   0.652& 0.802&  0.766&   \textbf{0.809}&   0.740&   0.758\\
			&UIQM&   0.089&   3.848&   \textbf{3.909}&   3.753&   0.358&   3.124&   \textcolor{green}{4.061}& 3.863&  3.600&   \textbf{4.190}&   3.363&   2.918\\
			&UCIQE&   0.283&   0.475&   0.453&   0.470&   0.311&   \textbf{0.501}&   0.478&  0.478& 0.493&   \textbf{0.502}&   0.472&   0.478\\
			\midrule
			&UICM& -84.82&  -2.079& -16.27& -17.00& -71.70& -18.49& \textcolor{green}{29.53}& -16.97&  \textbf{1.695}&   \textbf{8.351}& -10.65& -28.55\\
			&UISM&   1.889&   4.431&   \textbf{5.422}&  4.548&   0.351&   3.235&   2.289&  \textbf{4.814}& 3.076&   3.671&   3.131&   3.608\\
			B&UIConM&   0.539&   0.688&   0.778&  \textbf{ 0.799}&   0.637&   0.727&   0.639&  \textbf{0.799}& 0.754&   \textbf{0.787}&   0.723&   0.725\\
			&UIQM&   0.092&   3.709&   \textbf{3.925}&   3.719&   0.361&   3.034&   \textcolor{green}{3.794}&   3.651&   \textbf{4.134}& 3.800&  3.210&   2.852\\
			&UCIQE&   0.301&   0.463&   0.455&   0.471&   0.338&   \textbf{0.508}&   0.466& 0.477&  0.496&   \textbf{0.500}&   0.472&   0.486\\
			\midrule
			&UICM& -77.19&   \textbf{0.971}& -23.68& -26.65& -64.64& -22.23&  \textcolor{green}{11.58}& -17.79&-16.30&  \textbf{-7.198}& -32.91& -36.94\\
			&UISM&   3.291&   3.650&   \textbf{6.085}&   \textbf{5.057}&   1.131&   4.153&   3.513& 3.870&  4.486&   4.805&   4.252&   4.729\\
			A&UIConM&   0.696&   0.630&   0.785&   \textbf{0.841}&   0.715&   0.755&   0.721&  0.790& 0.804&   \textbf{0.815}&   0.787&   0.738\\
			&UIQM&   1.285&   3.358&   \textbf{3.937}&   3.749&   1.068&   3.300&   \textcolor{green}{3.940}&  3.467& 3.738&   \textbf{4.129}&   3.142&   2.991\\
			&UCIQE&   0.362&   0.375&   0.430&   0.449&   0.413&   \textbf{0.504}&   0.457& \textbf{0.489}&  0.484&   0.486&   0.462&   0.485\\
			\bottomrule
		\end{tabular}
	\end{center}
\end{table*}
\textbf{Qualitative comparisons:}
We compared the capabilities of the eleven methods to improve the image visibility on the subset UIQS. The qualitative comparison in Fig.~\ref{fig:grade_compare} demonstrate that most methods are able to achieve better enhancement for images with the quality levels A,B and C where the underwater scattering effect is subtle. The results of MSRCR appear appropriate color tones but not enough saturation and contrast. CLAHE and Fusion notably improve the image brightness, saturation and contrast, but the lack of an imaging model leads to evident hazy effects. Additionally, these three methods are unable to adaptively work in various scenarios due to their fixed parameter settings.
BP is effective for removing the effects caused by light scattering, but cannot deal with color cast well, especially when the water is greenish. UHP generates over-saturation and excessive contrast, smearing image details. DPATN as well as those algorithms with dehazing priors extended to underwater imaging can effectively remove haze-like effects and produce more natural scene. 

For the underwater images with the quality grades of D and E, the algorithm of MSRCR with a fixed set of parameters even aggravates the scattering effect. CLAHE and Fusion improve the contrast of these categories of images, but introduce considerable artifacts and keep the severe haze-like effects. BP works little on improving these severely degraded inputs. Another prior-based method UHP can yield relatively clearer results, especially for scenes farther from the camera. NOM tends to make the results severely reddish. It is worth noting that these reddish results still exhibit high UICM scores because the UICM measure favors the reddish hue. In contrast, DPATN, $\text{HLP}_{cb}$ and $\text{BCCR}_{cb}$ are able to remove the haze-like effects well in these challenging images. Among the three, $\text{HLP}_{cb}$ improves the image brightness recovering more image details, and $\text{BCCR}_{cb}$ performs the best on improving visibility and contrast.

\textbf{Quantitative comparisons:}
We employ two \emph{non-reference} metrics for the quantitative assessment of underwater image quality as no ground truth scene is available as the reference for real world sea images. One is the underwater image quality measure (UIQM) \cite{Panetta2016Human}, consisting of three underwater image attribute measures, i.e., the underwater image colourfulness measure (UICM), underwater image sharpness measure (UISM), and underwater image contrast measure (UIConM). UIQM is expressed as a linear combination of these three components as:
\begin{equation}
\text{UIQM}=c_1\times \text{UICM} +c_2\times \text{UISM}+c_3\times\text{UIConM}, 
\end{equation}
where $c_1$, $c_2$ and $c_3$ are the scale factors. We set $c_1=0.0282$, $c_2=0.2953$, and $c_3=3.5753$ as the original paper \cite{Panetta2016Human}.
The other one is the underwater color image quality evaluation (UCIQE) \cite{Yang2015An}, which uses a linear combination of the chroma, saturation, and contrast of underwater images in the CIElab color space.
The UCIQE score can be obtained as:
\begin{equation}
\text{UCIQE}=c_1\times \omega_c +c_2\times con_l+c_3\times\mu_s, 
\end{equation}
where $\omega_c$ is the standard deviation of chroma; $con_l$ is the contrast of brightness; $\mu_s$ is the average of saturation; $c_1$ and $c_2$ are the scale factors. Again, we set $c_1=0.4680$, $c_2=0.2745$, and  $c_3=0.2576$ as the original paper~\cite{Yang2015An}.

Table~\ref{table:RUI-I} gives the quantitative scores of the eleven UIE algorithms averaged on all images of the UIQS data set. In addition to the two comprehensive quality metrics, UICQM and UCIQE, we also list the values of the components of UICQM including UICM, UISM and UIConM, reflecting colourfulness, sharpness, and contrast of an image, respectively. The two highest two values are marked bold. It should be noted that, the excessive redness in  results of NOM boosts extremely high UCIM and UCIQM scores. Unfortunately, these results are so visually poor that they were considered as outliers and marked in green. We can see from Tab.~\ref{table:RUI-I} that the following algorithms perform over the others in terms of one single metric: MSRCR and $\text{BCCR}_{cb}$ have advantages in balancing color casts; CLAHE and prior-CNN-aggregated DPATN can generate sharper images; Fusion, DPATN and $\text{BCCR}_{cb}$ output images with higher contrasts. In terms of the two comprehensive metrics, DPATN, $\text{DCP}_{cb}$,  $\text{BCCR}_{cb}$, and the three model-free methods (MSRCR, CLAHE, and Fusion) stably improve image quality on all the five categories, and the gaps over the others are more evident on the categories of C, D and E with lower image quality. Among them, CLAHE and $\text{BCCR}_{cb}$ obtain higher UIQM scores, while UHP and $\text{BCCR}_{cb}$ yield higher UCIQE scores. 

As a summary, the prior-based BP algorithm is more suitable to process images with less degradation, while those model-free based methods including Fusion and CLAHE, and prior-aggregated DPATN are the better choices for severely degraded images.


\textbf{Discussion: Underwater image quality assessment}

There exist discrepancies between the qualitative images in Fig.~\ref{fig:grade_compare} and the quantitative scores in Table~\ref{table:RUI-I}. The algorithms producing results with higher scores do not always exhibit favorable appearance for human visual perception. Additionally, UCIQE and UIQM may yield inconsistent assessments on images. For example, NOM always tends to produce severe reddish color shift due to excessive enhancement, resulting in higher UIQM scores, especially on images of lower visual quality. The metric UCIQE favors the results with high contrasts, even for those of UHP showing unnatural excessive contrast. One possible explanation lies in that both UCIQE and UIQM metrics focus on the intensities of low-level features, e.g., contrast and saturation, but ignore higher semantic or prior knowledge from human perception. Also, the calculation of these metrics fails to test whether the intensity values fall within a reasonable range over the whole image. Therefore, the development of an appropriate and objective metric for underwater image quality assessments is still an open issue in this field.

Recently, data-driven CNNs comprising the information from human labels for non-reference quality assessment of natural images have rapidly developed and achieved remarkable performance~\cite{Le2014Convolutional,Bianco2016On,pan2018blind}. It is a promising direction to investigate how to immigrate the deep architectures for natural images to underwater scenarios. Training examples also play an important role for any deep learning approaches. From this respect, the real-world images showing different quality levels in our RUIE data set may contribute to this type of studies.



\subsection{Color Correction Comparisons on UCCS}
\begin{figure*}[t]
	\begin{center}
		\begin{tabular}{c@{\extracolsep{0.4em}}c@{\extracolsep{0.4em}}c@{\extracolsep{0.4em}}c@{\extracolsep{0.4em}}c@{\extracolsep{0.4em}}c@{\extracolsep{0.4em}}c}
			\includegraphics[width=.155\textwidth]{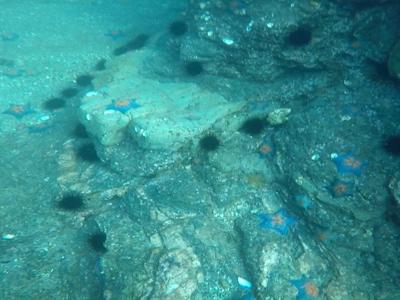}
			&\includegraphics[width=.155\textwidth]{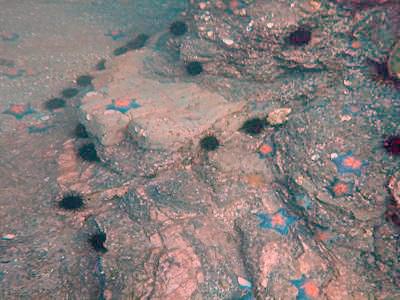}
			&\includegraphics[width=.155\textwidth]{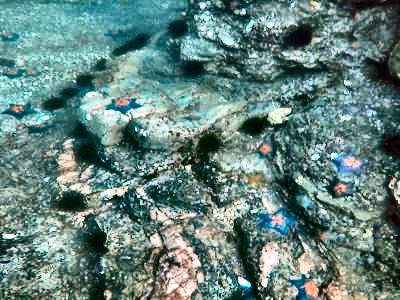}
			&\includegraphics[width=.155\textwidth]{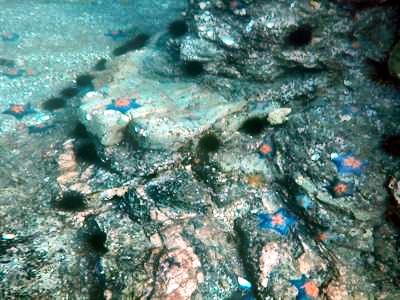}
			&\includegraphics[width=.155\textwidth]{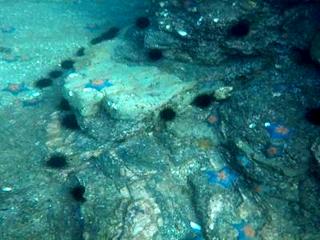}
			&\includegraphics[width=.155\textwidth]{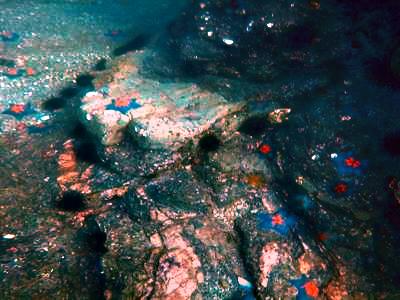}
			\\
			\includegraphics[width=.155\textwidth]{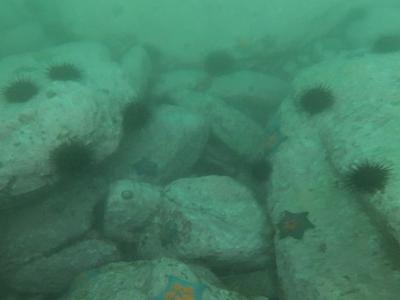}
			&\includegraphics[width=.155\textwidth]{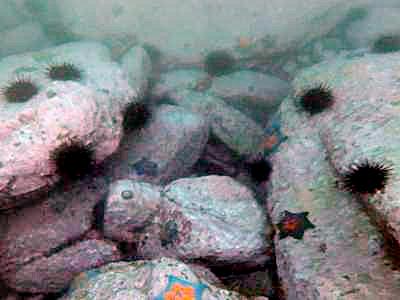}
			&\includegraphics[width=.155\textwidth]{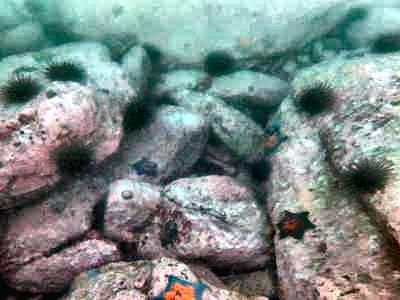}
			&\includegraphics[width=.155\textwidth]{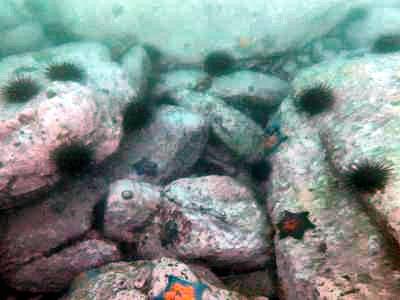}
			&\includegraphics[width=.155\textwidth]{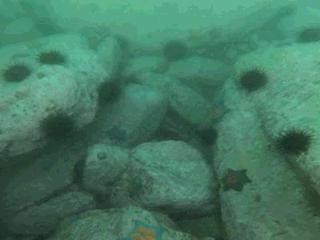}
			&\includegraphics[width=.155\textwidth]{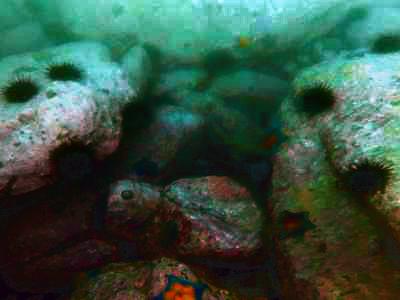}
			\\
			\includegraphics[width=.155\textwidth]{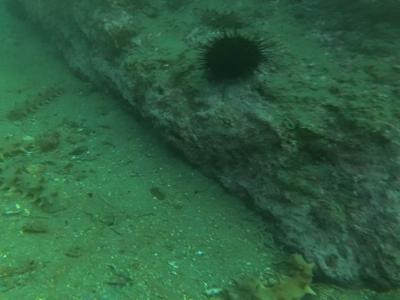}
			&\includegraphics[width=.155\textwidth]{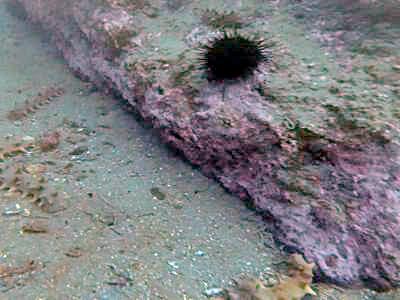}
			&\includegraphics[width=.155\textwidth]{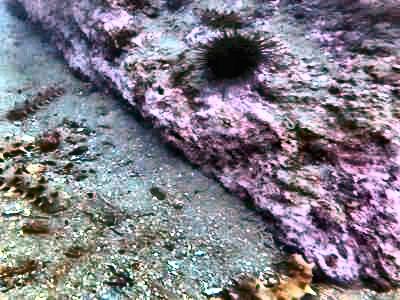}
			&\includegraphics[width=.155\textwidth]{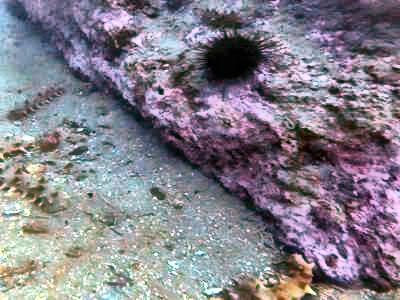}
			&\includegraphics[width=.155\textwidth]{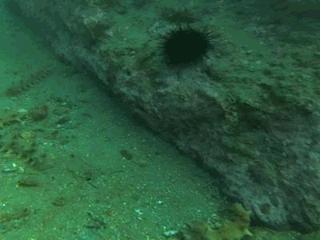}
			&\includegraphics[width=.155\textwidth]{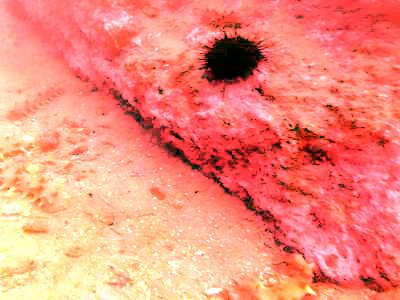}
			\\
			Input &MSRCR &CLAHE &Fusion &BP&UHP \\			
			\includegraphics[width=.155\textwidth]{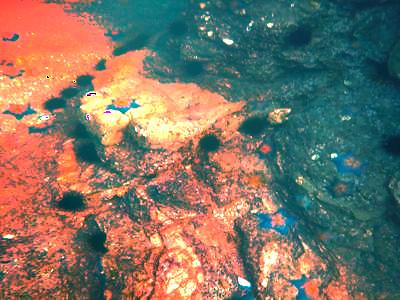}	
			&\includegraphics[width=.155\textwidth]{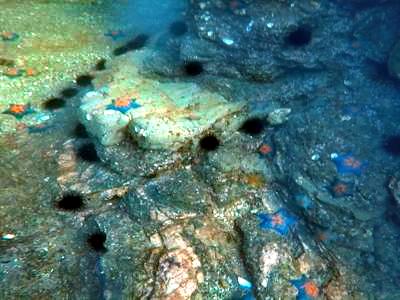}	
			&\includegraphics[width=.155\textwidth]{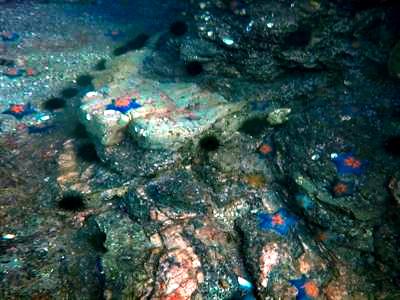}
			&\includegraphics[width=.155\textwidth]{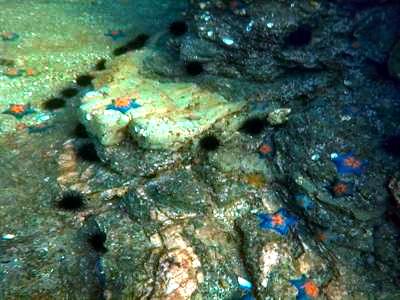}
			&\includegraphics[width=.155\textwidth]{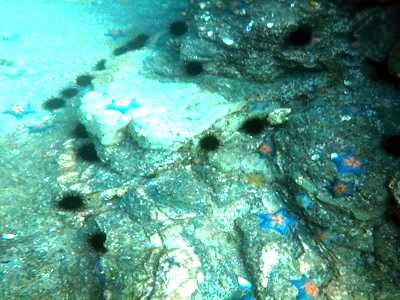}
			&\includegraphics[width=.155\textwidth]{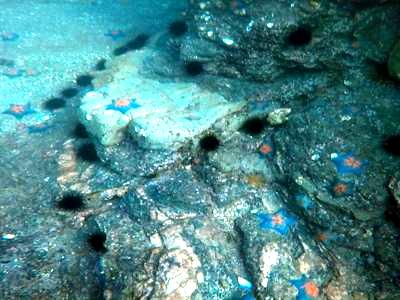}
			\\
			\includegraphics[width=.155\textwidth]{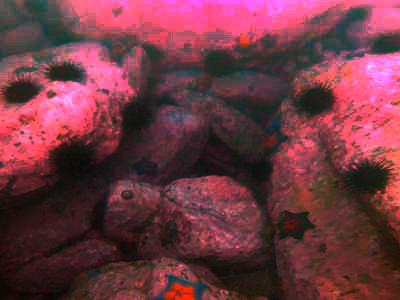}	
			&\includegraphics[width=.155\textwidth]{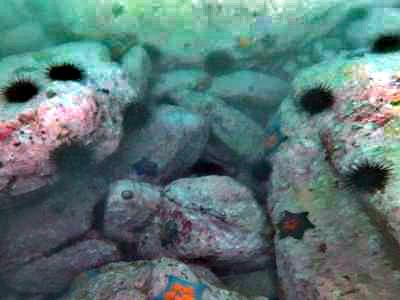}	
			&\includegraphics[width=.155\textwidth]{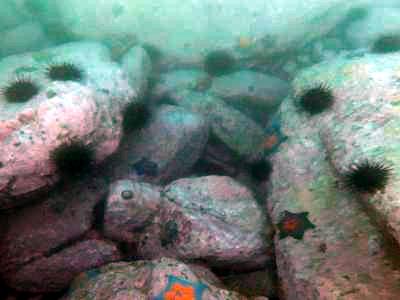}
			&\includegraphics[width=.155\textwidth]{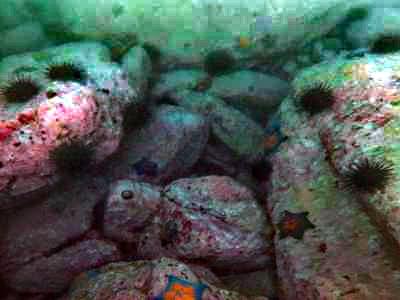}
			&\includegraphics[width=.155\textwidth]{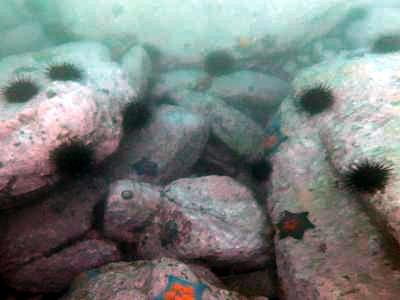}
			&\includegraphics[width=.155\textwidth]{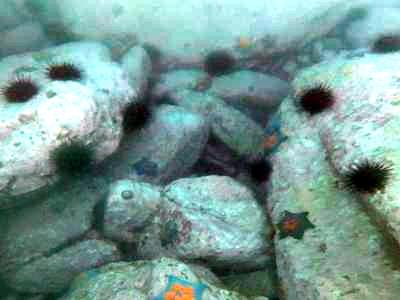}
			\\		
			\includegraphics[width=.155\textwidth]{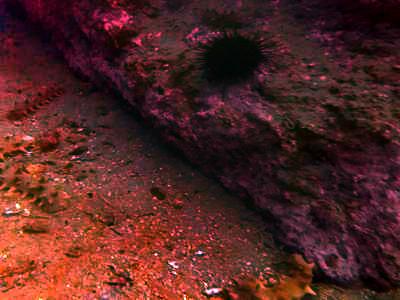}
			&\includegraphics[width=.155\textwidth]{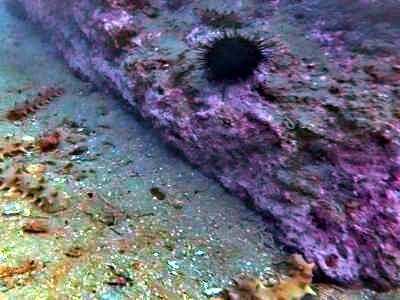}	
			&\includegraphics[width=.155\textwidth]{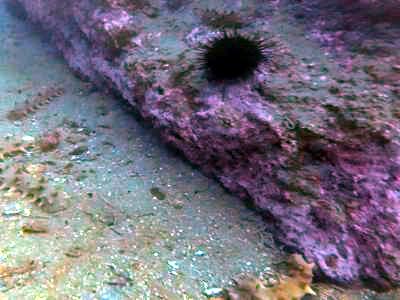}
			&\includegraphics[width=.155\textwidth]{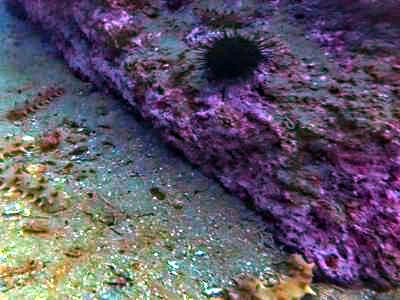}
			&\includegraphics[width=.155\textwidth]{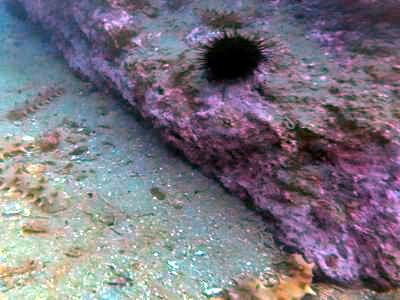}
			&\includegraphics[width=.155\textwidth]{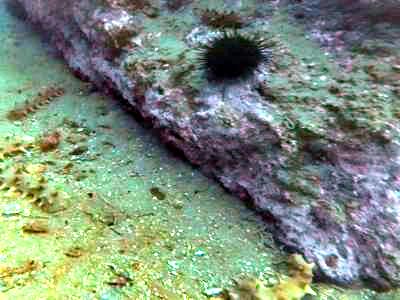}
			\\
			NOM &DPATN& $\text{DCP}_{cb}$ & $\text{BCCR}_{cb}$ &$\text{CAP}_{cb}$ &$\text{HLP}_{cb}$			
		\end{tabular}
	\end{center}
	\caption{The comparison on the dataset UCCS. The three input images from top to bottom are from subsets ``Blue'', ``Green-blue'', and ``Green'' of UCCS, respectively.}\label{fig:tone_compare}		
\end{figure*}

\begin{figure*}[!htbp]
	\begin{center}
		\begin{tabular}{c@{\extracolsep{0em}}c@{\extracolsep{0em}}c@{\extracolsep{0em}}c@{\extracolsep{0em}}c@{\extracolsep{0em}}c@{\extracolsep{0em}}c}
			\includegraphics[width=.2\textwidth]{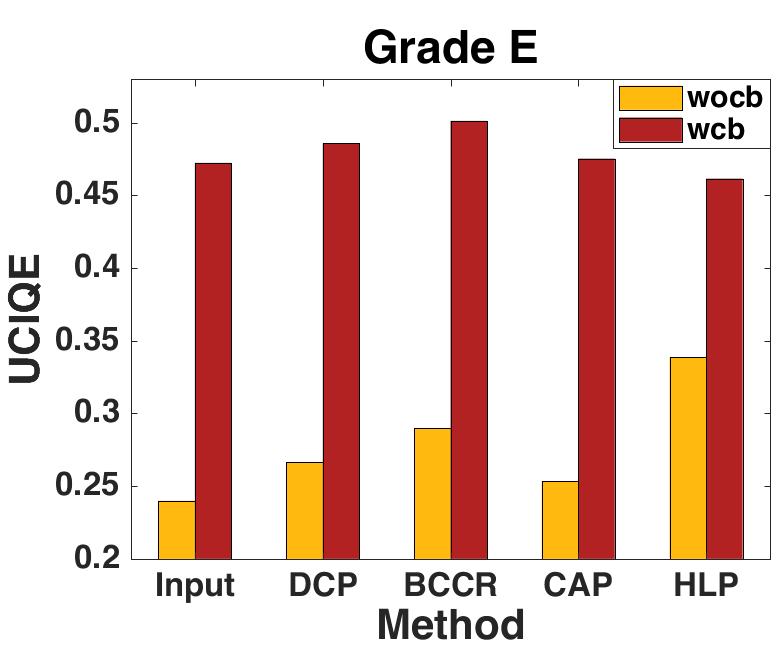}
			&\includegraphics[width=.2\textwidth]{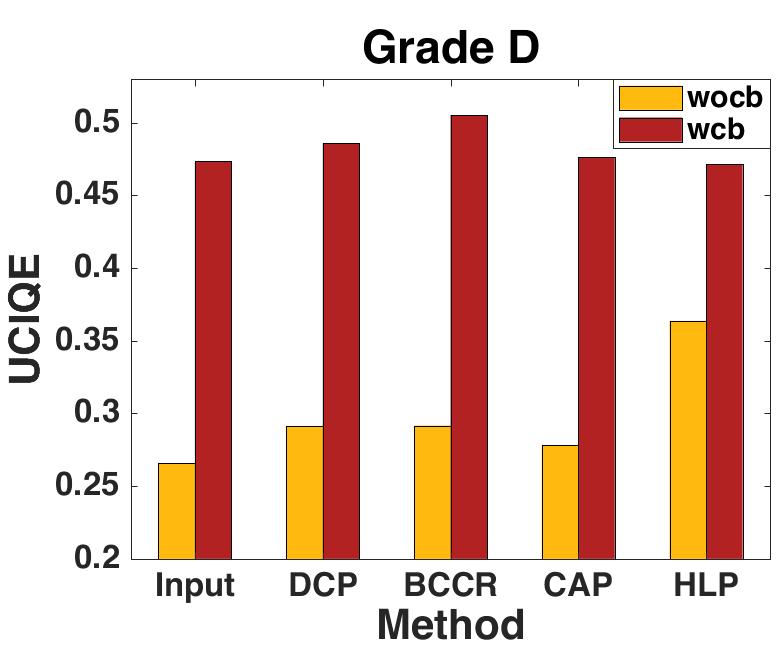}	
			&\includegraphics[width=.2\textwidth]{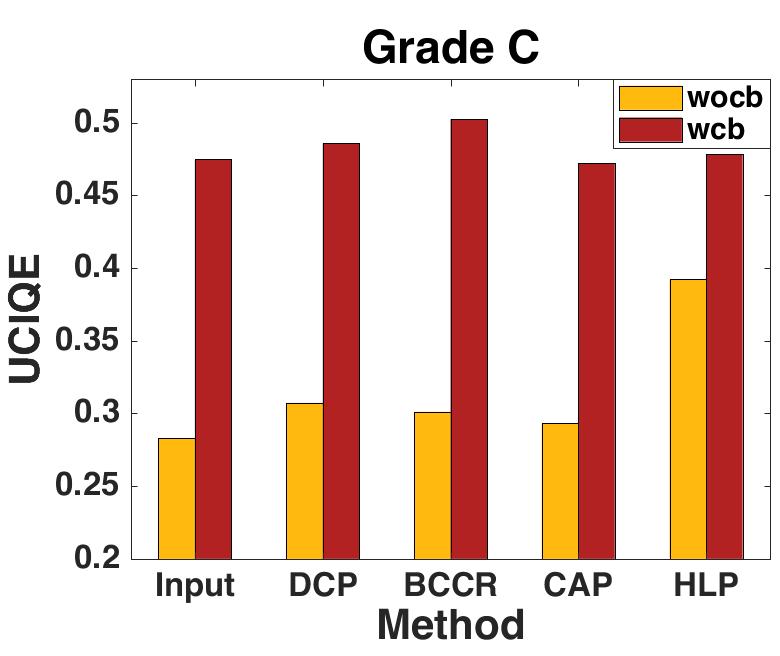}
			&\includegraphics[width=.2\textwidth]{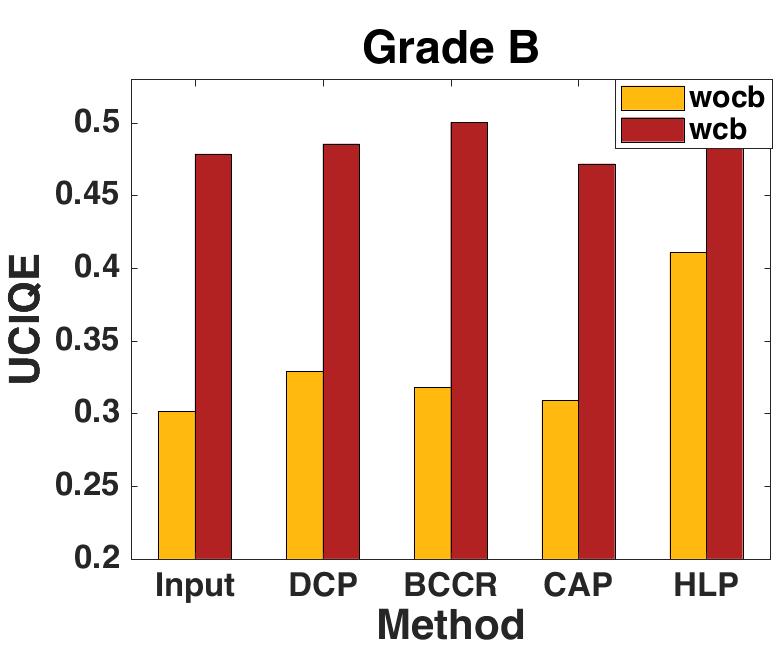}
			&\includegraphics[width=.2\textwidth]{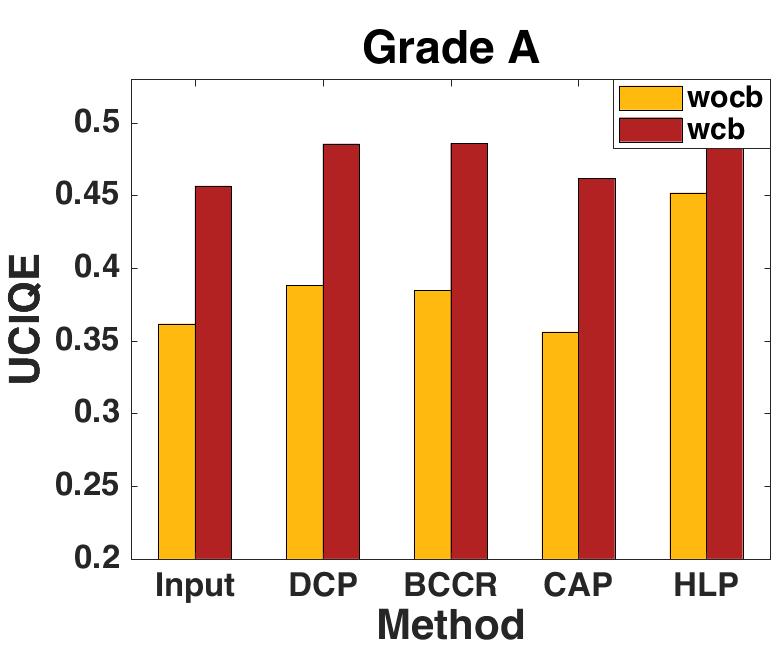}\\
			\includegraphics[width=.2\textwidth]{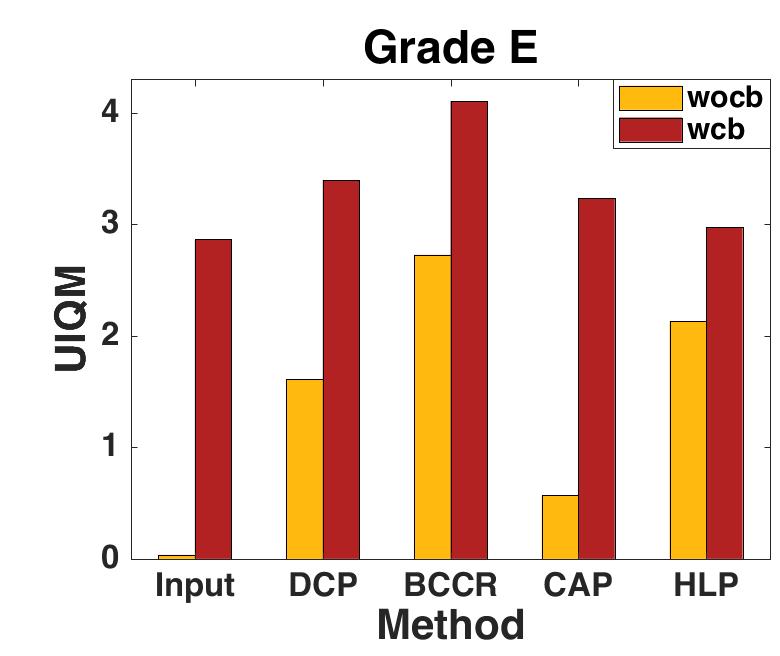}
			&\includegraphics[width=.2\textwidth]{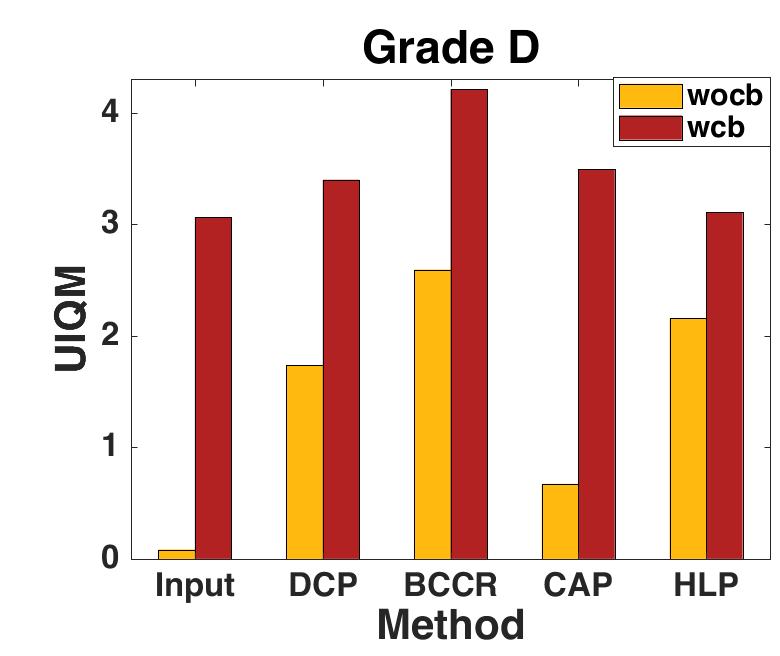}	
			&\includegraphics[width=.2\textwidth]{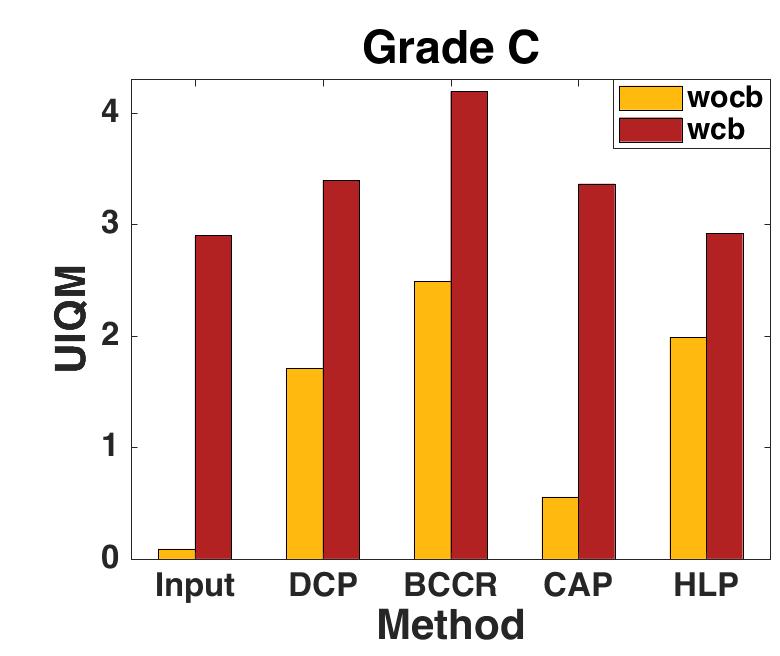}
			&\includegraphics[width=.2\textwidth]{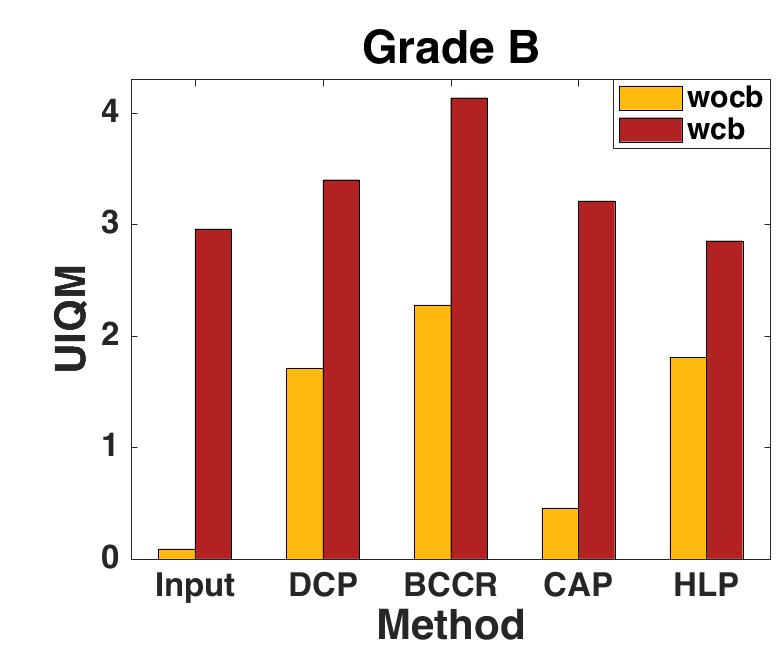}
			&\includegraphics[width=.2\textwidth]{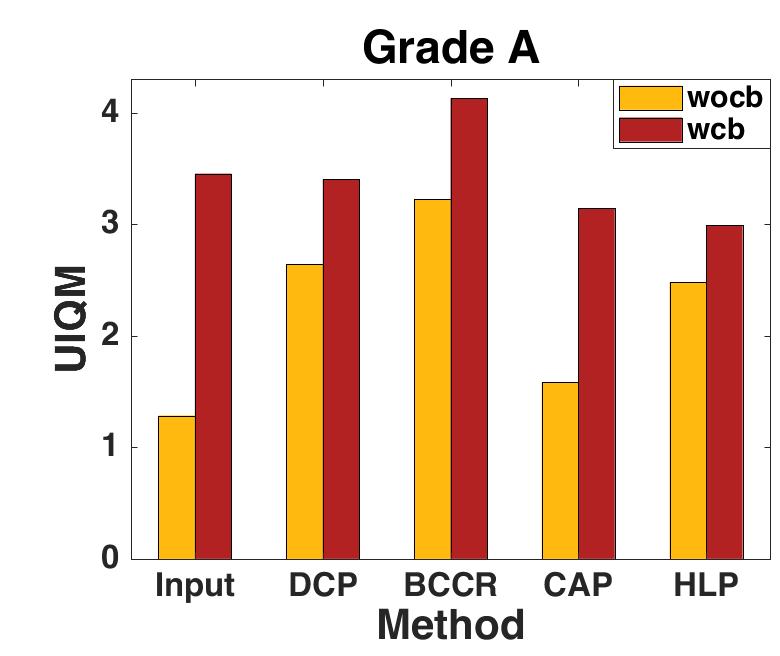}
			\\		
		\end{tabular}
	\end{center}
	\caption{Comparison of ``underwater dehazing without color balance (wocb)'' strategy and the ``dehazing with color balance (wcb)'' strategy on the five sub-datasets of UIQS.}\label{fig:bar_cb}		
\end{figure*}
Different lighting conditions, water depths, and levels of salinity produce significant changes on color tones of underwater images. Correcting to a natural tone is one of the important objectives for UIE. Therefore, we construct UCCS having great diversities of color tones, and use images of UCCS to evaluate the capabilities of UIE algorithms for color correction. 

\textbf{Qualitative comparisons:}
Figure~\ref{fig:tone_compare} demonstrates representative resultant images of the eleven methods performing on UCCS. At this point, we focus on the ability to correct color distortions. MSRCR can correct both greenish and bluish tones well. As for CLAHE and Fusion, the ability to handle greenish tones is superior than that to blue tones. BP enhances the contrast in bluish scenes, but tends to failure on greenish pictures. UHP may produce partial darkness, and NOM always generates over-corrected reddish results. Among the four model-based algorithms with direct applications of dehazing priors, $\text{DCP}_{cb}$ and $\text{BCCR}_{cb}$ can correct blue tone well giving more natural results, while $\text{HLP}_{cb}$ performs the best when dealing with low illumination and greenish tone.

\textbf{Quantitative comparisons:}
We also quantitatively evaluate the color correction ability of these methods by two metrics $ Avg_a$ and $Avg_b$, i.e., the average values of the channel a and b in the CIElab space, respectively. The metric $Avg_a$ characterizes the component of green to red, with the green in the negative direction while the red in the positive one. Similarly, $Avg_b$ represents the component from the blue to yellow. The values close to zero of $ Avg_a$ and $ Avg_b$ indicates low color cast. Table~\ref{table:RUI-I} gives the quantitative scores of the inputs images and outputs of the eleven UIE algorithms averaged in the three sub-datasets of UCCS. The most serious green bias appears in the ``Green'' set, and the green bias of the ``Blue'' set is not much different from that of the ``Green-blue''. In comparison, the blue bias of the ``Blue'' set is much more serious, and the three subsets are gradually positive biased in the blue-yellow component.

MSRCR performs the best on the ``Blue'' set, and exhibits excellent ability of correcting green bias in all three subsets, but tends to push $Avg_a$ and $Avg_b$ to positive values so that the results appear to be visually reddish. $\text{DCP}_{cb}$ and $\text{BCCR}_{cb}$ work better on the blue-green than the other tones. On the ``Green'' set, DPATN, $\text{DCP}_{cb}$, and the three model-free methods have good ability to correct green bias. Unfortunately, BP can handle none of the three subsets well, and sometimes produces resultant images showing extremely subtle difference with the input of low quality. NOM over-corrects color casts, resulting in poor visual effects. 

Considering both visual effects and quantitative analysis on UCCS, we can see that $\text{DCP}_{cb}$, $\text{BCCR}_{cb}$ and DPATN show more satisfactory performance on blueish images, while the model-free methods are more suitable for processing images with more green components.

\textbf{Discussion: Color correction and visibility improvement}

Color cast is one of challenging issues in underwater image enhancement as its effects vary with complex environmental factors such as water depth and salinity. The attenuation of red light through water causes the color cast as well as image details loss. This difficulty stumbles many UIE algorithms based on the transmission-related priors that work well in the context of image dehazing. For instance, the model-based BP algorithm performs poor on greenish images. On the other hand, the four-model based algorithms cascaded with a color balance module can achieve more appealing results. 

We further compare the performance of the strategies with and without color balance (CB)~\cite{Ancuti2012Enhancing} as postprocessing on the UIQS data set in order to peer into the impact from this simple technique tackling color cast. Figure~\ref{fig:bar_cb} demonstrates that every model-based dehazing-like UIE method somewhat improves the image quality. Among them, HLP increases the values of UCIQE the most, and BCCR is favored by UIQM the most. However, as shown in the first dark red bar of each chart of Fig.~\ref{fig:bar_cb}, even a single CB module improves the input of low quality much more than anyone of the four model-based algorithms without CB, i.e., DCP, BCCR, CAP and HLP, in terms of either UIQM or UCIQE. 

Figure~\ref{fig:bar_cb} also evidently illustrates that the strategy combining both model-based visibility enhancement and color correction works superior to either individual module. Different priors may exhibit different affinities with the color balance module. DCP and BCCR gain more evident increases when combing CB, while $\text{CAP}_{cb}$ and $\text{HLP}_{cb}$ produce a tiny gap over the single CB module especially on the C, D and E subsets showing severe degradation. Therefore, in the future, it is worthy developing a more elaborate scheme to collaborate these two modules instead of simple cascade. In a recent preliminary study, we jointly learn both prior-based transmission and color correction in a residual learning framework, showing promising results~\cite{hou2018joint}.

\begin{table}[t]
	\centering
	\caption{Average \textcolor{green}{$Avg_a$} / \textcolor{blue}{$Avg_b$} scores on UCCS, $Avg_a$ means the green-red component, with green in the negative direction. $Avg_b$ represents the blue-yellow component, with blue in the negative direction. The best two are shown in bold.} \label{table:tone}
	\begin{tabular}{lp{2cm}<{\centering}p{1.9cm}<{\centering}p{2.1cm}<{\centering}}\toprule
	Method&Blue& Green-blue& Green  \\\midrule
	Input& -25.84 / -6.56 & -24.36 / 4.24 & -30.97 / 12.10 \\
	MSRCR & \textbf{1.17} / \textbf{0.47} & 2.42 / 1.05 & 2.58 / \textbf{0.31} \\
	CLAHE & -10.95 / -2.71 & -6.73 / 1.67 & -1.68 / 1.46 \\
	Fusion & -10.27 / -2.67 & -6.07 / 1.66 & -1.21 / \textbf{1.42} \\
	BP & -24.23 / -5.90 & -23.15 / 4.85 & -29.77 / 12.35 \\
	UHP & -11.70 / -1.08 & -7.87 / 6.59 & -9.84 / 6.42 \\
	NOM & \textbf{0.88} / 6.90 & 35.25 / 9.05 & 36.01 / 22.35 \\
	DPATN & -10.15 / -3.14 & -4.16 / 2.03 & \textbf{-1.15} / 1.49 \\
	$\text{DCP}_{cb}$& -12.21 / -2.37 & \textbf{1.77} / \textbf{0.84} & \textbf{0.76} / 1.58 \\
	$\text{BCCR}_{cb}$& -8.31 / \textbf{1.00} & \textbf{1.59} / 4.76 & 3.76 / 2.43 \\
	$\text{CAP}_{cb}$& -15.83 / -2.15 & -4.69 / 5.46 & 1.70 / 2.79 \\
	$\text{HLP}_{cb}$ & -16.14 / -6.16 & -7.68 / \textbf{0.94} & -8.30 / 5.17 \\
	\bottomrule	
    \end{tabular}
\end{table}

\subsection{Higher-level Task-driven Comparison on UHTS}
We apply a common marine object classification module to the enhanced images given by the eleven UIE algorithm on UHTS, and evaluate the detection accuracy in terms of the mean Average Precision (mAP) and detection number~(Num). Several dehazing studies~\cite{li2017aod,li2019benchmarking,pei2018does} introduced similar task-driven evaluation on the performance of dehazing algorithms. Nevertheless, to our best knowledge, the task-specific evaluation on UIE algorithms still remains untouched, mainly attributing to no label of marine objects of interest available in any existing underwater benchmarks.

\begin{figure}[t]
	\begin{center}
		\begin{tabular}{c@{\extracolsep{0.3em}}c}
			\includegraphics[width=.235\textwidth]{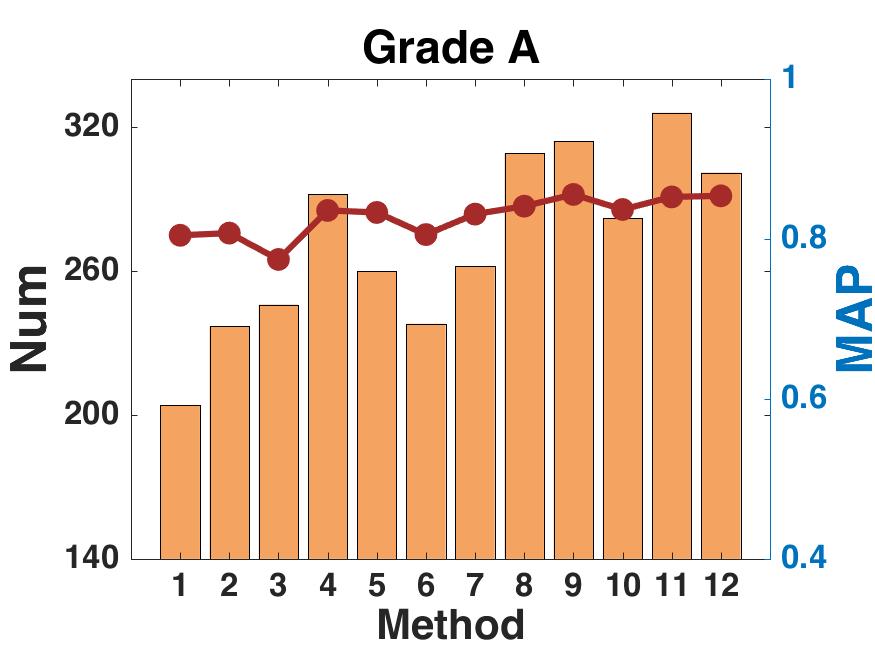}
			&\includegraphics[width=.235\textwidth]{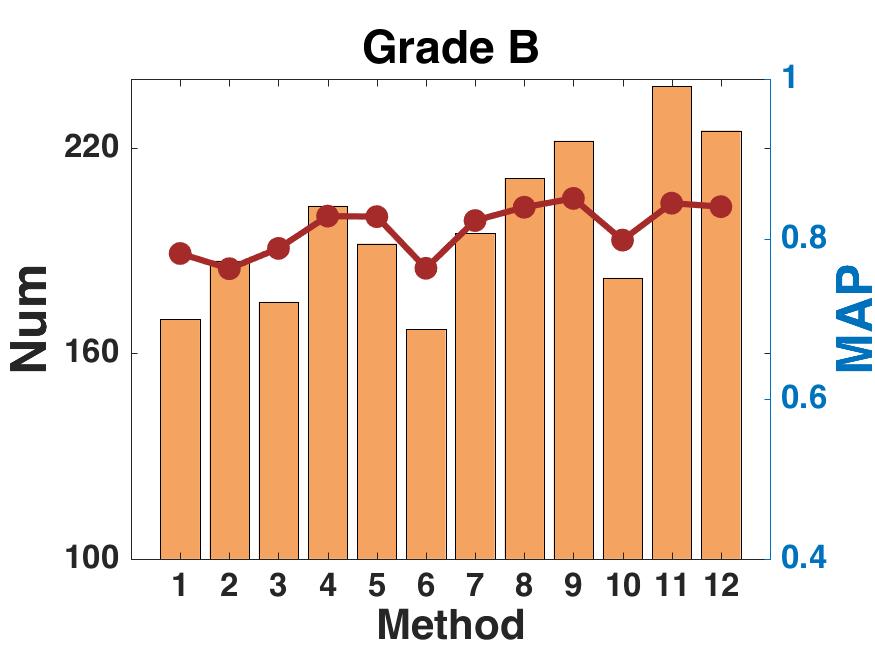}\\	
			\includegraphics[width=.235\textwidth]{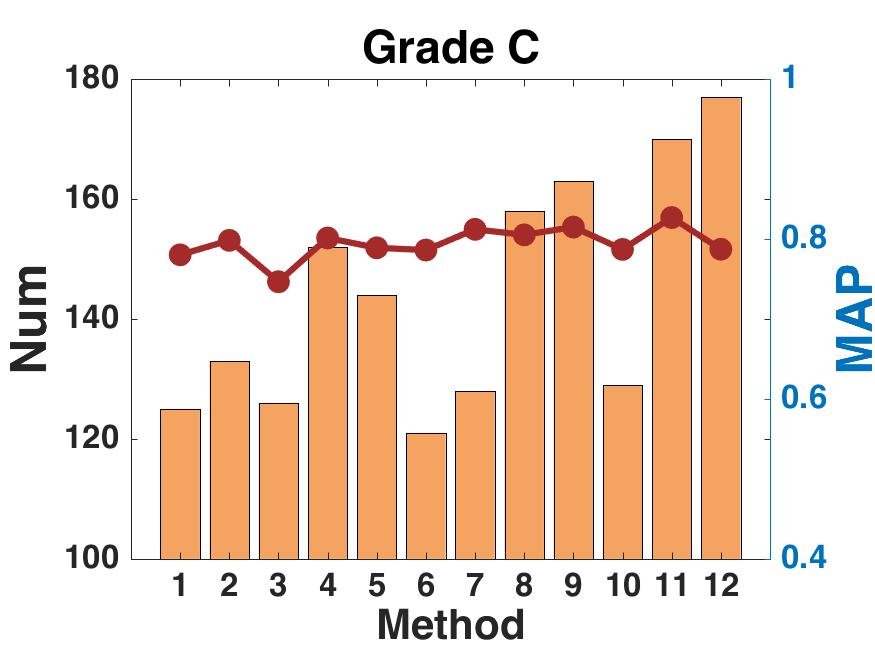}
			&\includegraphics[width=.235\textwidth]{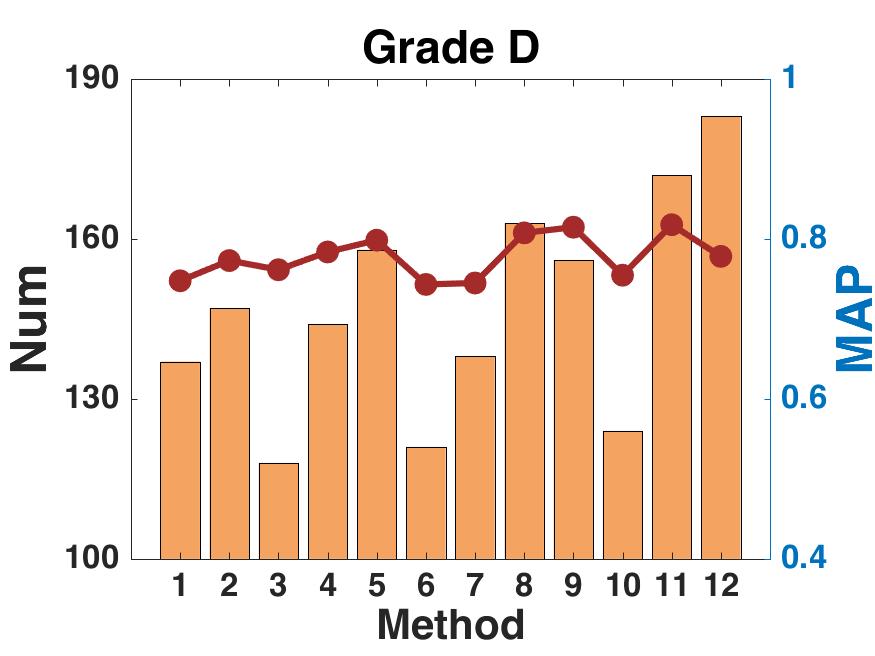}\\
			\includegraphics[width=.235\textwidth]{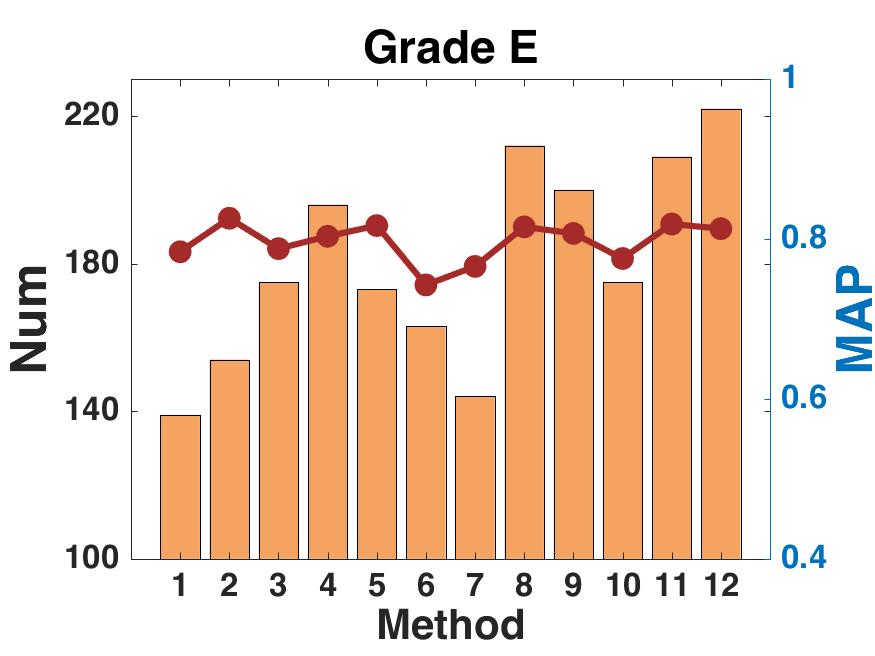}
			&\includegraphics[width=.235\textwidth]{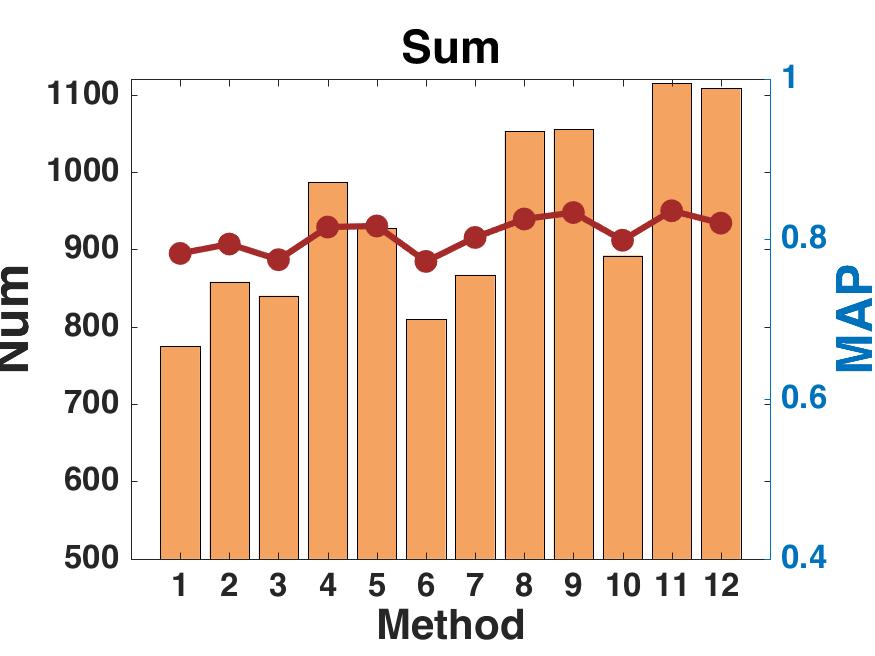}
			\\
		\end{tabular}
	\end{center}
	\caption{Object detection number and mAP on UHTS. The histogram represents the detection number and the polyline represents mAP. Number 1 to 12 are respectively Input, MSRCR, CLAHE, Fusion, BP, UHP, NOM, DPATN, $\text{DCP}_{cb}$, $\text{BCCR}_{cb}$, $\text{CAP}_{cb}$ $\text{HLP}_{cb}$.
	}\label{fig:detect_real_sort}		
\end{figure}

We labeled $1,800$ underwater images and re-trained the YOLO-V3~\cite{redmon2018yolov3} network to detect and classify three types of marine objects, i.e., sea urchins, sea cucumbers, and scallops. Figure~\ref{fig:detect_real_sort} lists the values of mAP and Num. Most UIE algorithms can improve the detection numbers, but hardly bring significant benefits to mAP. The model-based algorithms, $\text{BCCR}_{cb}$ and $\text{HLP}_{cb}$, perform the best by significantly improving Num while stably increasing the values of mAP on all the subsets of UHTS. Model-free CLAHE and Fusion play a positive role for Num, and CLAHE keeps or slightly increases mAP while Fusion has a negative effect on mAP. Unfortunately, NOM performs little or even negative on both mAP and Num, especially on the subsets of B and C. 

\begin{figure*}[!htbp]
	\begin{center}
		\begin{tabular}{c@{\extracolsep{0.4em}}c@{\extracolsep{0.4em}}c@{\extracolsep{0.4em}}c@{\extracolsep{0.4em}}c@{\extracolsep{0.4em}}c@{\extracolsep{0.4em}}c}
			\includegraphics[width=.155\textwidth]{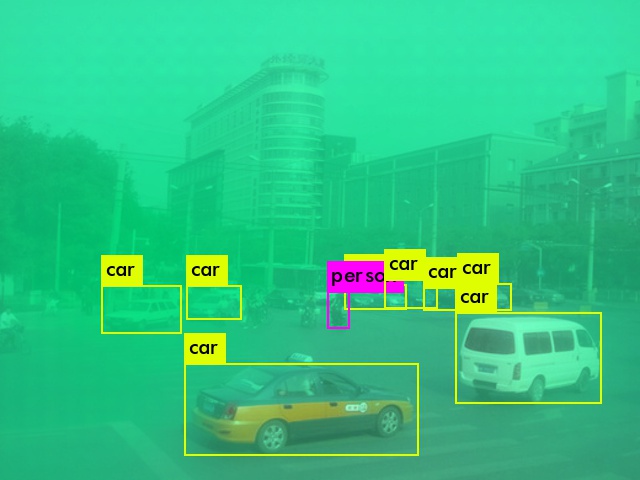}
			&\includegraphics[width=.155\textwidth]{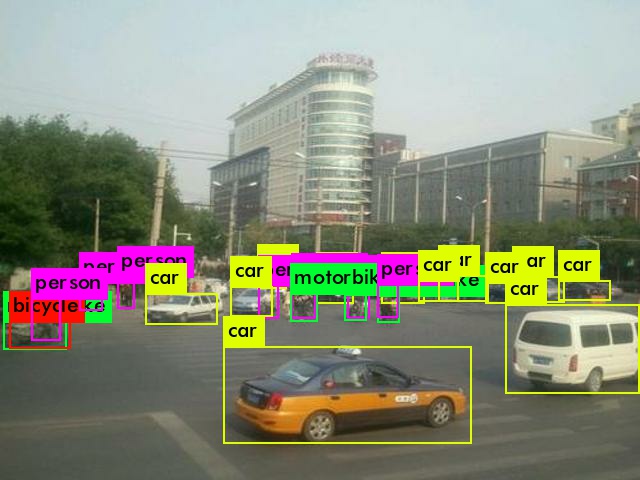}	
			&\includegraphics[width=.155\textwidth]{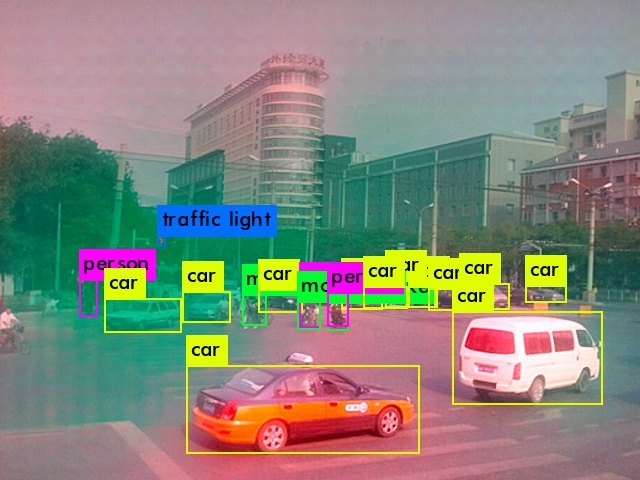}
			&\includegraphics[width=.155\textwidth]{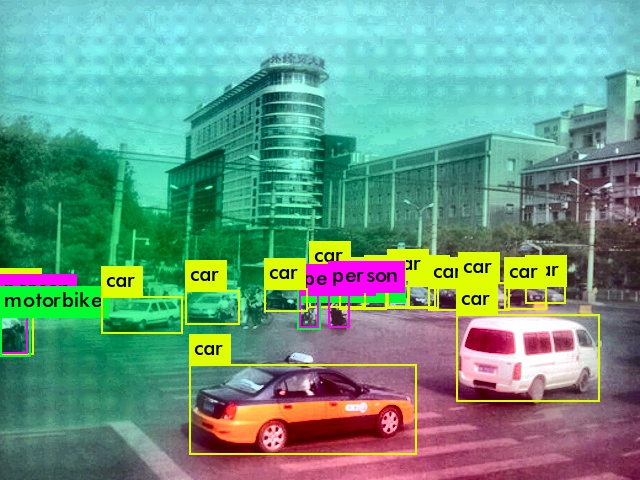}
			&\includegraphics[width=.155\textwidth]{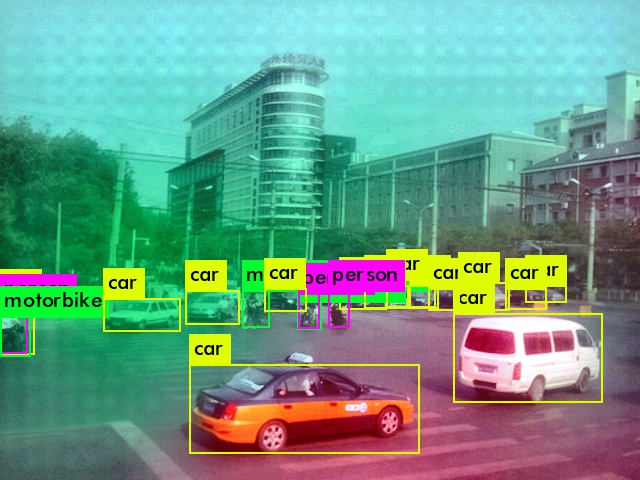}
			&\includegraphics[width=.155\textwidth]{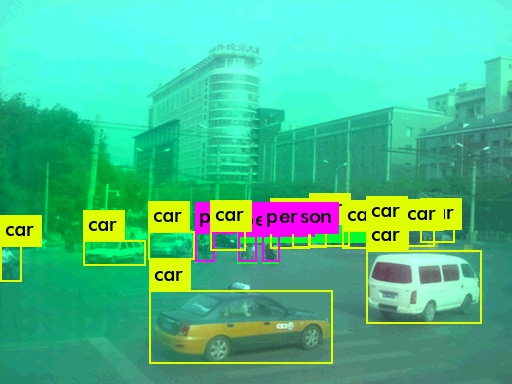}
			\\	
			\includegraphics[width=.155\textwidth]{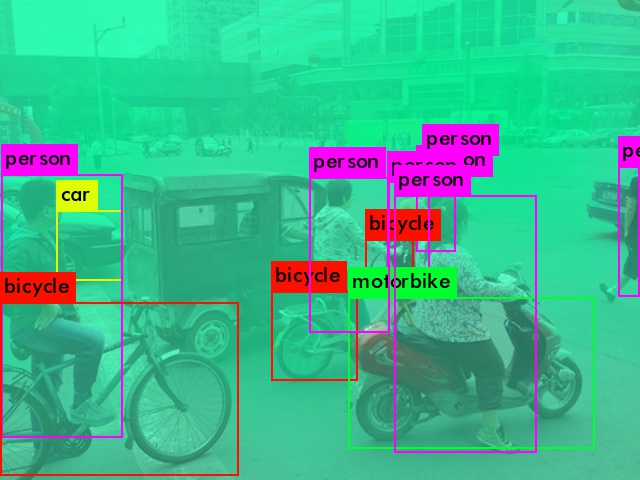}
			&\includegraphics[width=.155\textwidth]{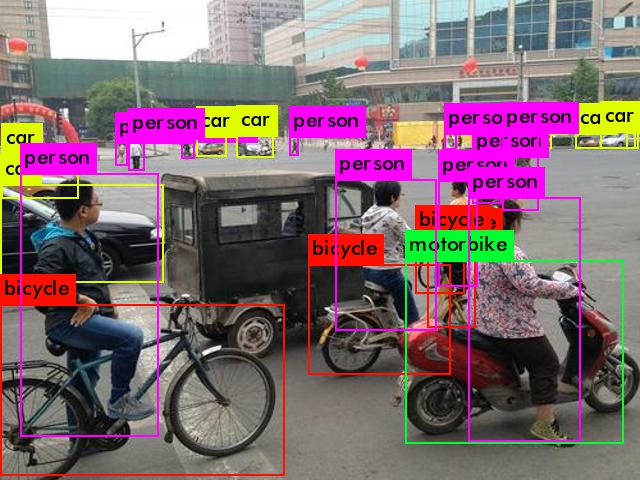}	
			&\includegraphics[width=.155\textwidth]{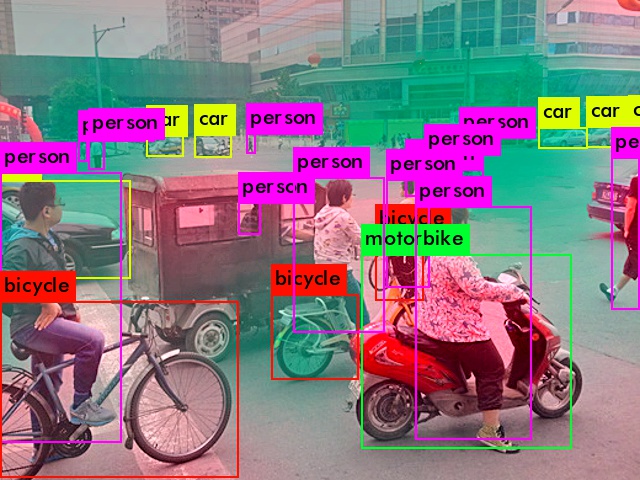}
			&\includegraphics[width=.155\textwidth]{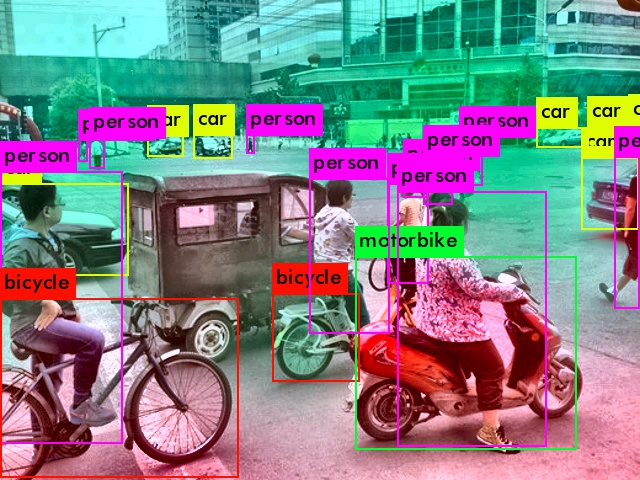}
			&\includegraphics[width=.155\textwidth]{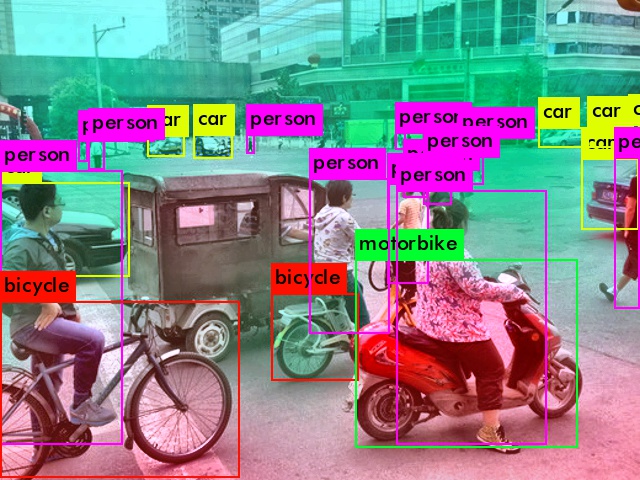}
			&\includegraphics[width=.155\textwidth]{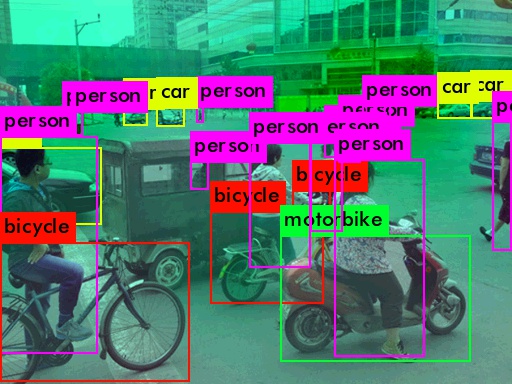}
			\\
			Input & GT &MSRCR &CLAHE &Fusion &BP\\
			\includegraphics[width=.155\textwidth]{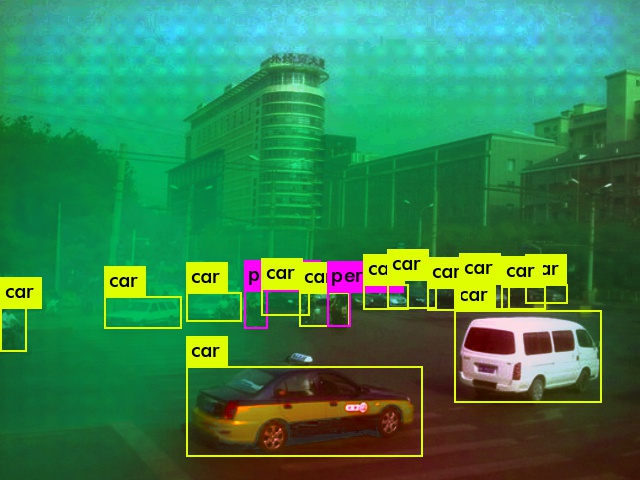}
			&\includegraphics[width=.155\textwidth]{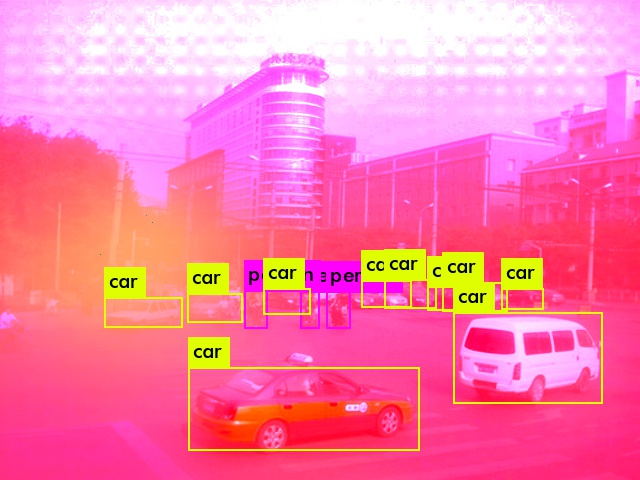}	
			&\includegraphics[width=.155\textwidth]{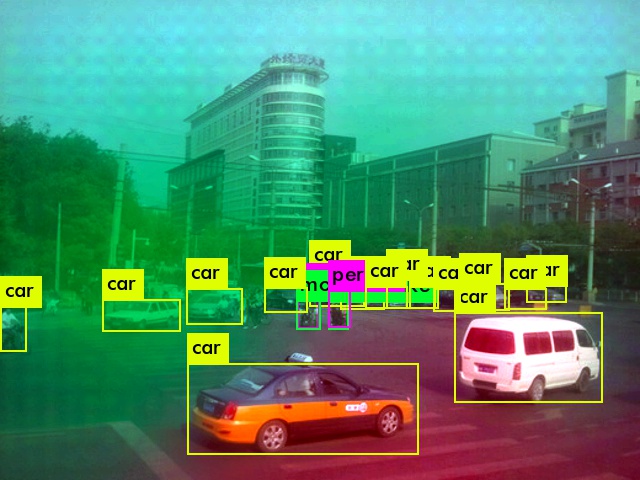}
			&\includegraphics[width=.155\textwidth]{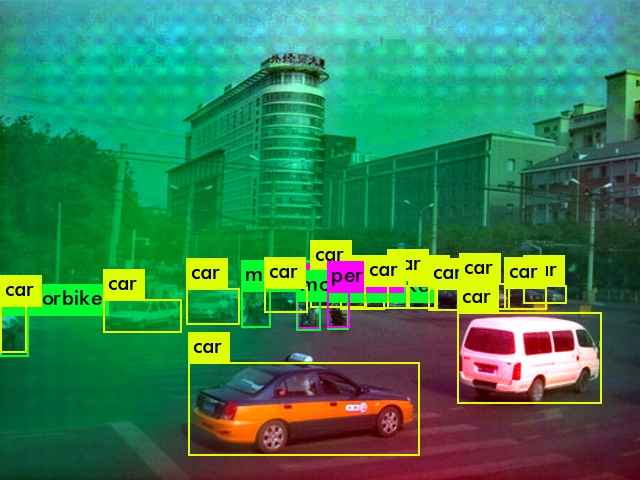}
			&\includegraphics[width=.155\textwidth]{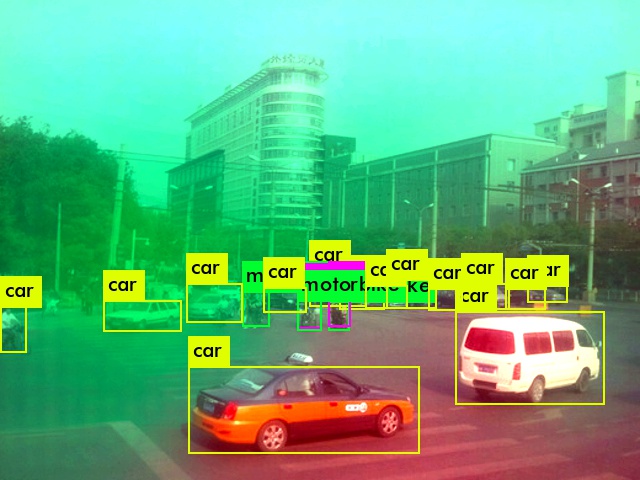}
			&\includegraphics[width=.155\textwidth]{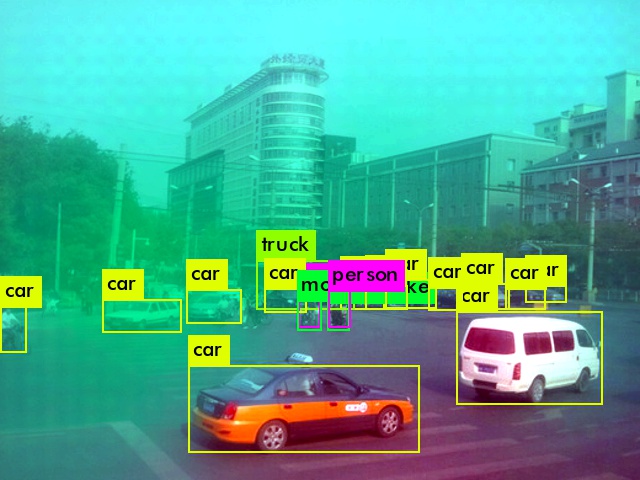}
			\\
			\includegraphics[width=.155\textwidth]{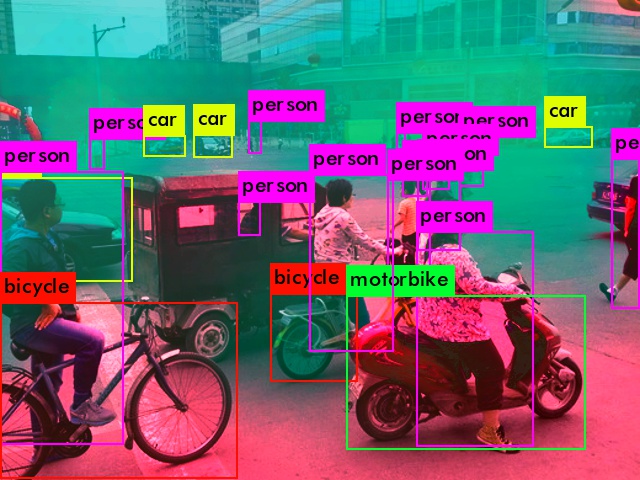}
			&\includegraphics[width=.155\textwidth]{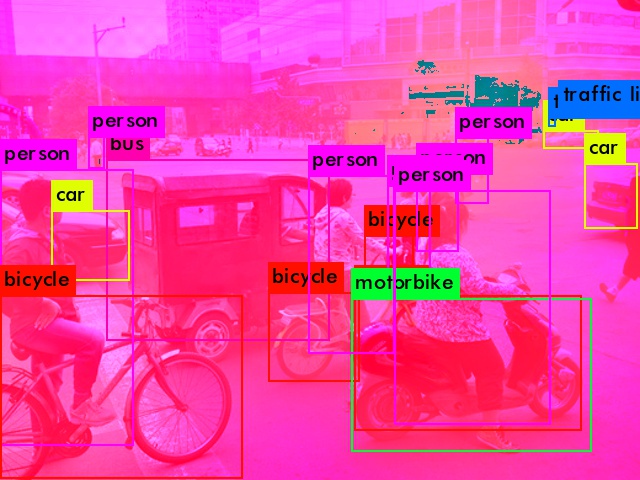}	
			&\includegraphics[width=.155\textwidth]{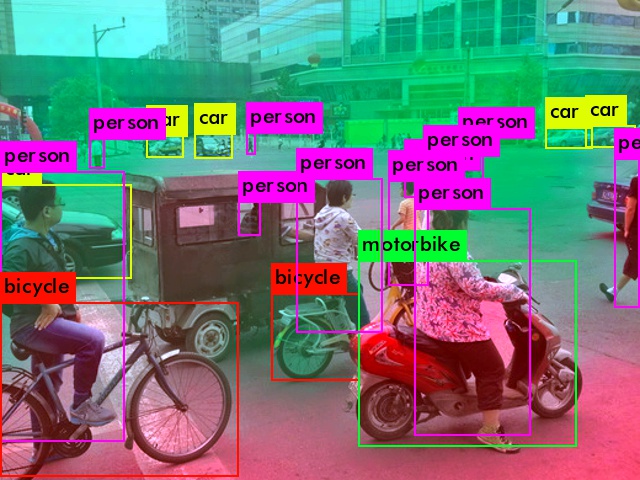}
			&\includegraphics[width=.155\textwidth]{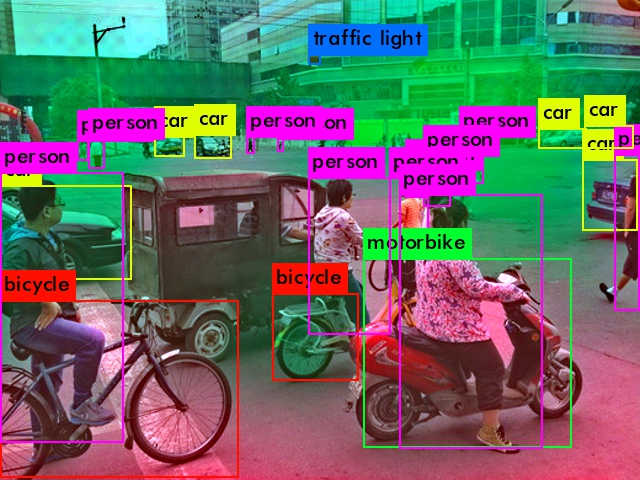}
			&\includegraphics[width=.155\textwidth]{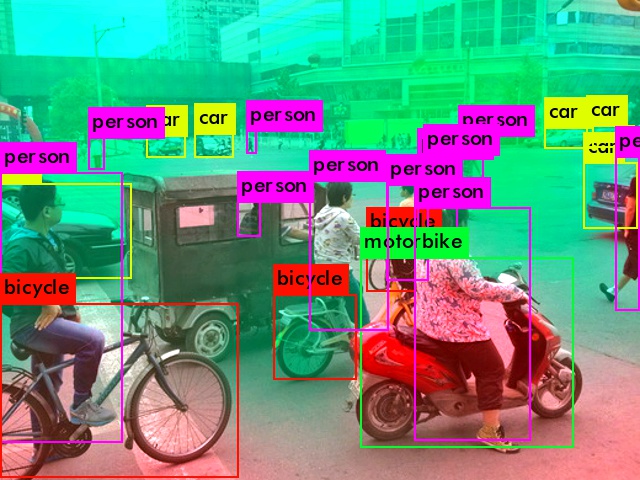}
			&\includegraphics[width=.155\textwidth]{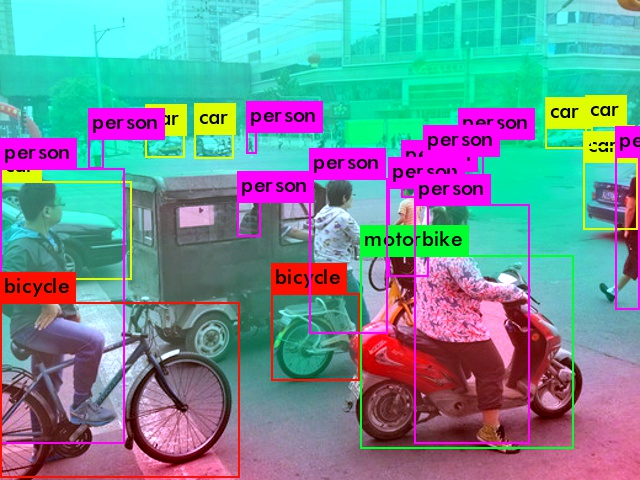}
			\\
			UHP &NOM & $\text{DCP}_{cb}$ & $\text{BCCR}_{cb}$ &$\text{CAP}_{cb}$ &$\text{HLP}_{cb}$
			
		\end{tabular}
	\end{center}
	\caption{Visualization of object detection results ( on synthetic dataset) after applying different dehazing algorithms.
	}\label{fig:detect-syn}		
\end{figure*}

Furthermore, we employed WaterGan~\cite{li2018watergan} to \emph{synthesize} sixty underwater images containing 21 classes of outdoor objects (e.g., bicycle, person, and bus) by using $3,000$ randomly selected real world underwater images from UIQS and $3,000$ outdoor RGB-D images~\cite{li2019benchmarking}. Subsequently, we applied the original trained YOLO-V3 model in \cite{redmon2018yolov3} to detect the outdoor objects from the resultant images of the eleven UIE algorithms. Table \ref{table:detect_syn} and Fig.~\ref{fig:detect-syn} provide the detection results on this synthetic data set. The mAP results on the synthetic data in Table~\ref{table:detect_syn} are consistent with those on the real world set UHTS. The model-based UIE algorithms followed by a simple color balance module, $\text{CAP}_{cb}$ and $\text{DCP}_{cb}$, and the aggregated prior-data network, DPATN, obtain higher mAP values, while model-free based MSRCR, UHP and NOM increase little or even decrease mAP.

\textbf{Discussion: The role of underwater image enhancement on higher-level object detection}

\begin{figure*}[t]
	\begin{center}
		\begin{tabular}{c@{\extracolsep{0em}}c@{\extracolsep{0em}}c@{\extracolsep{0em}}c@{\extracolsep{0em}}c@{\extracolsep{0em}}c@{\extracolsep{0em}}c}
			\includegraphics[width=.2\textwidth]{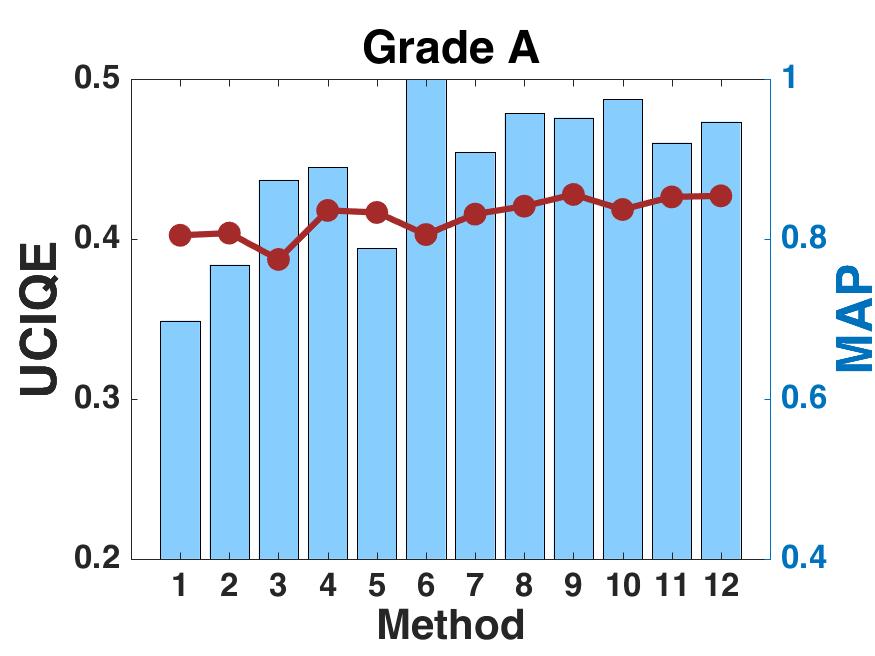}
			&\includegraphics[width=.2\textwidth]{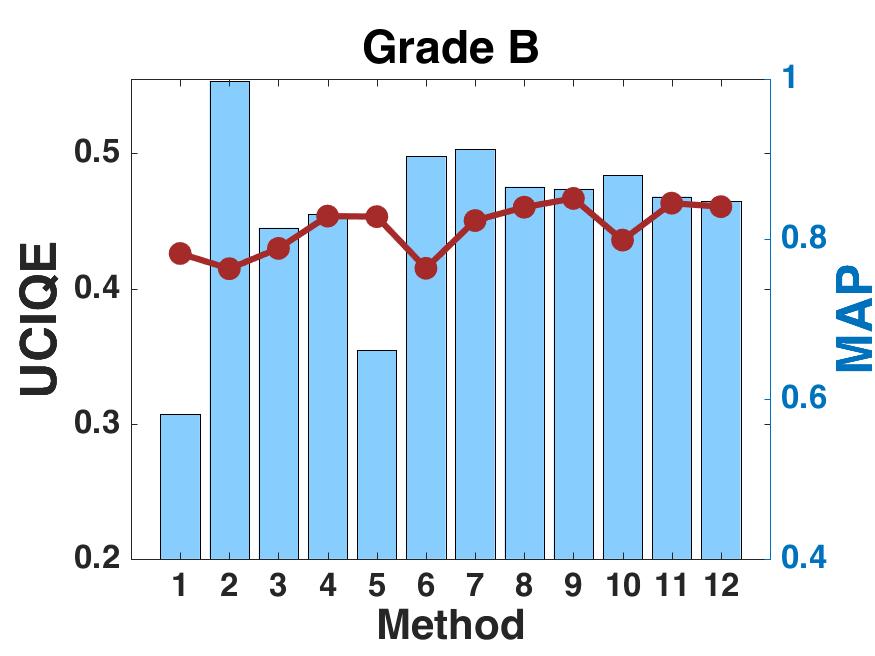}	
			&\includegraphics[width=.2\textwidth]{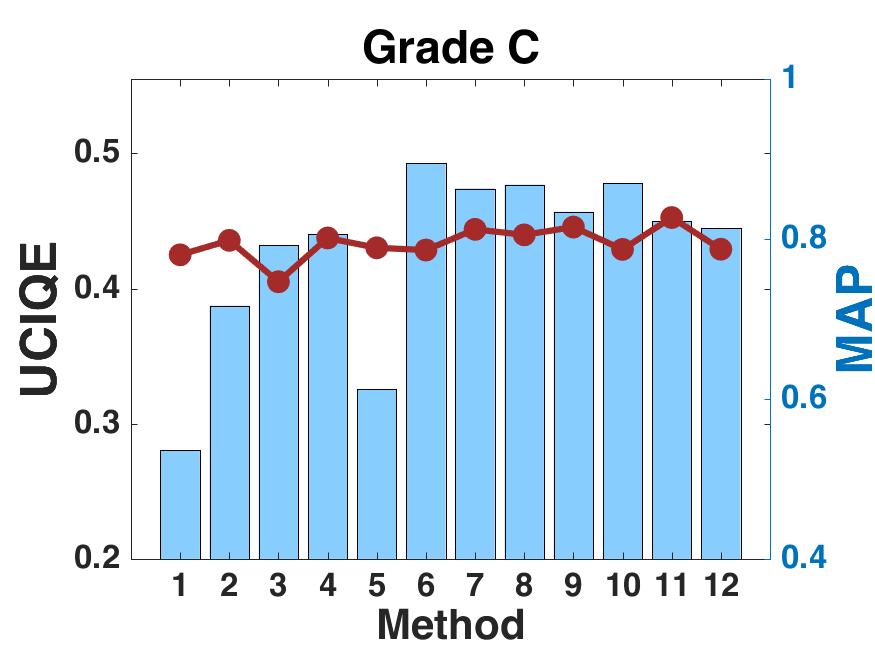}
			&\includegraphics[width=.2\textwidth]{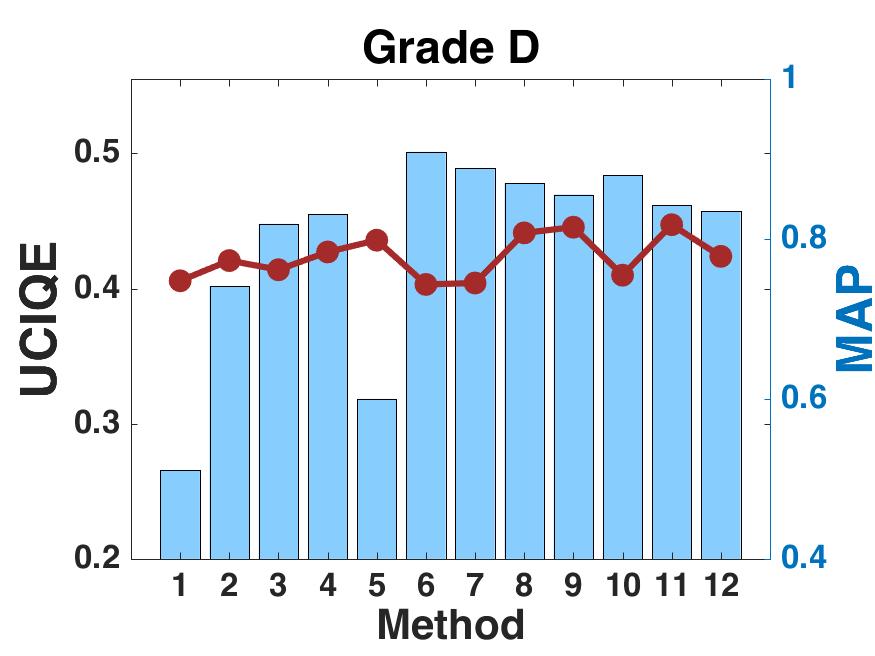}
			&\includegraphics[width=.2\textwidth]{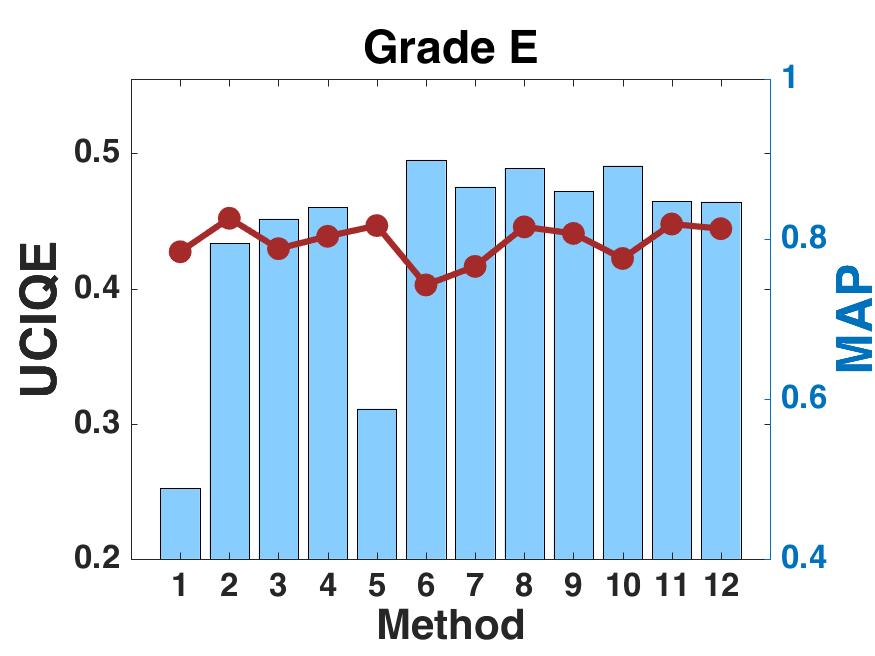}\\
			\includegraphics[width=.2\textwidth]{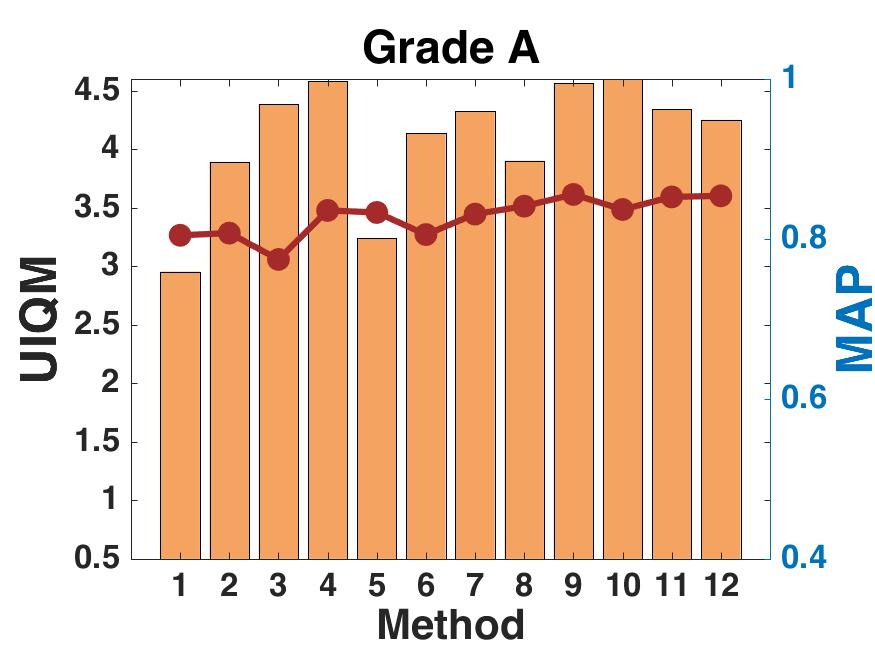}
			&\includegraphics[width=.2\textwidth]{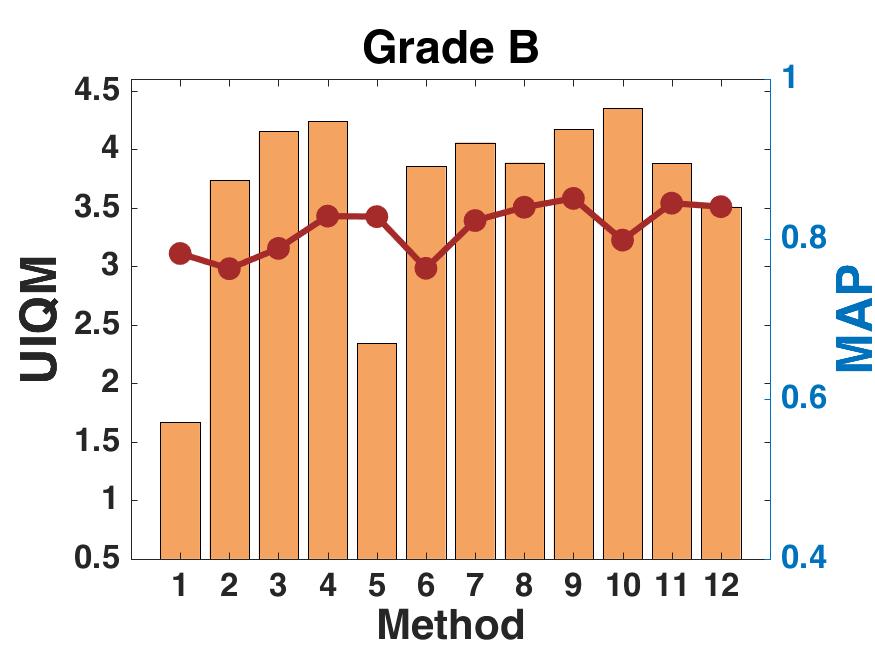}	
			&\includegraphics[width=.2\textwidth]{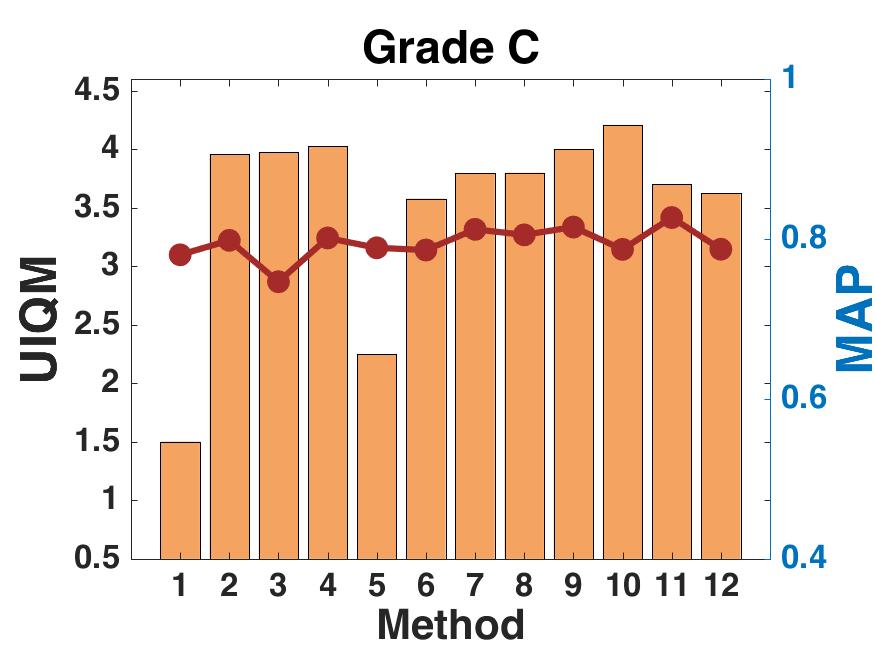}
			&\includegraphics[width=.2\textwidth]{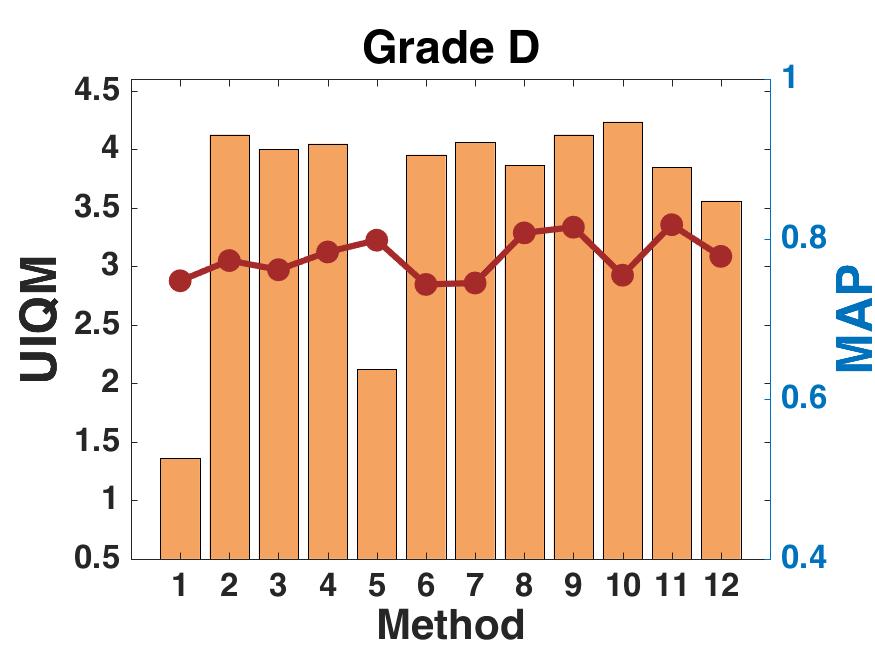}
			&\includegraphics[width=.2\textwidth]{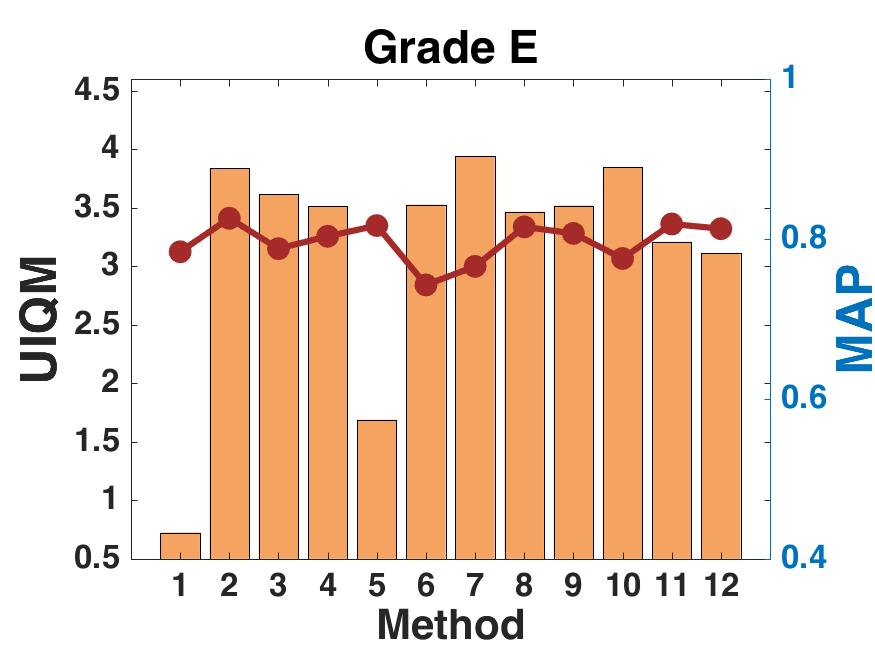}
			\\		
		\end{tabular}
	\end{center}
	\caption{The mAP and UCIQE/ UIQM scores on UHTS. The histogram represents UCIQW/UIQM scores and the polyline represents mAP, and methods 1 to 12 are respectively Input, MSRCR, CLAHE, Fusion, BP, UHP, NOM, DPATN, $\text{DCP}_{cb}$, $\text{BCCR}_{cb}$, $\text{CAP}_{cb}$ $\text{HLP}_{cb}$.
	}\label{fig:detect_score_prob}		
\end{figure*}

We also conducted experiments on UHTS to investigate how the detection accuracy is related to the no-reference image quality metric, both given by the enhanced images of UIE algorithms. A weak correlation occurs in Fig.~\ref{fig:detect_score_prob} showing the comparisons of the mAP results with the average UCIQE/UIQM scores on UHTS. On the subset B, both CLAHE and Fusion can greatly improve the UCIQE and UIQM scores, but their mAP results sometimes are even lower than those of degraded inputs. On the subsets of A, D and E, UHP significantly improves mAP, but its quality evaluation is the worst among all methods. Therefore, comparisons on the higher-level tasks might be necessary to comprehensively evaluate UIE algorithms. 

Pei \emph{et al.} reported a similar phenomenon on dehazing that most existing dehazing algorithms cannot significantly improve the accuracy for higher-level image classification~\cite{pei2018does}. The objective of low-level enhancement typically differs from that of classification so that enhancement algorithms can hardly recover features favoring higher-level tasks.  Therefore, higher-level classification may prefer to train end-to-end deep models directly from labeled degraded examples other than two separate steps. There are two possible approaches to obtaining training examples that are critical to this end-to-end model: one is to directly label real world degraded images; the other is to use labeled images of natural scene together with unlabeled real world underwater images to synthesize degraded training examples with abundant labels, which requires designated GAN. For this respect, real-world images with different levels of image quality in our RUIE are not only helpful for evaluating low-level UIE but also boosting high-level tasks.

\begin{figure*}[!htbp]
	\begin{center}
		\begin{tabular}{c@{\extracolsep{0.4em}}c@{\extracolsep{0.4em}}c@{\extracolsep{0.4em}}c@{\extracolsep{0.4em}}c@{\extracolsep{0.4em}}c@{\extracolsep{0.4em}}c}	
			\includegraphics[width=.155\textwidth]{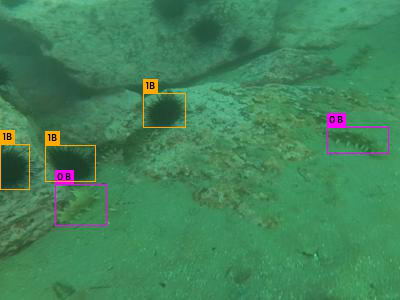}
			&\includegraphics[width=.155\textwidth]{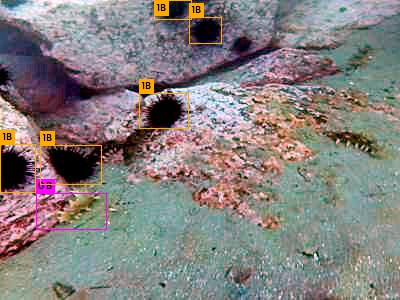}
			&\includegraphics[width=.155\textwidth]{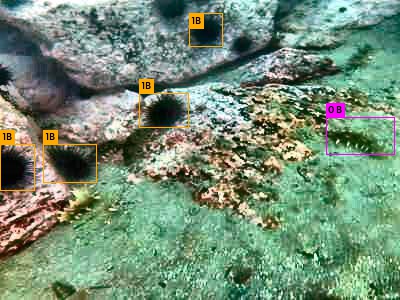}
			&\includegraphics[width=.155\textwidth]{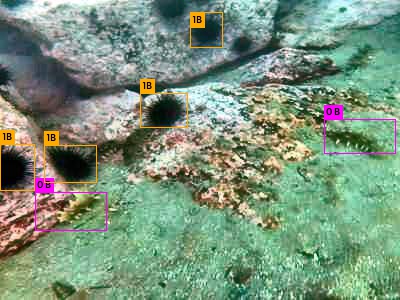}
			&\includegraphics[width=.155\textwidth]{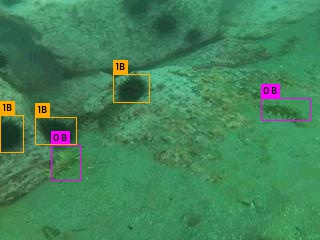}
			&\includegraphics[width=.155\textwidth]{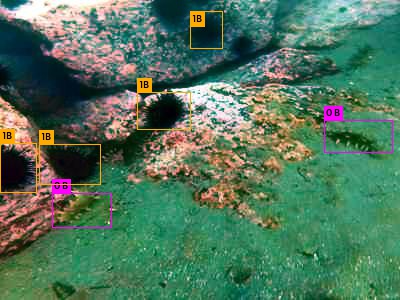}
			\\
			\includegraphics[width=.155\textwidth]{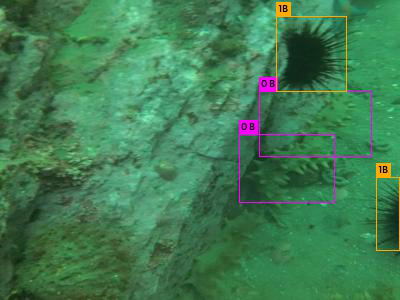}	
			&\includegraphics[width=.155\textwidth]{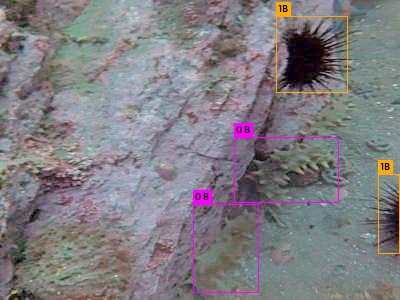}
			&\includegraphics[width=.155\textwidth]{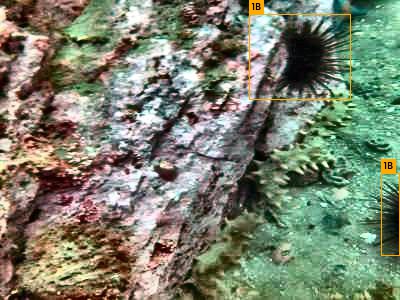}
			&\includegraphics[width=.155\textwidth]{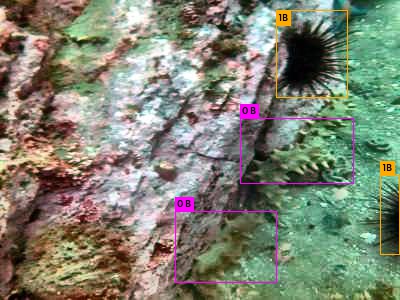}
			&\includegraphics[width=.155\textwidth]{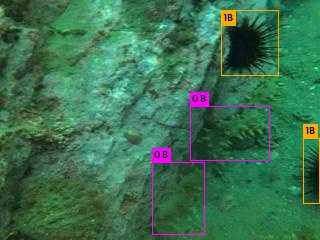}
			&\includegraphics[width=.155\textwidth]{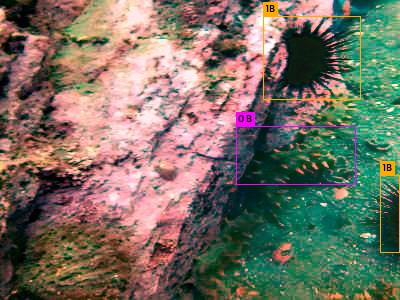}
			\\
			Input&MSRCR &CLAHE &Fusion &BP&UHP \\
			\includegraphics[width=.155\textwidth]{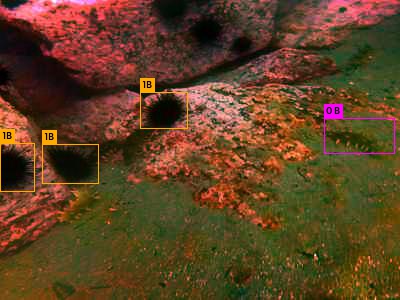}				
			&\includegraphics[width=.155\textwidth]{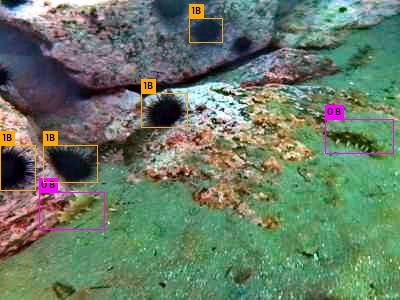}	
			&\includegraphics[width=.155\textwidth]{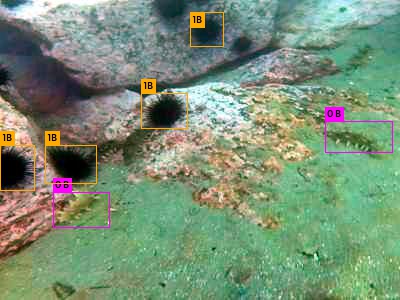}
			&\includegraphics[width=.155\textwidth]{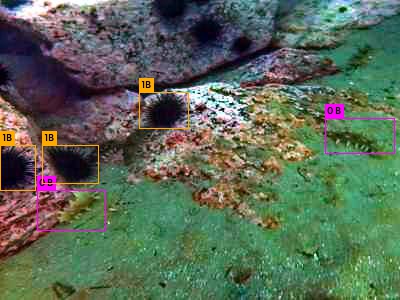}
			&\includegraphics[width=.155\textwidth]{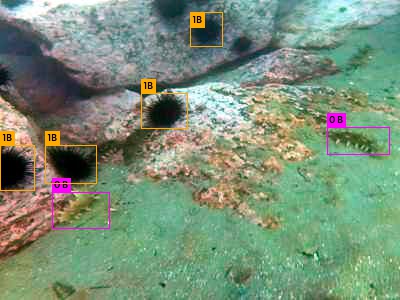}
			&\includegraphics[width=.155\textwidth]{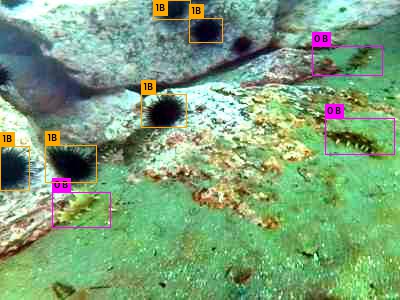}
			\\			
			\includegraphics[width=.155\textwidth]{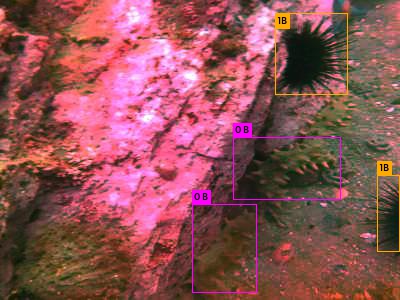}
			&\includegraphics[width=.155\textwidth]{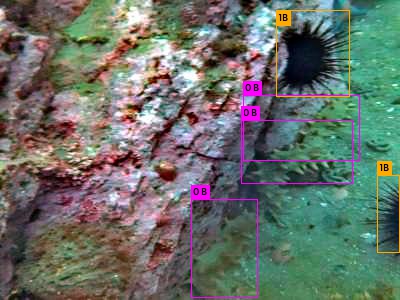}		
			&\includegraphics[width=.155\textwidth]{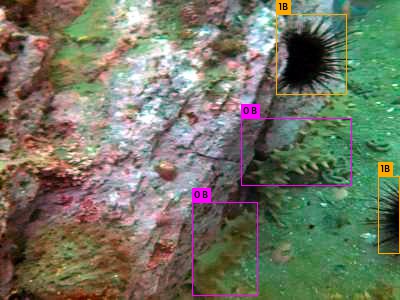}
			&\includegraphics[width=.155\textwidth]{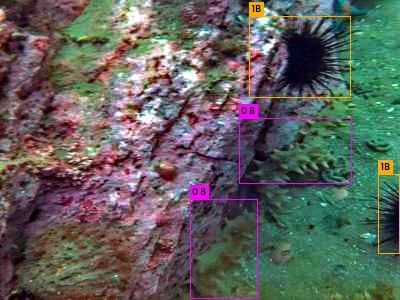}
			&\includegraphics[width=.155\textwidth]{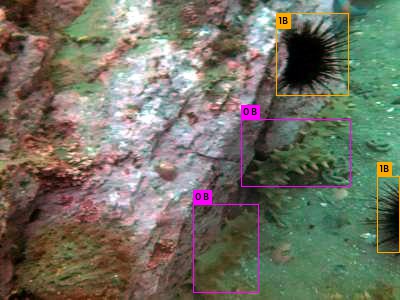}
			&\includegraphics[width=.155\textwidth]{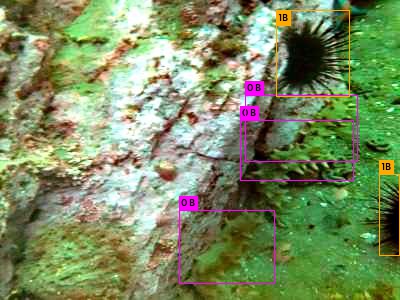}		
			\\
			NOM & DPATN& $\text{DCP}_{cb}$ & $\text{BCCR}_{cb}$ &$\text{CAP}_{cb}$ &$\text{HLP}_{cb}$			
		\end{tabular}
	\end{center}
	\caption{Comparison on UHTS. 
	}\label{fig:detect_real_image}		
\end{figure*}

\begin{table*}[!htbp]
	\centering
	\caption{MAP and total detection number on the synthetic dataset.} \label{table:detect_syn}
	\begin{tabular}{lp{0.7cm}<{\centering}p{0.7cm}<{\centering}p{0.9cm}<{\centering}p{0.9cm}<{\centering}p{0.8cm}<{\centering}p{0.8cm}<{\centering}p{0.8cm}<{\centering}p{0.8cm}<{\centering}p{0.8cm}<{\centering}p{0.8cm}<{\centering}p{0.8cm}<{\centering}p{0.8cm}<{\centering}p{0.8cm}<{\centering}}\toprule
		Category&Input& GT & MSRCR &CLAHE &Fusion &BP& UHP &NOM & DPATN & $\text{DCP}_{cb}$ & $\text{BCCR}_{cb}$ &$\text{CAP}_{cb}$ &$\text{HLP}_{cb}$\\ \midrule
		mAP &0.703 &0.725 &0.707 &0.720 &0.717 &0.714 &	0.702 &	0.695 &	0.723&0.\textbf{726} &	0.721 &0.\textbf{725} &	0.721\\
		Num&650 & 1038 & \textbf{893} & 882 & \textbf{888} & 767 & 733 & 668 & 804& 823 & 837 & 819 &  835\\	
		\bottomrule
	\end{tabular}
\end{table*}

\begin{table*}[!htbp]
	\centering
	\caption{Average running time on 300$\times $400 images. } \label{table:time}
	\begin{tabular}{lp{1cm}<{\centering}p{1cm}<{\centering}p{1cm}<{\centering}p{1cm}<{\centering}p{1cm}<{\centering}p{1cm}<{\centering}p{1cm}<{\centering}p{1cm}<{\centering}p{1cm}<{\centering}p{1cm}<{\centering}p{1cm}<{\centering}}\toprule
		Method&MSRCR &CLAHE &Fusion &BP& UHP &NOM & DPATN&$\text{DCP}_{cb}$ & $\text{BCCR}_{cb}$ &$\text{CAP}_{cb}$ &$\text{HLP}_{cb}$\\ \midrule
		Time(s)&0.0119&  0.0126&   0.0232 & 4.5847&  0.7985 & 0.6012 & 0.8758 & 0.8565 & 0.6568 & 0.5112& 0.4256\\
		
		\bottomrule
	\end{tabular}
\end{table*}

\section{Conclusions and Future Work  }
\label{sec:conclusions}
In this paper, we setup an undersea imaging system and construct the RUIE benchmark. The benchmark consists of three subsets UIQS, UCCS and UHTS,  targeting at the three challenging aspects or tasks for enhancement, i.e., visibility degradation, color cast, and higher-level detection/classification, respectively. Moreover, we evaluate eleven representative UIE algorithms on these three subsets in terms of various metrics on respective tasks. The experimental results demonstrate that no one single UIE algorithm can work the best for all tasks upon all criteria. UHP and $ \text{BCCR}_{cb}$ yield high scores of UCIQE and UIQM; DCP, CLAHE and Fusion are the most competitive in terms of color correction; $\text{UHLP}_{cb}$ presents the most appealing subjective quality; $\text{UHLP}_{cb}$ and $ \text{BCCR}_{cb}$ produce superior detection performance on real world underwater images. Besides, no strong correlation exists between the image quality assessment and detection accuracy (mAP). Therefore, tremendous efforts are highly demanded to more effective quality assessment, more elaborate paradigms simultaneously improving visibility and correcting color cast, and designated deep networks with more accurate detection/classification for underwater objects.

We advocate to develop data-driven non-reference assessment incorporating human perception, joint learning framework for visibility enhancement and color correction, and end-to-end networks for underwater object detection/classification in the future. All these possible directions would benefit from the proposed benchmark having a large number of real world images as well as showing a wide range of diversities.
We also expect this study to draw more interest from the computer vision community to work toward the challenging underwater tasks.

\section*{Acknowledgments}
Dalian Zhangzidao Group and Zhongji Oceanic Engineering Group provide great support for setting up the undersea image capturing system.

\bibliographystyle{IEEEtran}
\bibliography{egbib}

\end{document}